\documentclass[nohyperref]{article}

\usepackage{microtype}
\usepackage{graphicx}
\usepackage{booktabs} 

\usepackage{hyperref}
\usepackage{style}

\usepackage[accepted]{icml2023}

\usepackage{amsmath}
\usepackage{amssymb}
\usepackage{mathtools}
\usepackage{amsthm}

\usepackage[capitalize,noabbrev]{cleveref}

\theoremstyle{plain}

\theoremstyle{definition}

\theoremstyle{remark}

\icmltitlerunning{Searching Large Neighborhoods for ILPs with Contrastive Learning}

\begin{document}

\twocolumn[
\icmltitle{Searching Large Neighborhoods for \\Integer Linear Programs with Contrastive Learning}



\icmlsetsymbol{equal}{*}

\begin{icmlauthorlist}
\icmlauthor{Taoan Huang}{USC}
\icmlauthor{Aaron Ferber}{USC}
\icmlauthor{Yuandong Tian}{FAIR}
\icmlauthor{Bistra Dilkina}{USC}
\icmlauthor{Benoit Steiner}{FAIR}
\end{icmlauthorlist}

\icmlaffiliation{USC}{ University of Southern California}
\icmlaffiliation{FAIR}{Meta AI, FAIR}

\icmlcorrespondingauthor{Taoan Huang}{taoanhua@usc.edu}

\icmlkeywords{Machine Learning, ICML}

\vskip 0.3in
]



\printAffiliationsAndNotice{}  

\begin{abstract}
Integer Linear Programs  (ILPs) are powerful tools for modeling and solving a large number of combinatorial optimization problems. Recently, it has been shown that Large Neighborhood Search (LNS), as a heuristic algorithm, can find high quality solutions to ILPs faster than Branch and Bound. However, how to find the right heuristics to maximize the performance of LNS remains an open problem. In this paper, we propose a novel approach, \CL, that delivers state-of-the-art anytime performance on several ILP benchmarks measured by  metrics including the primal gap, the primal integral, survival rates and the best performing rate. Specifically, 
\CL collects positive and negative solution samples from an expert heuristic that is slow to compute and learns a more efficient one with contrastive learning. We use graph attention networks and a richer set of features to further improve its performance. 
\end{abstract}

\section{Introduction}

Algorithm designs for combinatorial optimization problems (COPs) are important and challenging tasks. A wide variety of real-world problems are COPs, such as vehicle routing \cite{toth2002vehicle}, network design \cite{johnson1978complexity}, path planning \cite{pohl1970heuristic} and mechanism design \cite{de2003combinatorial} problems, and a majority of them are NP-hard to solve. In the past few decades, algorithms, including optimal algorithms, approximation algorithms and heuristic algorithms, have been studied extensively due to the importance of COPs. Those algorithms are mostly designed by human through costly processes that often require deep understanding of the problem domains and their underlying structures as well as considerable time and effort. 

Recently, there has been an increased interest to automate algorithm designs for COPs with machine learning (ML). Many  ML approaches learn to either construct  or improve solutions within an algorithmic framework, such as greedy search, local search or tree search, for a specific COP, such as the traveling salesman problem (TSP) \cite{xin2021neurolkh,zheng2021combining}, vehicle routing problem 
 (VRP) \cite{kool2018attention} or independent set problem \cite{li2018combinatorial}, and are often not easily applicable to other COPs. 

In contrast, Integer Linear Programs (ILPs) can flexibly encode and solve a broad family of COPs, such as minimum vertex cover, set covering and facility location problems. ILPs can be solved by Branch and Bound (BnB) \cite{land2010automatic}, an optimal tree search algorithm that can achieve state-of-the-art for ILPs. Over the past decades, BnB has been improved tremendously to become the core of many popular ILP solvers such as SCIP \cite{BestuzhevaEtal2021OO}, CPLEX \cite{cplex2009v12} and Gurobi \cite{gurobi}. However, due to its exhaustive search nature, it is hard for BnB to scale to large instances \cite{khalil2016learning,gasse2019exact}. 

On the other hand, Large Neighborhood Search (LNS) has recently been shown to  find high quality solutions much faster than BnB for large ILP instances \cite{song2020general,wu2021learning,sonnerat2021learning,huang2022local}. LNS starts from an initial solution (i.e., a feasible assignment of values to variables) and then improves the current best solution by iteratively picking a subset of variables to reoptimize while leaving others fixed. Picking which subset to reoptimize, i.e., the \emph{destroy heuristic}, is a critical component in LNS. Hand-crafted destroy heuristics, such as the randomized heuristic \cite{song2020general,sonnerat2021learning} and the Local Branching (LB) heuristic \cite{fischetti2003local}, are often either inefficient (slow to find good subsets) or ineffective (find subsets of bad quality). ML-based destroy heuristics have also been proposed and outperform hand-crafted ones. State-of-the-art approaches include \DM~\citep{sonnerat2021learning} that uses imitation learning (IL) to imitates the LB heuristic and \RL~\citep{wu2021learning} that uses a similar framework to \DM but trained with reinforcement learning (RL). 


In this paper, we propose a novel ML-based LNS for ILPs, namely \emph{\CL}, that uses contrastive learning (CL) \cite{chen2020simple,khosla2020supervised} to learn efficient and effective destroy heuristics. Similar to \DM~\cite{sonnerat2021learning}, we learn to imitate the \emph{Local Branching (LB)} heuristic, a destroy heuristic that selects the optimal subset of variables within the Hamming ball of the incumbent solutions. LB requires
solving another ILP with same size as the original problem and thus is computationally expensive. We not only use the optimal subsets provided by LB as the expert demonstration (as in \DM), but also leverage intermediate solutions and perturbations. When solving the ILP for LB, intermediate solutions are found and those that are close to optimal in term of effectiveness become \emph{positive samples}. 
We also collect \emph{negative samples} by randomly perturbing the optimal subset.
With both positive and negative samples, instead of a classification loss as in \DM, we use a contrastive loss that encourages the model to predict the subset similar to the positive samples but dissimilar to the negative ones with similarity measured by dot products \cite{oord2018representation,he2020momentum}. 
Finally, we also use a richer set of features and use graph attention networks (GAT) instead of GCN, to further boost performance. 

Empirically, we show that \CL outperforms state-of-the-art ML and non-ML approaches at different runtime cutoffs ranging from a few minutes to an hour in terms of multiple metrics, including the primal gap, the primal integral, the best performing rate and the survival rate, demonstrating the effectiveness and efficiency of \CL. In addition, \CL shows great generalization performance on test instances two times larger than training instances. 

\section{Background}
In this section, we first define ILPs and then introduce LNS for ILP solving and the Local Branching (LB) heuristic.

\subsection{ILPs}

An \textit{integer linear program (ILP)} is defined as 
\begin{equation}
    \min \bc^{\mathsf{T}}\bx \textrm{ s.t. } \bA \bx\leq \bb \textrm{ and } \bx \in \{0,1\}^n, \label{eqn::ILPformulation}
\end{equation}
where $\bx = (x_1,\ldots, x_n)^\sfT$ denotes the $n$ binary variables to be optimized, $\bc\in \mathbb{R}^n$ is the vector of objective coefficients, $\bA\in \mathbb{R}^{m\times n}$ and $\bb\in \mathbb{R}^{m}$ specify $m$ linear constraints.
A \textit{solution} to the ILP is a feasible assignment of values to the variables.
In this paper, we focus on the formulation above that consists of only binary variables, but our methods can be applied to mixed integer linear programs with continuous variables and/or non-binary integer variables.  

\subsection{LNS for ILP solving}\label{sec::LNSforILPs}

LNS is a heuristic algorithm that starts with an initial solution and then iteratively destroys and reoptimizes a part of the solution until a runtime limit is exceeded or some stopping condition is met. Let $\mathcal{I}=(\bA, \bb, \bc)$ be the input ILP, where $\bA, \bb$ and  $\bc$ are the coefficients defined in Equation (\ref{eqn::ILPformulation}), and $\bx^0$ be the initial solution (typically found by running BnB for a short runtime). In iteration $t\geq 0$ of LNS, given the \textit{incumbent solution} $\bx^t$,  defined as the best solution found so far, a \textit{destroy heuristic} selects a subset of $k^t$ variables $\calX^t= \{x_{i_1},\ldots, x_{i_{k^t}}\}$. The reoptimization is done by solving a sub-ILP with $\calX^t$ being the variables while fixing the values of $x_j\notin \calX^t$ the same as in $\bx^t$. The solution to the sub-ILP is the new incumbent solution $\bx^{t+1}$ and then LNS proceeds to iteration $t+1$.
Compared to BnB, LNS is more effective in improving the objective value $\bc^{\mathsf{T}}x$ especially on difficult instances \cite{song2020general,sonnerat2021learning,wu2021learning}. Compared to other local search methods, LNS explores a large neighborhood in each step and thus, is more effective in avoiding local minima. 

 \paragraph{Adaptive Neighborhood Size}
Adaptive methods are commonly used to set the neighborhood size $k^t$ in previous work \cite{sonnerat2021learning,huang2022local}. The initial neighborhood size $k^0$ is set to a constant or a fraction of the number of variables. In this paper, we consider the following adaptive method \cite{huang2022local}: in iteration $t$, if LNS finds an improved solution, we let $k^{t+1}=k^t$, otherwise $k^{t+1} =\min\{ \gamma\cdot k^t, \beta\cdot n\}$ where  $\gamma>1$ is a constant and we upper bound  $k^t$ to a constant fraction $\beta<1$ of the number of variables to make sure the sub-ILP is not too large (thus, too difficult) to solve. Adaptively setting $k^t$ helps LNS escape local minima by expanding the search neighborhood when it fails to improve the solution.

\subsection{LB Heuristic}

The LB Heuristic \cite{fischetti2003local} is originally proposed as a primal heuristic in BnB but also applicable in LNS for ILP solving \cite{sonnerat2021learning,liu2022learning}. 
Given the incumbent solution $\bx^t$ in iteration $t$ of LNS, LB aims to find the subset of variables to destroy $\calX^t$ such that it leads to the optimal $\bx^{t+1}$ that differs from $\bx^t$ on at most $k^t$ variables, i.e., it computes the optimal solution $\bx^{t+1}$ that sits within a given Hamming ball of radius $k^t$ centered around $\bx^t$.
To find $\bx^{t+1}$, the LB heuristic solves the LB ILP that is exactly the same ILP from input but with one additional constraint that limits the distance between $\bx^t$ and $\bx^{t+1}$:
$\sum_{i\in[n]:x^t_i=0}x^{t+1}_i + \sum_{i\in[n]:x^t_i=1}(1-x^{t+1}_i)\leq k^t. $
The LB ILP is of the same size of the input ILP (i.e., it has the same number of variables and one more constraint), therefore, it is often too slow to be useful in practice.

\section{Related Work}

In this section, we summarize related work on LNS for ILPs and other COPs, learning to solve ILPs with BnB and contrastive learning for COPs. We also summarize additional related work on LNS-based primal heuristics for BnB and learning to solve other COPs in Appendix.

\subsection{LNS for ILPs and Other COPs}
Huge effort has been made to improve BnB for ILPs in the past decades, but LNS for ILPs has not been studied extensively.  
Recently, \citet{song2020general} show that even a randomized destroy heuristic in LNS can outperform state-of-the-art BnB. They also show that an ML-guided decomposition-based LNS can achieve even better performance, where they apply RL and IL to learn destroy heuristics that decompose the set of variables into equally-sized subsets using a classification loss.  
\citet{sonnerat2021learning} learn to select variables by imitating LB. \RL~\citep{wu2021learning} uses a similar framework but trained with RL and outperforms \citet{song2020general}. 
Both \citet{wu2021learning} and \citet{sonnerat2021learning} use the bipartite graph representations of ILPs to learn the destroy heuristics represented by GCNs.
Another line of related work focuses on improving LB. \citet{liu2022learning} use ML to tune the runtime limit and neighborhood sizes for LB. \citet{huang2022local} propose \LBRELAX to select variables by solving  the LP relaxation of LB.

Besides ILPs, LNS has been applied to solve many COPs, such as VRP \cite{ropke2006adaptive,azi2014adaptive}, TSP \cite{smith2017glns}, scheduling \cite{kovacs2012adaptive,vzulj2018hybrid} and path planning problems \cite{LiAAAI22,LiIJCAI21}. ML methods have also been applied to improve LNS for those applications \cite{chen2019learning,lu2019learning,hottung2020neural,li2021learning,huang2022anytime}. 

\subsection{Learning to Solve ILPs with BnB}

Several studies have applied ML to improve BnB. The majority of works focus on learning to either  select variables to branch on \cite{khalil2016learning,gasse2019exact,gupta2020hybrid,zarpellon2021parameterizing} or select nodes to expand \cite{he2014learning,labassi2022learning}. There are also works on learning to schedule and run primal heuristics \cite{khalil2017learning,chmiela2021learning} and to select cutting planes \cite{tang2020reinforcement,paulus2022learning,huang2022learning}. 

\subsection{Contrastive Learning for COPs}

While contrastive learning of visual representations \cite{hjelm2018learning,he2020momentum,chen2020simple} and graph representations \cite{you2020graph,tong2021directed} 
have been studied extensively, it has not been explored much for COPs. \citet{mulamba2021contrastive} derive a contrastive loss for decision-focused learning to solve COPs with uncertain inputs that can be learned from historical data, where they view non-optimal solutions as negative samples. \citet{duan2022augment} use contrastive pre-training to learn good representations for the boolean satisfiability problem. 

\section{Contrastive Learning for LNS}

Our goal is to learn a policy, a destroy heuristic represented by an ML model, that selects a subset of variables to destroy and reoptimize in each LNS iteration. Specifically, let $\bs^t=(\calI, \bx^t)$ be the current state in iteration $t$ of LNS where $\mathcal{I}=(\bA, \bb, \bc)$ is  the ILP and $\bx^t$ is the incumbent solution, the policy predicts an action $\ba^t=(a_1^t,\ldots, a_n^t)\in \{0,1\}^n$, a binary representation of the selected variables $\calX^t$ 
indicating whether $x_i$ is selected ($a^t_i=1$) or not ($a^t_i=0$). We use contrastive learning to learn to predict high quality $\ba^t$ such that, after solving the sub-ILP derived from $\ba^t$ (or $\calX^t$), the resulting incumbent solution $\bx^{t+1}$ is improved as much as possible.
Next, we describe how we prepare data for contrastive learning, the policy network and the contrastive loss used in training, and finally introduce how the learned policy is used in \CL. 

\subsection{Data Collection}

Following previous work by \citet{sonnerat2021learning}, we use LB as the expert policy to collect good demonstrations to learn to imitate. Formally, for a given state $\bs^t=(\calI,\bx^t)$, we use LB to find the optimal action $\ba^t$ that leads to the minimum $\bc^{\mathsf{T}}\bx^{t+1}$ after solving the sub-ILP. 
Different from the previous work, we use contrastive learning to learn to make discriminative predictions of  $\ba^t$ by contrasting positive and negative samples (i.e., good and bad examples of actions $\ba^t$). In the following, we describe how we collect the positive sample set $\calS_{\sfp}^t$ and the negative sample set $\calS_{\sfn}^t$.

\paragraph{Collecting Positive Samples $\calS_{\sfp}^t$}

During data collection, given $\bs^t=(\calI, \bx^t)$, we solve the LB ILP with the incumbent solution $\bx^t$ and neighborhood size $k^t$ to find the optimal $\bx^{t+1}$. 
LNS proceeds to  iteration $t+1$ with $\bx^{t+1}$ until no improving solution $\bx^{t+1}$ could be found by the LB ILP within a runtime limit. 
In experiments, the LB ILP is solved with SCIP 8.0.1 \cite{BestuzhevaEtal2021OO} with an hour runtime limit  and $k^t$ is fine-tuned for each type of instances. After each solve of the LB ILP, in addition to the best solution found, SCIP records all intermediate solutions found during the solve. We look for intermediate solutions $\bx'$ whose resulting improvements on the objective value is at least $0<\alpha_{\sfp}\leq 1$ times the best improvement  (i.e., $\bc^{\mathsf{T}}(\bx^t-\bx')\geq \alpha_{\sfp} \cdot \bc^{\mathsf{T}}(\bx^t-\bx^{t+1})$) and consider their corresponding actions as positive samples. We limit the number of the positive samples $|\calS^t_{\sfp}|$ to  $u_{\sfp}$.
If more than $u_{\sfp}$ positive samples are available, we record the top $u_{\sfp}$ ones to avoid large computational overhead with too many samples when computing the contrastive loss (see Section \ref{sec::trainingwithCL}).
$\alpha_{\sfp}$ and $u_{\sfp}$ are set to $0.5$ and $10$, respectively, in experiments. 

\paragraph{Collecting Negative Samples $\calS_{\sfn}^t$}

Negative samples are critical parts of contrastive learning to help distinguish between good and bad demonstrations. We collect a set of $c_{\sfn}^t$ negative samples $\calS_{\sfn}^t$, where $c_{\sfn}^t=\kappa |\calS_{\sfp}^t|$ and $\kappa$ is a hyperparameter to control the ratio between the numbers of positive and negative samples.
Suppose $\calX^t$ is the optimal set of variables selected by LB. We then perturb $\calX^t$ to get $\hat{\calX}^{t}$ by replacing $5\%$ of the variables in  $\calX^t$ with the same number of those not in $\calX^t$ uniformly at random. We then solve the corresponding sub-ILP derived from $\hat{\calX}^{t}$ to get a new incumbent solution $\hat{\bx}^{t+1}$. If the resulting improvement of $\hat{\bx}^{t+1}$ is less than $0\leq \alpha_{\sfn}<1$ times the best improvement  (i.e., $\bc^{\mathsf{T}}(\bx^t-\hat{\bx}^{t+1})\leq \alpha_{\sfn} \cdot \bc^{\mathsf{T}}(\bx^t-\bx^{t+1})$), we consider its corresponding action as a negative sample. We repeat this $c_{\sfn}^t$ times to collect negative samples. If less than $c_{\sfn}^t$ negative samples is collected, we increase the perturbation rate from $5\%$ to $10\%$ and generate another $c_{\sfn}^t$ samples. We keep increasing the perturbation rate at an increment of $5\%$ until $c_{\sfn}^t$ negative samples are found or it reaches $100\%$. In experiments, we set $\kappa = 9$ and $\alpha_{\sfn}= 0.05$, and it takes less than 3 minutes to collect negative samples for each state. 

\subsection{Policy Network}
\label{sec::policyNetwork}
Following previous work on learning for ILPs \cite{gasse2019exact, sonnerat2021learning, wu2021learning}, we use a bipartite graph representation of ILP to encode a state $\bs^{t}$. The bipartite graph consists of $n+m$ nodes representing the $n$ variables and $m$ constraints on two sides, respectively, with an edge connecting a variable and a constraint if the variable has a non-zero coefficient in the constraint. Following \citet{sonnerat2021learning}, we use features proposed in \citet{gasse2019exact} for node features and edge features in the bipartite graph and also include a fixed-size window of most recent incumbent values as variable node features with the window size set to 3 in experiments. In addition to features used in \citet{sonnerat2021learning}, we include features proposed in \citet{khalil2016learning} computed at the root node of BnB to make it a richer set of variable node features.

We learn a policy $\pi_{\btheta}(\cdot)$ represented by a graph attention network (GAT) \cite{brody2021attentive} parameterized by learnable weights $\btheta$. The policy takes as input the state $\bs^t$ and output a score vector $\pi_{\btheta}(\bs^t)\in [0,1]^n$, one score per variable.
To increase the modeling capacity and to manipulate node interactions proposed by our architecture, we use embedding layers to map each node feature and edge feature to space $\mathbb{R}^d$. 
Let $\bfv_j, \bfc_i, \bfe_{i,j}\in\mathbb{R}^d$ be the embeddings of the $i$-th variable, $j$-th constraint and the edge connecting them output by the embedding layers. Since our graph is bipartite, following previous work \cite{gasse2019exact}, we perform two rounds of message passing through the GAT.  In the first round, each constraint node $\bfc_i$ attends to its neighbors $\calN_i$ using an attention structure with $H$ attention heads
to get updated constraint embeddings $\bfc'_i$ (computed as a function of $\bfv_j, \bfc_i, \bfe_{i,j}$).
In the second round, similarly, each variable node attends to its neighbors to get  updated variable embeddings $\bfv'$ (computed as a function of $\bfv_j, \bfc'_i, \bfe_{i,j}$) with another set of attention weights. 
After the two rounds of message passing, the final representations of variables $\bfv'$ are passed through a multi-layer perceptron (MLP) to obtain a scalar value for each variable and, finally, we apply the sigmoid  function to get a score  between 0 and 1. 
Full details of the network architecture are provided in Appendix. 
In experiments, $d$ and $H$ are set to $64$ and $8$, respectively.

\begin{table*}[htbp]
\centering
\caption{ Names and the average numbers of variables and constraints of the test instances. 
\label{table::instanceSize}}
\scriptsize
\begin{tabular}{c|crrr|crrr}
\hline
              & \multicolumn{4}{c|}{Small Instances}                                                                              & \multicolumn{4}{c}{Large Instances}                                                                                \\ \hline
Name          & \multicolumn{1}{c}{MVC-S}  & \multicolumn{1}{c}{MIS-S} & \multicolumn{1}{c}{CA-S}  & \multicolumn{1}{c|}{SC-S} & \multicolumn{1}{c}{MVC-L}   & \multicolumn{1}{c}{MIS-L}  & \multicolumn{1}{c}{CA-L}  & \multicolumn{1}{c}{SC-L} \\ \hline
\#Variables   & \multicolumn{1}{r}{1,000}  & \multicolumn{1}{r}{6,000} & \multicolumn{1}{r}{4,000} & 4,000                     & \multicolumn{1}{r}{2,000}   & \multicolumn{1}{r}{12,000} & \multicolumn{1}{r}{8,000} & 8,000                    \\ 
\#Constraints & \multicolumn{1}{r}{65,100} & \multicolumn{1}{r}{23,977}      & \multicolumn{1}{r}{2,675}      & 5,000                     & \multicolumn{1}{r}{135,100} & \multicolumn{1}{r}{48,027}       & \multicolumn{1}{r}{5,353}      & 5,000                    \\ \hline
\end{tabular}
\end{table*}

\subsection{Training with a Contrastive Loss}\label{sec::trainingwithCL}

Given a set of ILP instance for training, we follow the expert's trajectory to collect training data. Let $\calD = \{(\bs, \calS_{\sfp}, \calS_{\sfn})\}$ be the set of states with their corresponding sets of positive and negative samples in the training data. 
A contrastive loss is a function whose value is low when the predicted action $\pi_{\btheta}(\bs)$ is similar to the positive samples $\calS_{\sfp}$ and dissimilar to the negative samples $\calS_{\sfn}$. With similarity measured by dot products, a form of supervised contrastive loss, called InfoNCE \cite{oord2018representation,he2020momentum}, 
is used in this paper:
$$\small
\calL(\btheta) = \sum_{(\bs, \calS_{\sfp}, \calS_{\sfn})\in \calD}\frac{-1}{|\calS_{\sfp}|}\sum_{\ba\in \calS_{\sfp}}\log\frac{\exp(\ba^{\mathsf{T}}\pi_{\btheta}(\bs)/\tau)}{\sum_{\ba'\in \calS_{\sfn}\cup{\{\ba\}}}\exp(\ba'^{\mathsf{T}}\pi_{\btheta}(\bs)/\tau)}
$$
where  $\tau$ is a temperature hyperparameter set to 0.07 \cite{he2020momentum} in experiments.

\begin{figure*}[tbp]
    \centering
    \includegraphics[width=0.8\textwidth]{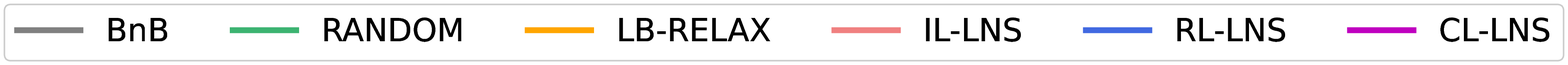}

    \subfloat[MVC-S (left) and MVC-L (right).]
    {
        \includegraphics[width=0.24\textwidth]{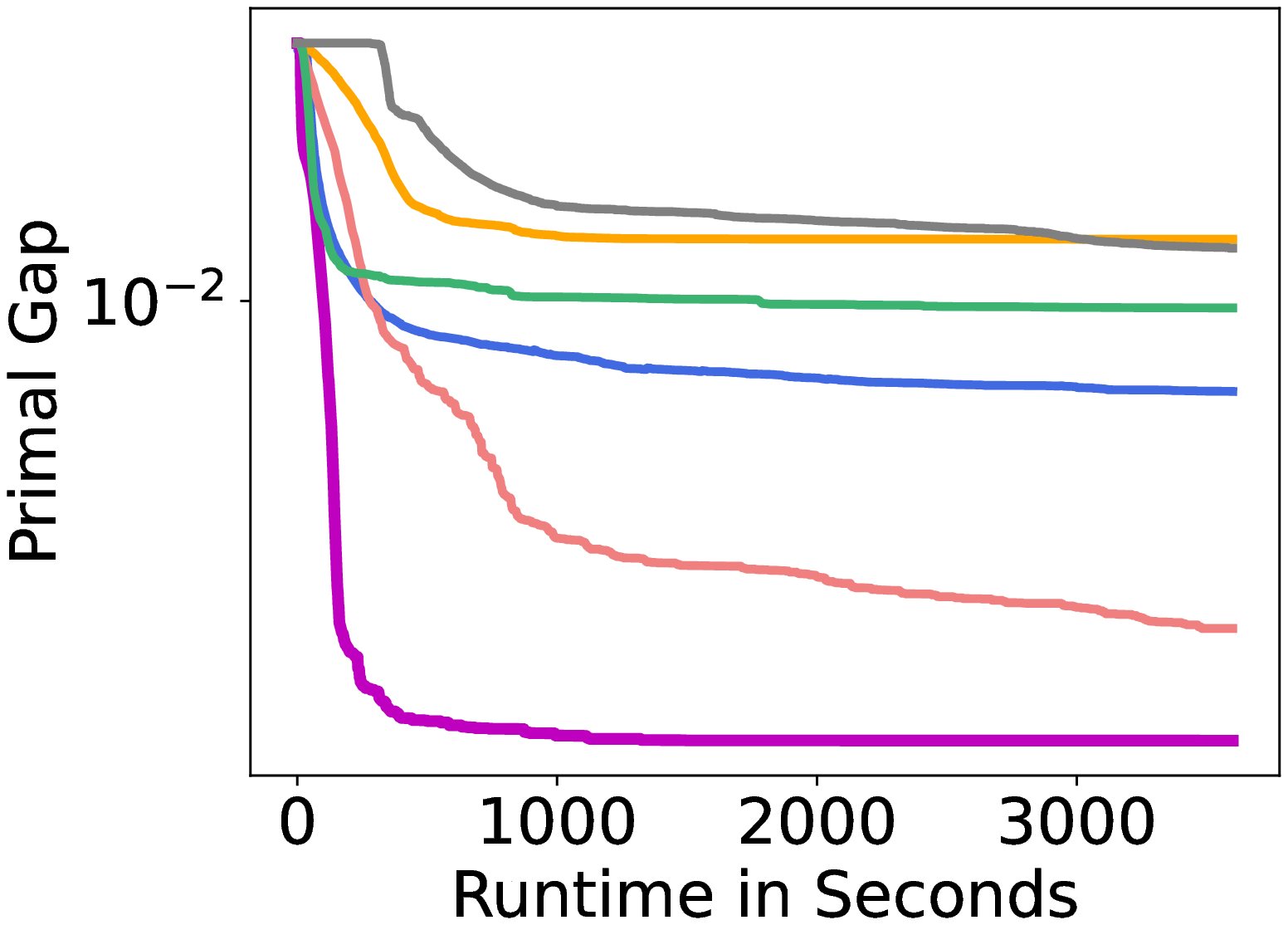}
        \includegraphics[width=0.24\textwidth]{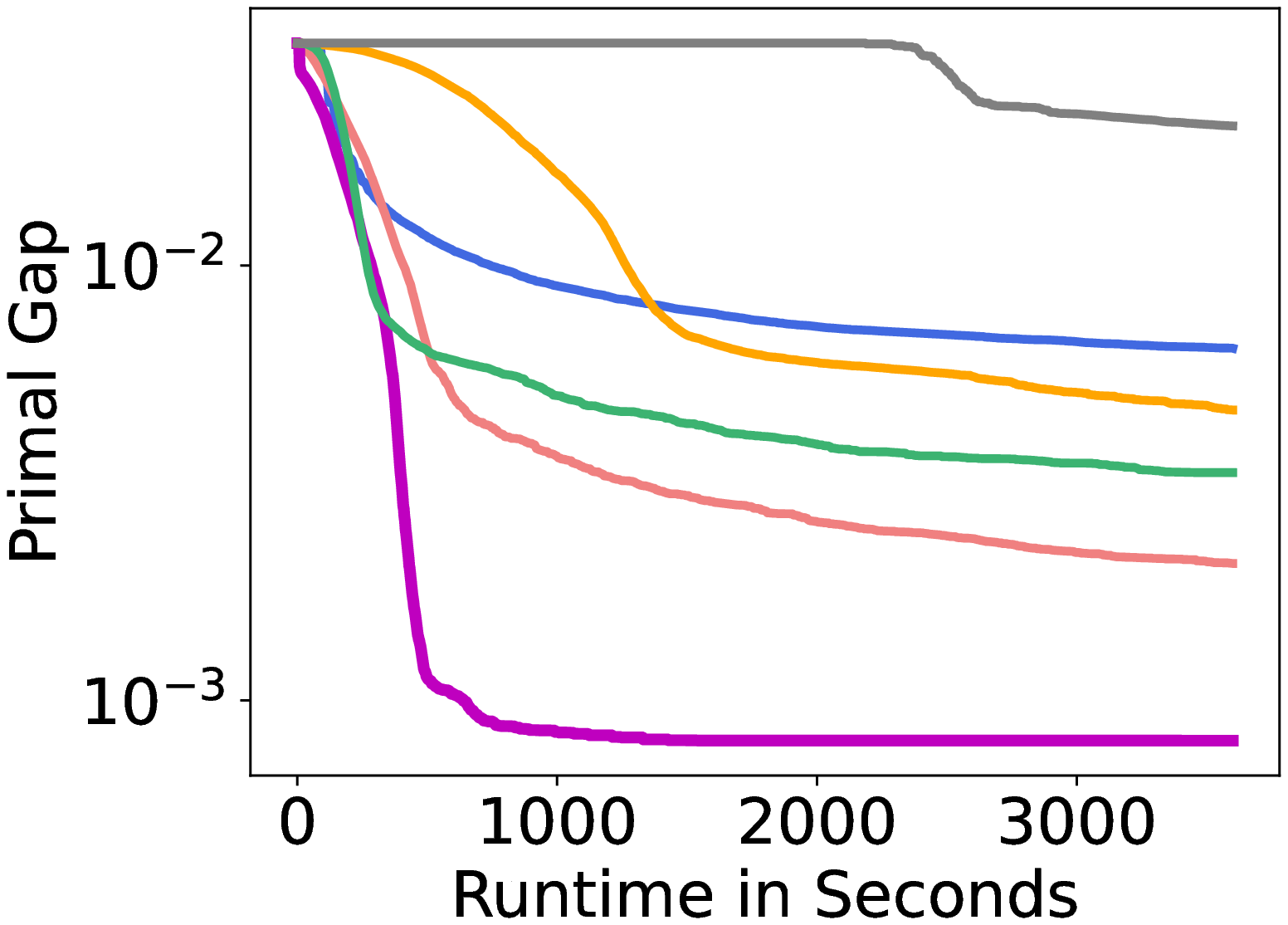}    
    }
    \subfloat[MIS-S (left) and MIS-L (right).]
    {
        \includegraphics[width=0.24\textwidth]{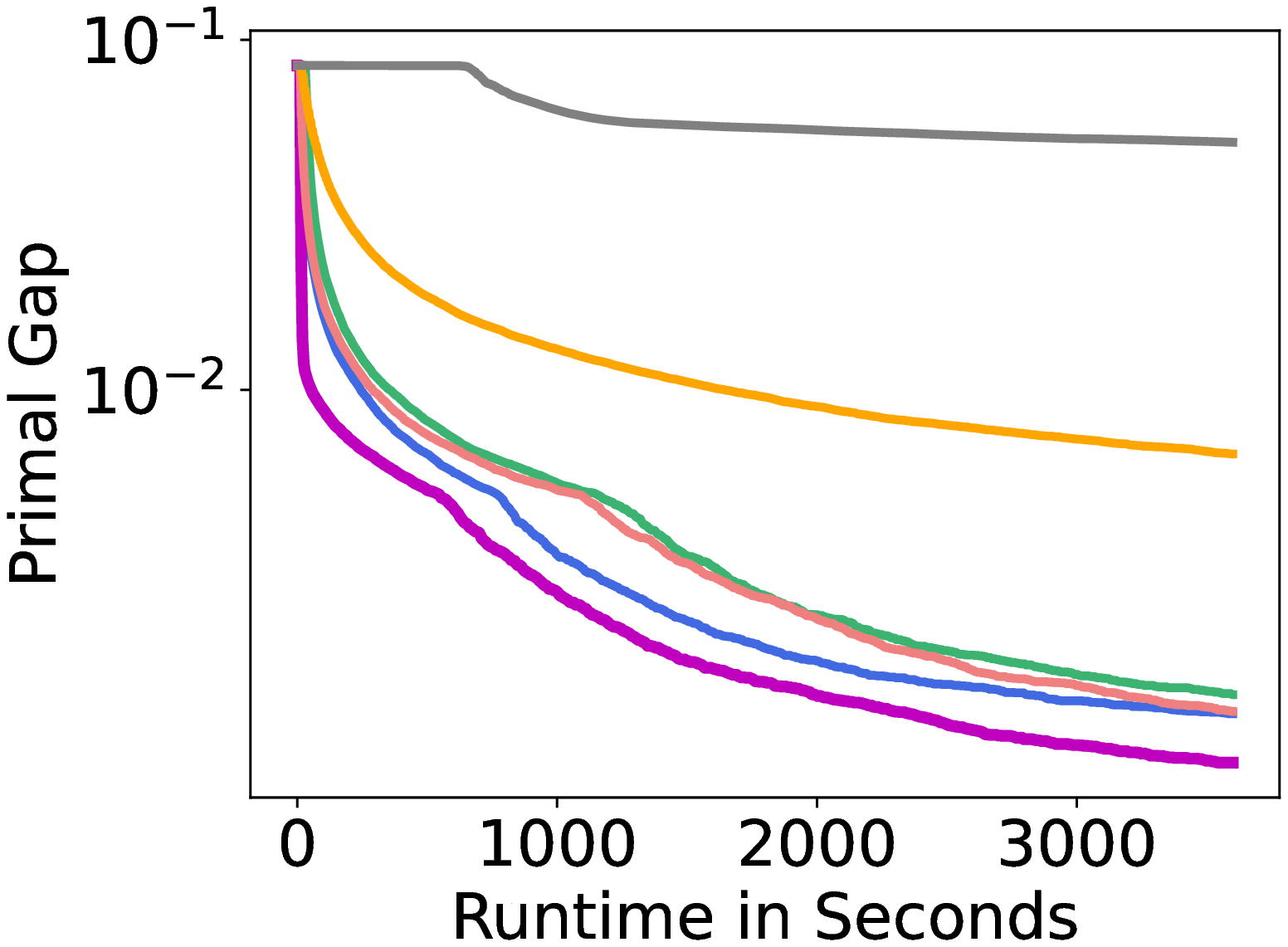}
        \includegraphics[width=0.24\textwidth]{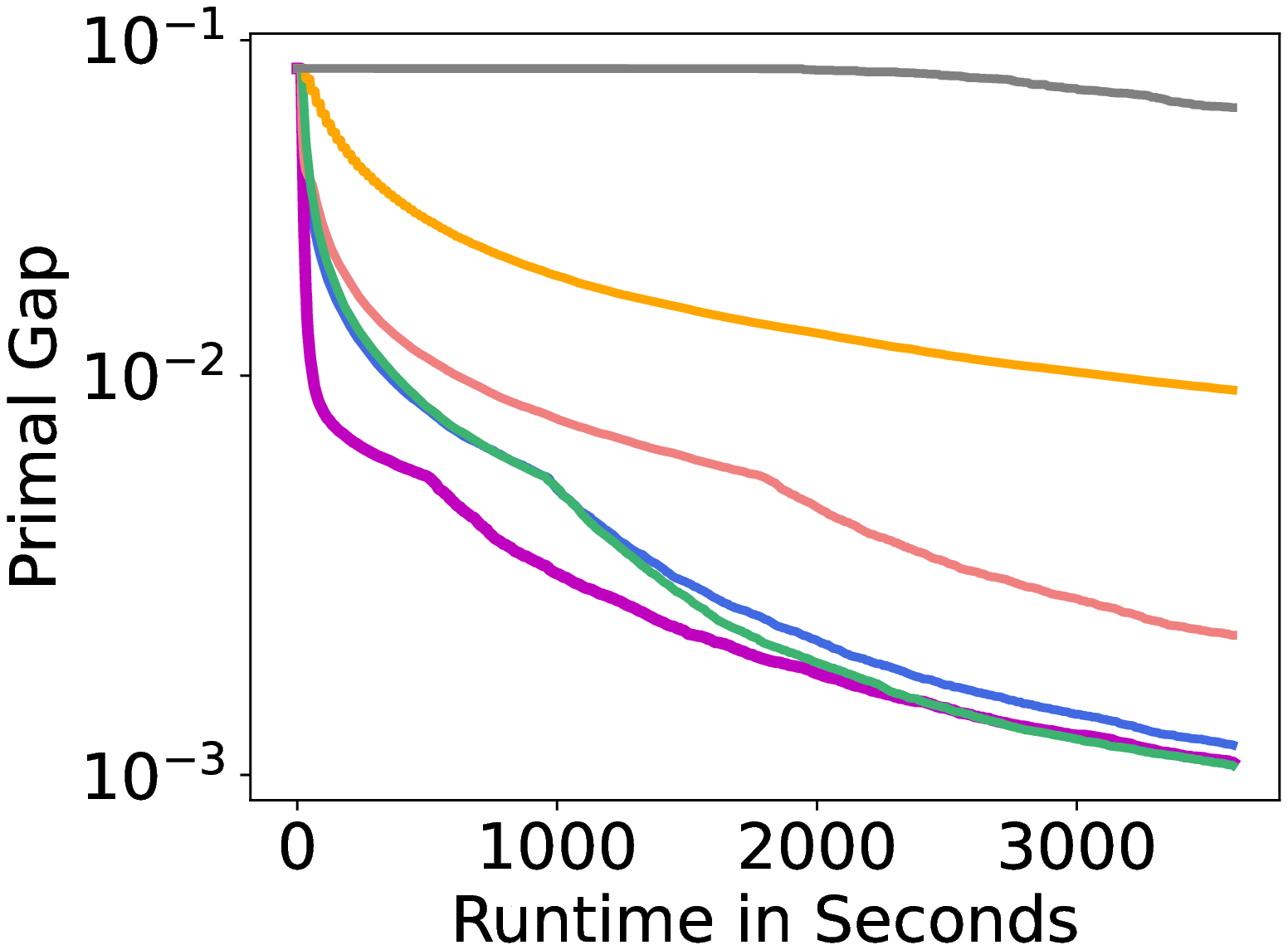}    
    }\\
    \subfloat[CA-S (left) and CA-L (right).]
    {
        \includegraphics[width=0.24\textwidth]{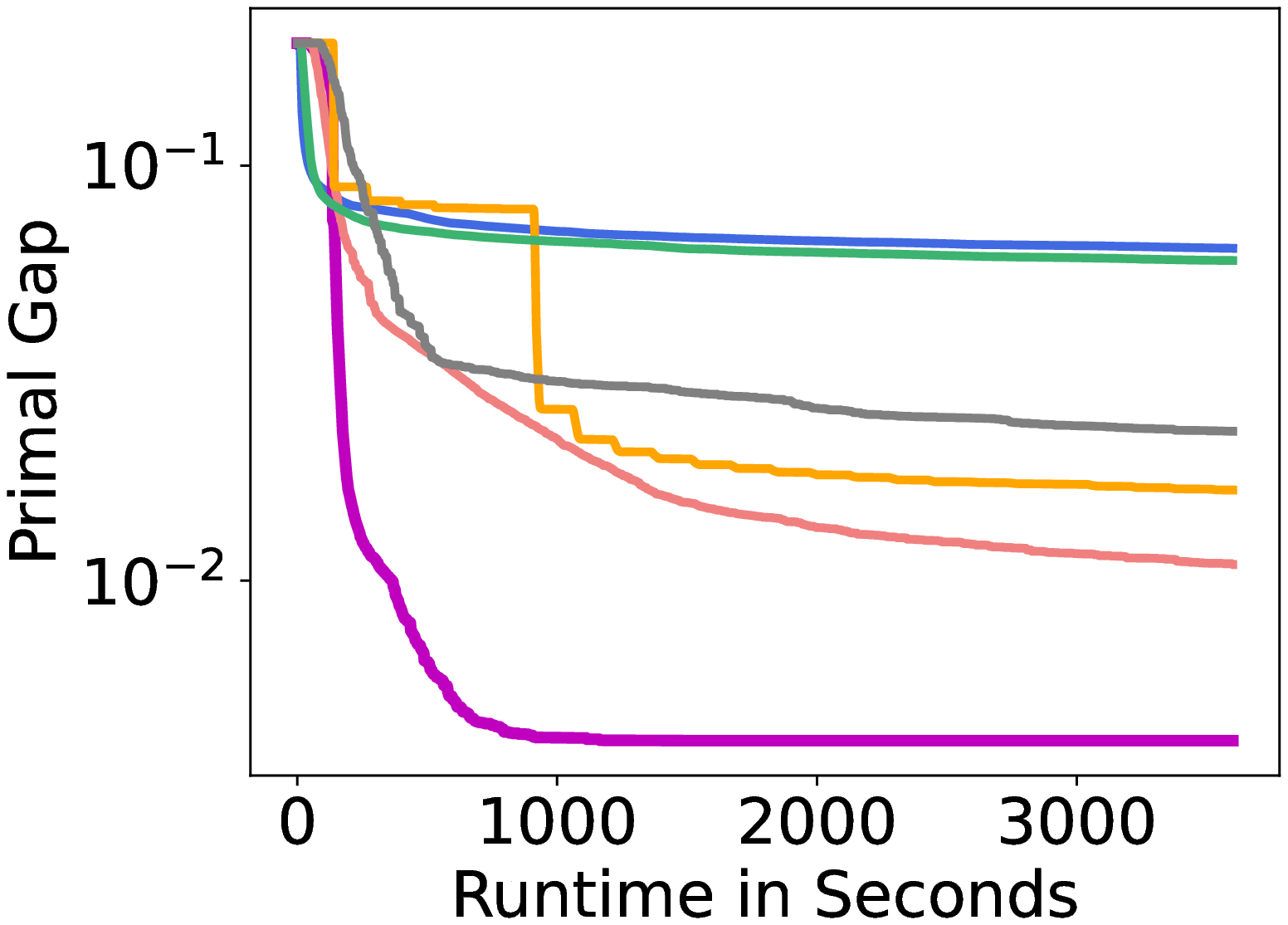}
        \includegraphics[width=0.24\textwidth]{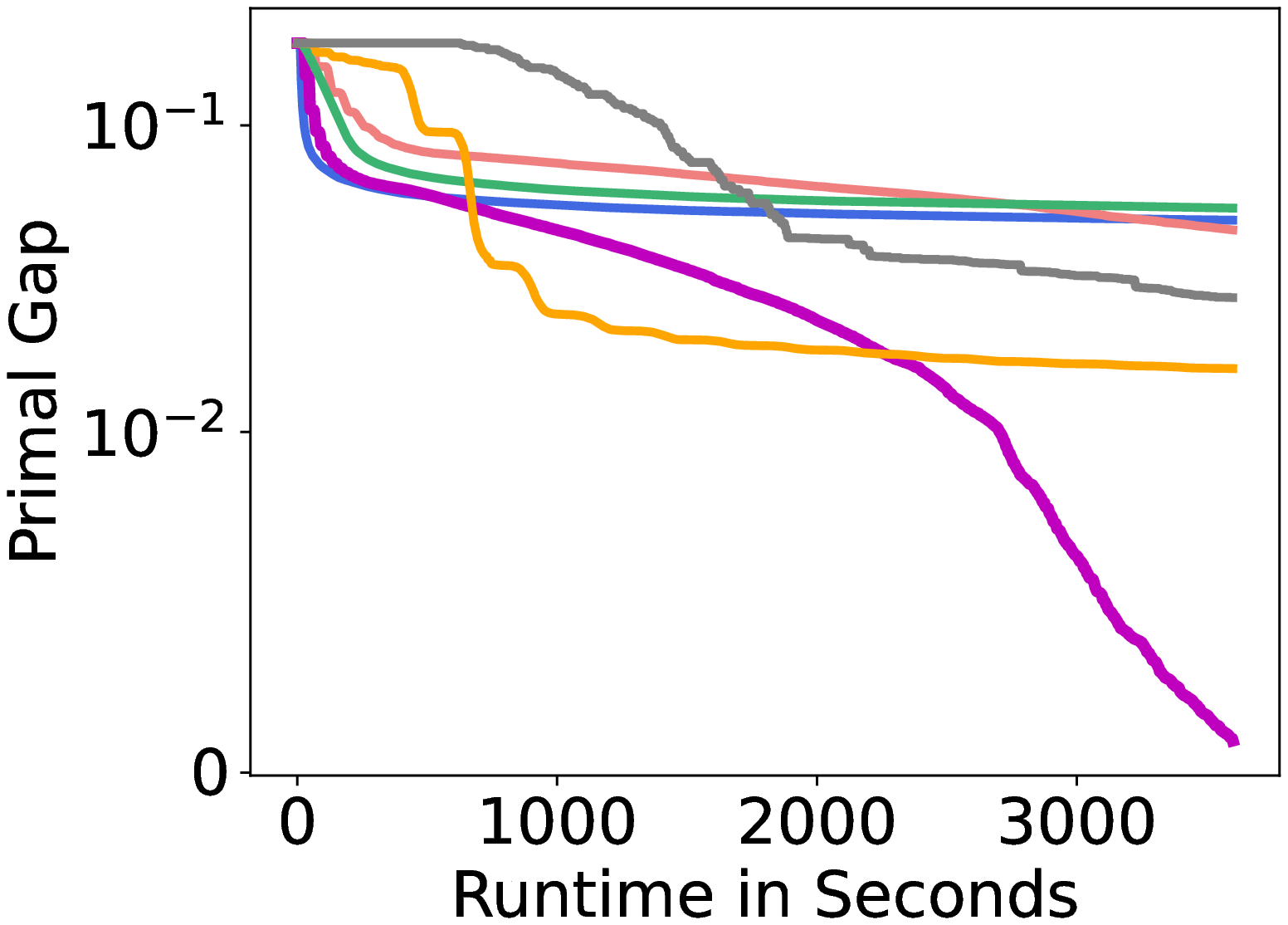}    
    }
    \subfloat[SC-S (left) and SC-L (right).]
    {
        \includegraphics[width=0.24\textwidth]{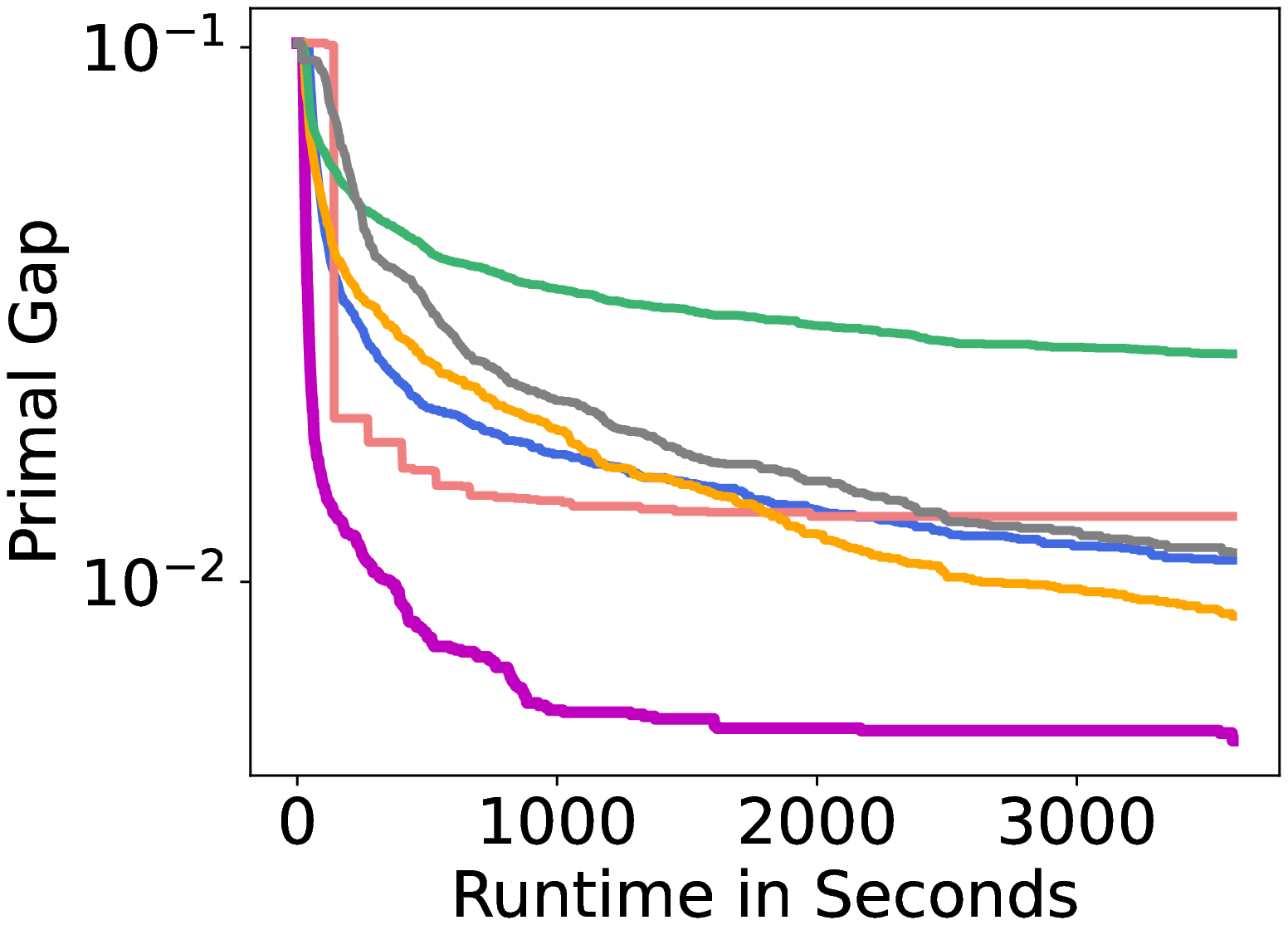}
        \includegraphics[width=0.24\textwidth]{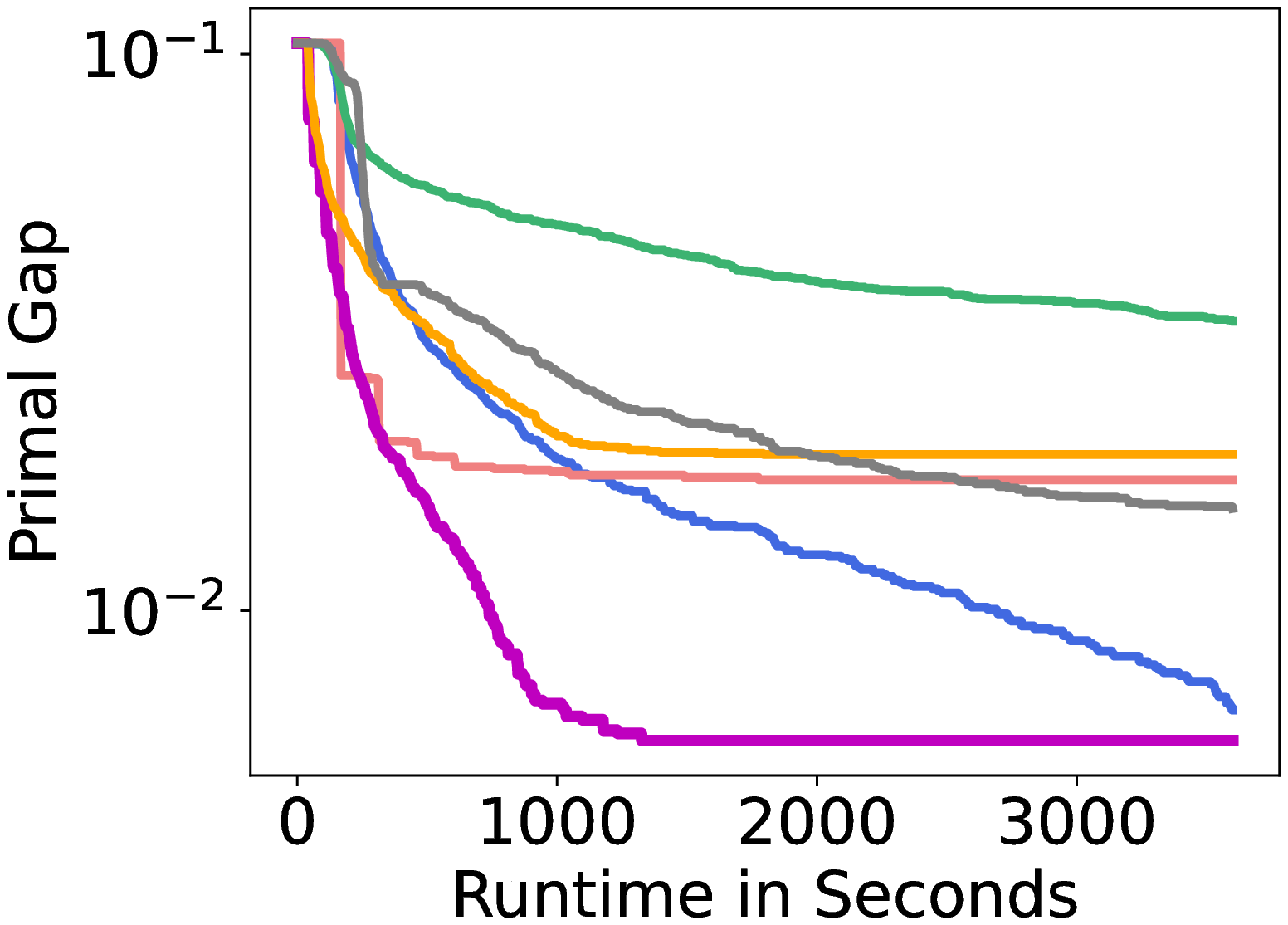}    
    }
    \caption{The primal gap (the lower the better) as a function of runtime, averaged over 100 test instances. For ML approaches, the policies are trained on only small training instances but tested on both small and large test instances. \label{res::gap}}
\end{figure*}

\subsection{Applying Learned Policy $\pi_{\btheta}$} \label{sec::usingthepolicy}
We apply the learned policy $\pi_{\btheta}$ in LNS. 
In iteration $t$, let $(v_1,\cdots,v_n):=\pi_{\btheta}(\bs^t)$ be the variable scores output by the policy. To select $k^t$ variables, \CL greedily selects those with the highest scores. Previous works \cite{sonnerat2021learning,wu2021learning} commonly use sampling methods to select the variables, but those sampling methods are empirically worse than our greedy method in \CL. However, when the adaptive neighborhood size $k^t$ reaches its upper bound $\beta\cdot n$, \CL may repeat the same prediction due to deterministic selection process. When this happens, we switch to the sampling method introduced in \cite{sonnerat2021learning}. The sampling method selects variables sequentially: at each step, a variable $x_i$ that has not been selected yet is selected with probability proportional to $v_i^{\eta}$, where $\eta$ is a temperature parameter set to $0.5$ in experiments.

\section{Empirical Evaluation}

In this section, we introduce our evaluation setup and then present the results. Our code will be made available to the public upon publication. 

\subsection{Setup}

\paragraph{Instance Generation} We evaluate on four NP-hard problem benchmarks that are widely used in existing studies \cite{wu2021learning,song2020general,scavuzzo2022learning}, which consist of two graph optimization problems, namely the minimum vertex cover (MVC) and maximum independent set (MIS) problems, and two non-graph optimization problems, namely the combinatorial auction (CA) and set covering (SC) problems.  
We first generate a test set of 100 \emph{small instances} for each problem, namely MVC-S, MIS-S, CA-S and SC-S. 
MVC-S instances are generated according to the Barabasi-Albert random graph model \cite{albert2002statistical}, with 1,000 nodes and average degree 70 following \cite{song2020general}.
MIS-S instances are generated according to the Erdos-Renyi random graph model \cite{erdos1960evolution}, with 6,000 nodes and average degree 5 following \cite{song2020general}.
CA-S instances are generated with 2,000 items and 4,000 bids according to the arbitrary relations in \citet{leyton2000towards}.
SC-S instances are generated with 4,000 variables and 5,000 constraints following \citet{wu2021learning}. 
We then generate another test set of 100 \emph{large instances} for each problem by doubling the number of variables, namely MVC-L, MIS-L, CA-L and SC-L.
For each test set,  Table \ref{table::instanceSize} shows its average numbers of variables and constraints. More details of instance generation are included in Appendix.

 For data collection and training, we generate another set of 1,024 small instances for each problem. We split these instances into training and validation sets, each consisting of 896 and 128 instances, respectively.

\paragraph{Baselines}

We compare \CL with five baselines:
    (1) BnB: using SCIP (v8.0.1), the state-of-the-art open-source ILP solver, with the aggressive mode fine-tuned to focus on improving the objective value;
    (2) \RANDOM: LNS which selects the neighborhood by uniformly sampling $k^t$ variables without replacement;
    (3) \LBRELAX \cite{huang2022local}: LNS which selects the neighborhood with the \LBRELAX heuristics;
    (4) \DM \cite{sonnerat2021learning};
    (5) \RL \cite{wu2021learning}. We compare with two more baselines in Appendix. 
For each ML approach,  a separate model is trained for each problem on the small training set and tested on both small and large test sets. We implement \DM  and fine tune its hyperparameters for each problem since the authors do not fully open source the code. \DM uses the same training dataset as \CL but uses only the positive samples. For \RL, we use the code and hyperparameters provided by the authors and train the models with five random seeds to select one with the best performance on the validation sets. We do not compare to the approach by \citet{song2020general} since it performs worse than \RL on multiple problems \cite{wu2021learning}.  

\paragraph{Metrics}
We use the following metrics to evaluate all approaches:
(1) The \textit{primal bound} is the objective value of the ILP;
(2) The \textit{primal gap} \cite{berthold2006primal} is the normalized difference between the primal bound $v$ and a precomputed best known objective value $v^*$, defined as $\frac{|v-v^*|}{\max(v,v^*,\epsilon)}$ if $v$ exists and $v\cdot v^*\geq 0$, or 1 otherwise. We use $\epsilon=10^{-8}$ to avoid division by zero; 
(3) The \textit{primal integral} \cite{achterberg2012rounding} at time $q$  is the integral on $[0,q]$ of the primal gap as a function of runtime. It captures the quality of and the speed at which solutions are found;
(4) The \textit{survival rate}  to meet a certain primal gap threshold is the fraction of instances with primal gaps below the threshold \cite{sonnerat2021learning}; (5) The \emph{best performing rate}  of an approach is  the fraction of instances on which it achieves the best primal gap (including ties) compared to all approaches at a given runtime cutoff.
Since BnB and LNS are both anytime algorithms, we  show these metrics as a function of runtime or the number of iterations in LNS (when applicable) to demonstrate their anytime performance.

\begin{table}[bp]
\scriptsize
\centering
\caption{
Primal gap (PG) (in percent), primal integral (PI) at 60 minutes runtime cutoff, averaged over 100 test instances and their standard deviations. ``$\downarrow$'' means the lower the better. For ML approaches, the policies are trained on only small training instances but tested on both small and large test instances. \label{res::smalltable60}}

\begin{tabular}{c|rr|rr}
\hline
                                   & \multicolumn{1}{c}{PG (\%) $\downarrow$}         & \multicolumn{1}{c|}{PI $\downarrow$}               & \multicolumn{1}{c}{PG (\%) $\downarrow$}         & \multicolumn{1}{c}{PI $\downarrow$} \\ \hline

                       & \multicolumn{2}{c|}{MVC-S}                                                                        & \multicolumn{2}{c}{MIS-S}                                                                         \\ \hline
BnB                     & {1.32$\pm$0.43} & 66.1$\pm$13.1            & {5.10$\pm$0.69} & 222.8$\pm$25.9           \\ 
\RANDOM                & {0.96$\pm$1.26} & 38.0$\pm$44.8             & {0.24$\pm$0.14} & 22.1$\pm$5.0             \\ 
\LBRELAX                 & {1.38$\pm$1.51} & 57.0$\pm$51.2             & {0.65$\pm$0.20} & 46.9$\pm$6.5             \\ 
\DM        & {0.29$\pm$0.23} & {19.2$\pm$10.2}               & {0.22$\pm$0.17} & { 19.4$\pm$5.8}             \\ 
\RL               & {0.61$\pm$0.34} & 29.6$\pm$11.5             & {0.22$\pm$0.14} & 17.2$\pm$5.2            \\ 
{\bf\CL}                & {\bf 0.17$\pm$0.09} & {\bf 8.7$\pm$6.7}             & {\bf 0.15$\pm$0.15} & {\bf12.8$\pm$5.4}            \\ \hline
\multicolumn{1}{l|}{} & \multicolumn{2}{c|}{CA-S}                                                                         & \multicolumn{2}{c}{SC-S}                                                                          \\ \hline
BnB                & 2.28$\pm$0.59 & 137.4$\pm$25.9      & 1.13$\pm$0.95 & 86.7$\pm$37.9                         \\ 
\RANDOM             & 5.90$\pm$1.02 & 235.6$\pm$34.9      & 2.67$\pm$1.29 & 124.3$\pm$45.4                    \\ 
\LBRELAX       & 1.65$\pm$0.57 & 140.5$\pm$18.3      & 0.86$\pm$0.83 & 63.2$\pm$31.6    \\ 
\DM             &1.09$\pm$0.51 & 90.0$\pm$20.8   & 1.33$\pm$0.97 & 63.2$\pm$34.3            \\ 
\RL            & 6.32$\pm$1.03 & 249.2$\pm$35.9   & 1.10$\pm$0.77 & 77.8$\pm$28.9            \\ 
{\bf\CL} & {\bf 0.65$\pm$0.32} & {\bf50.7$\pm$22.7} & {\bf0.50$\pm$0.58} & {\bf26.2$\pm$12.8}\\ \hline
\hline    

                       & \multicolumn{2}{c|}{MVC-L}                                                                        & \multicolumn{2}{c}{MIS-L}                                                                         \\ \hline
BnB                    & {2.41$\pm$0.40} & 130.2$\pm$11.1            & {6.29$\pm$1.62} & 285.1$\pm$18.2           \\ 
\RANDOM                 & {0.38$\pm$0.24} & 22.7$\pm$8.0              & {\bf0.11$\pm$0.08} & 19.0$\pm$3.1             \\ 
\LBRELAX                 & {0.46$\pm$0.23} & 48.4$\pm$7.5             & {0.91$\pm$0.16} & 68.6$\pm$5.5             \\ 
\DM               & {0.27$\pm$0.23} & {21.2$\pm$8.1}               & {0.29$\pm$0.15} & { 27.1$\pm$5.5}             \\ 
\RL               & {0.59$\pm$0.30} & 37.3$\pm$9.6             & {0.14$\pm$0.12} & 18.9$\pm$4.1            \\ 
{\bf\CL}                & {\bf 0.05$\pm$0.04} & {\bf 9.1$\pm$3.4}             & { 0.12$\pm$0.11} & {\bf12.9$\pm$4.4}            \\ \hline
\multicolumn{1}{l|}{} & \multicolumn{2}{c|}{CA-L}                                                                         & \multicolumn{2}{c}{SC-L}                                                                          \\ \hline
BnB        & 2.74$\pm$1.87 & 320.9$\pm$83.1           & 1.54$\pm$1.33 & 115.0$\pm$42.5            \\ 
\RANDOM     & 5.37$\pm$0.75 & 229.2$\pm$24.4          & 3.31$\pm$1.79 & 166.4$\pm$61.3         \\ 
\LBRELAX    & 1.61$\pm$1.50 & 153.0$\pm$50.3          & 1.91$\pm$1.42 & 88.3$\pm$48.9            \\ 
\DM         & 4.56$\pm$0.98 & 254.2$\pm$33.4    & 1.72$\pm$1.19 & 79.1$\pm$42.4            \\ 
\RL         & 4.91$\pm$0.81 & 197.0$\pm$28.5     & 0.66$\pm$0.72 & 116.2$\pm$27.1            \\ 
{\bf\CL}   & {\bf0.09$\pm$0.10} & {\bf116.1$\pm$18.0} & {\bf0.58$\pm$0.45} & {\bf39.2$\pm$23.2}\\ \hline
\end{tabular}
\end{table}

\begin{figure*}[bpth]
    \centering
    \includegraphics[width=0.8\textwidth]{figure_all/legend_ML_horizontal_timeVSobj_3601.eps}

    \subfloat[MVC-S (left) and MVC-L (right).]
    {
        \includegraphics[width=0.24\textwidth]{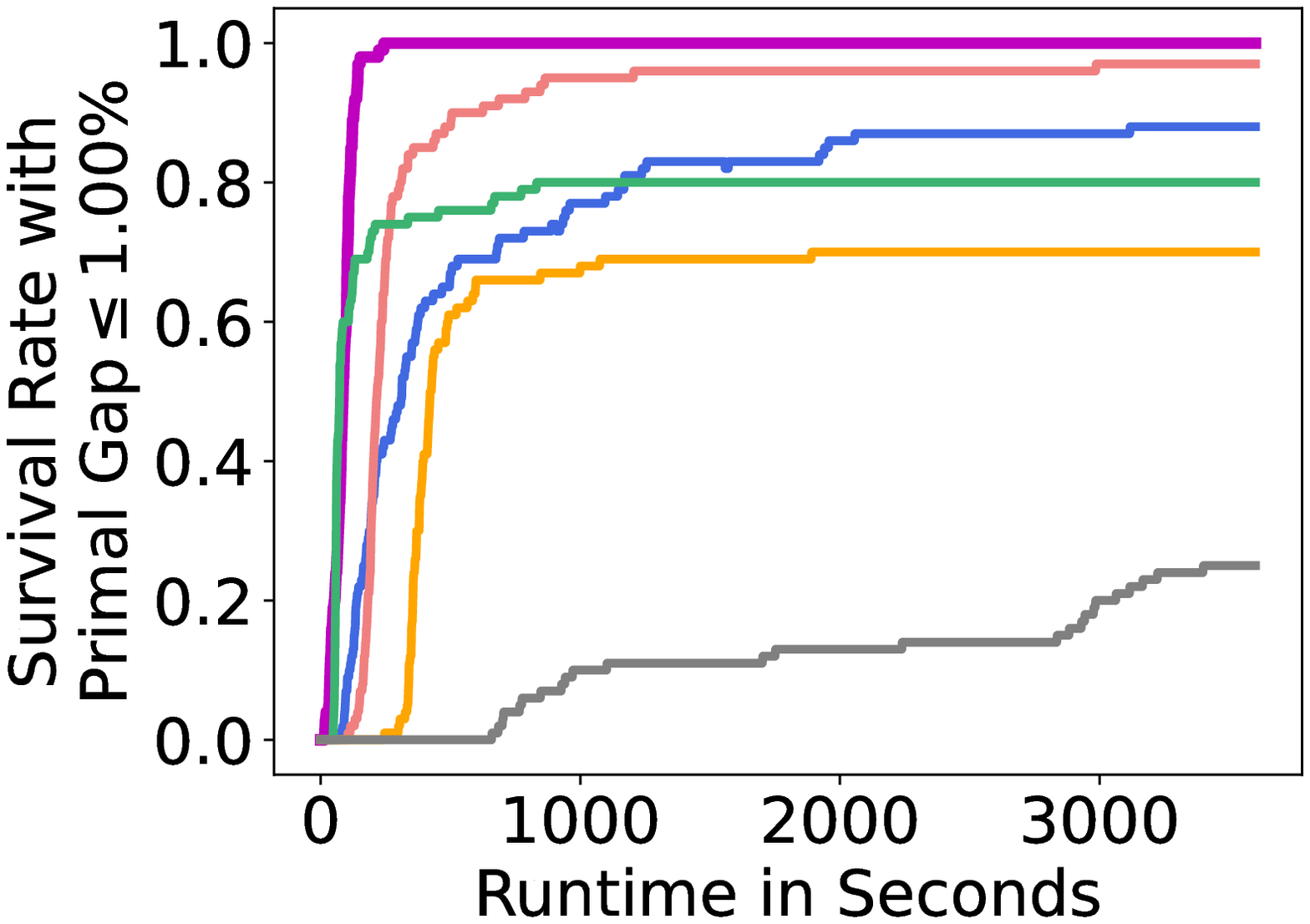}
        \includegraphics[width=0.24\textwidth]{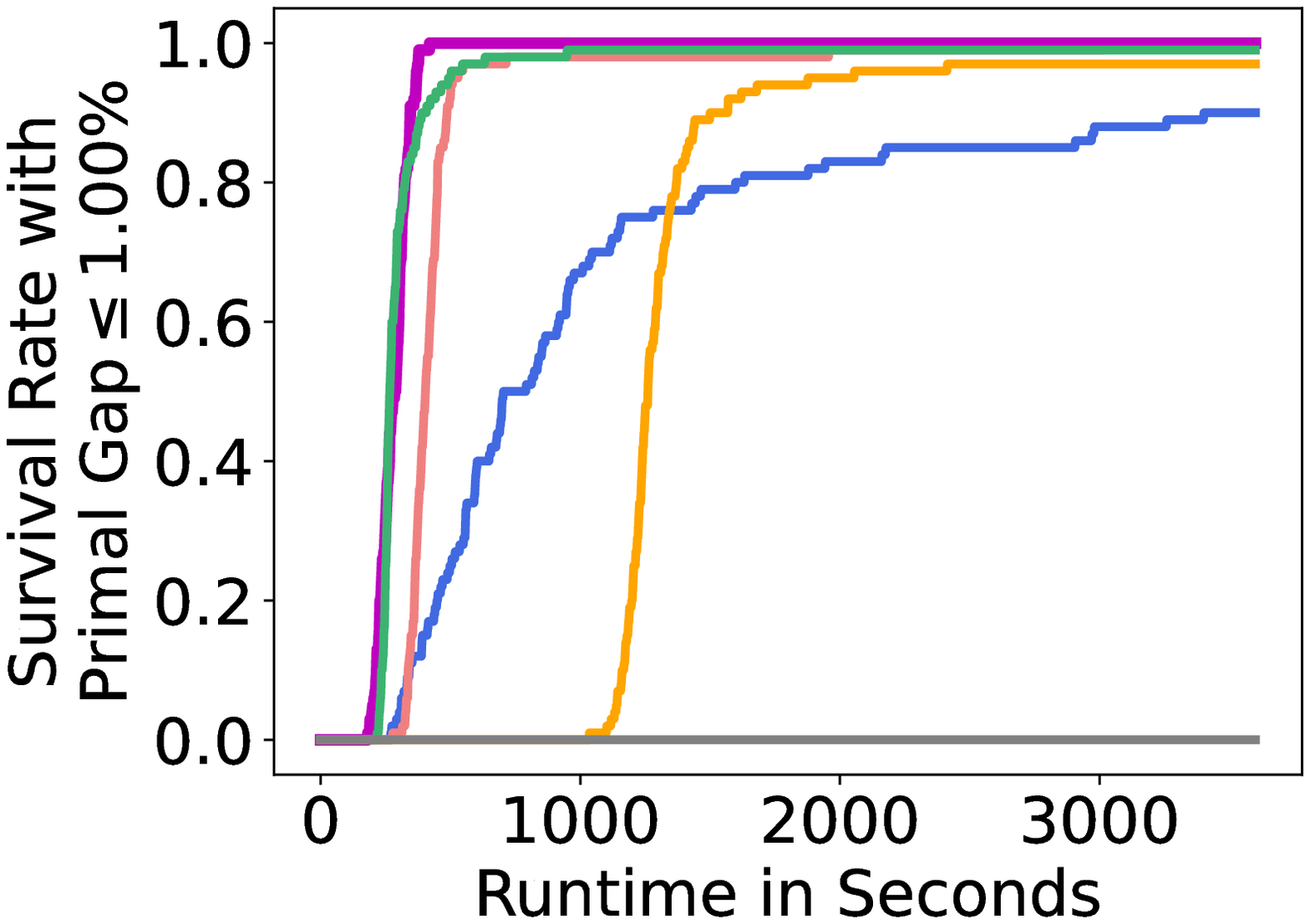}    
    }
    \subfloat[MIS-S (left) and MIS-L (right).]
    {
        \includegraphics[width=0.24\textwidth]{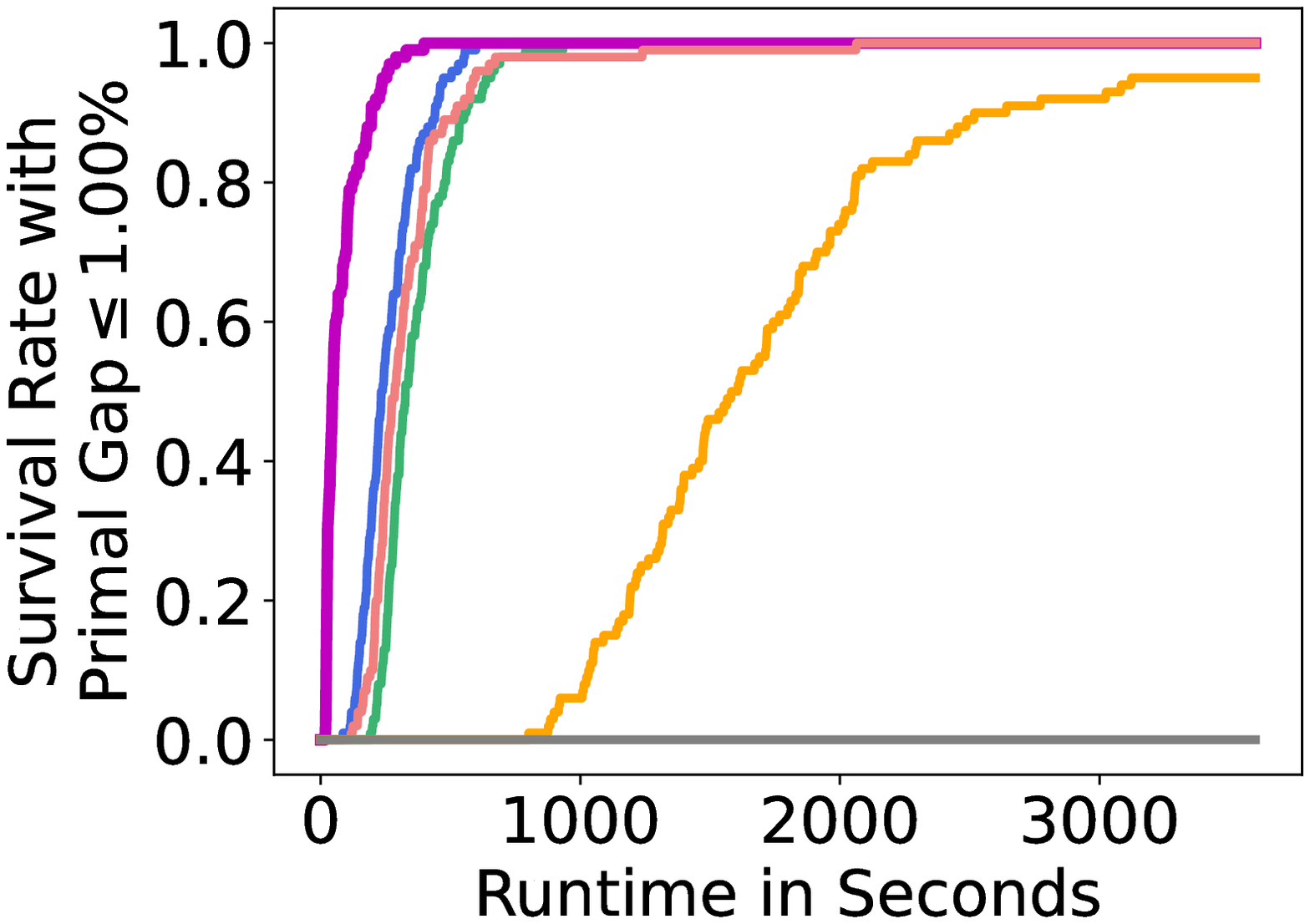}
        \includegraphics[width=0.24\textwidth]{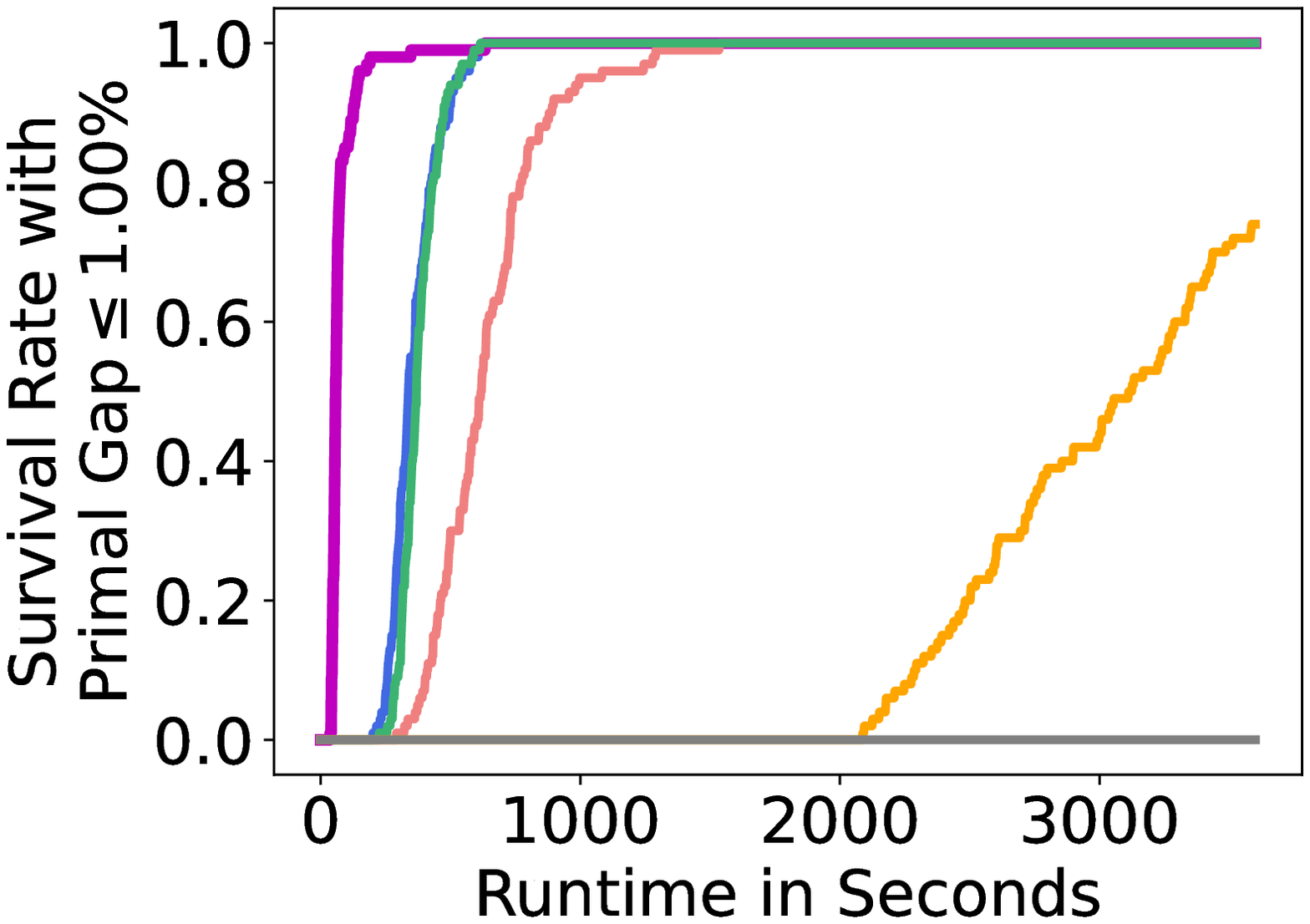}    
    }\\
    \subfloat[CA-S (left) and CA-L (right).]
    {
        \includegraphics[width=0.24\textwidth]{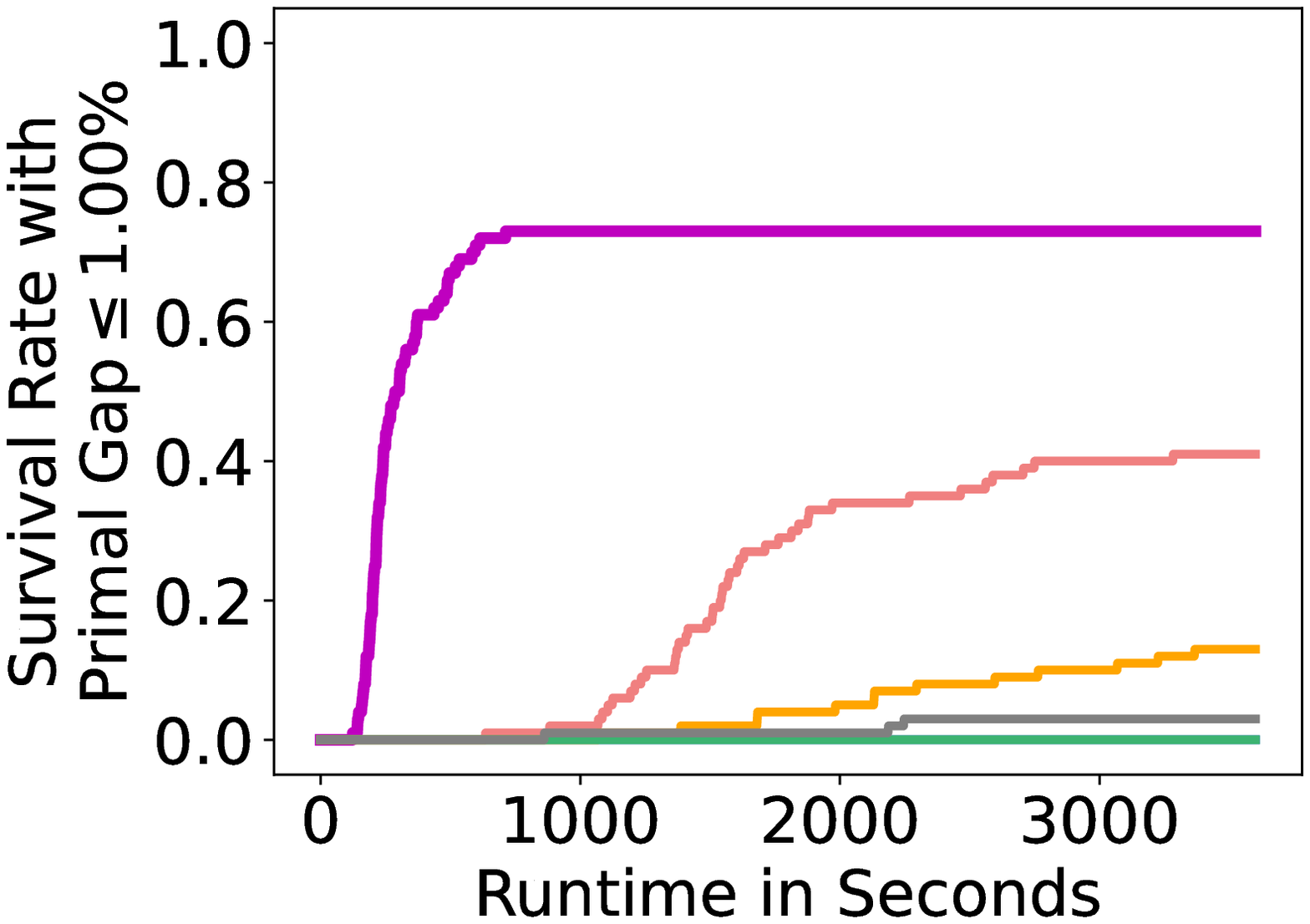}
        \includegraphics[width=0.24\textwidth]{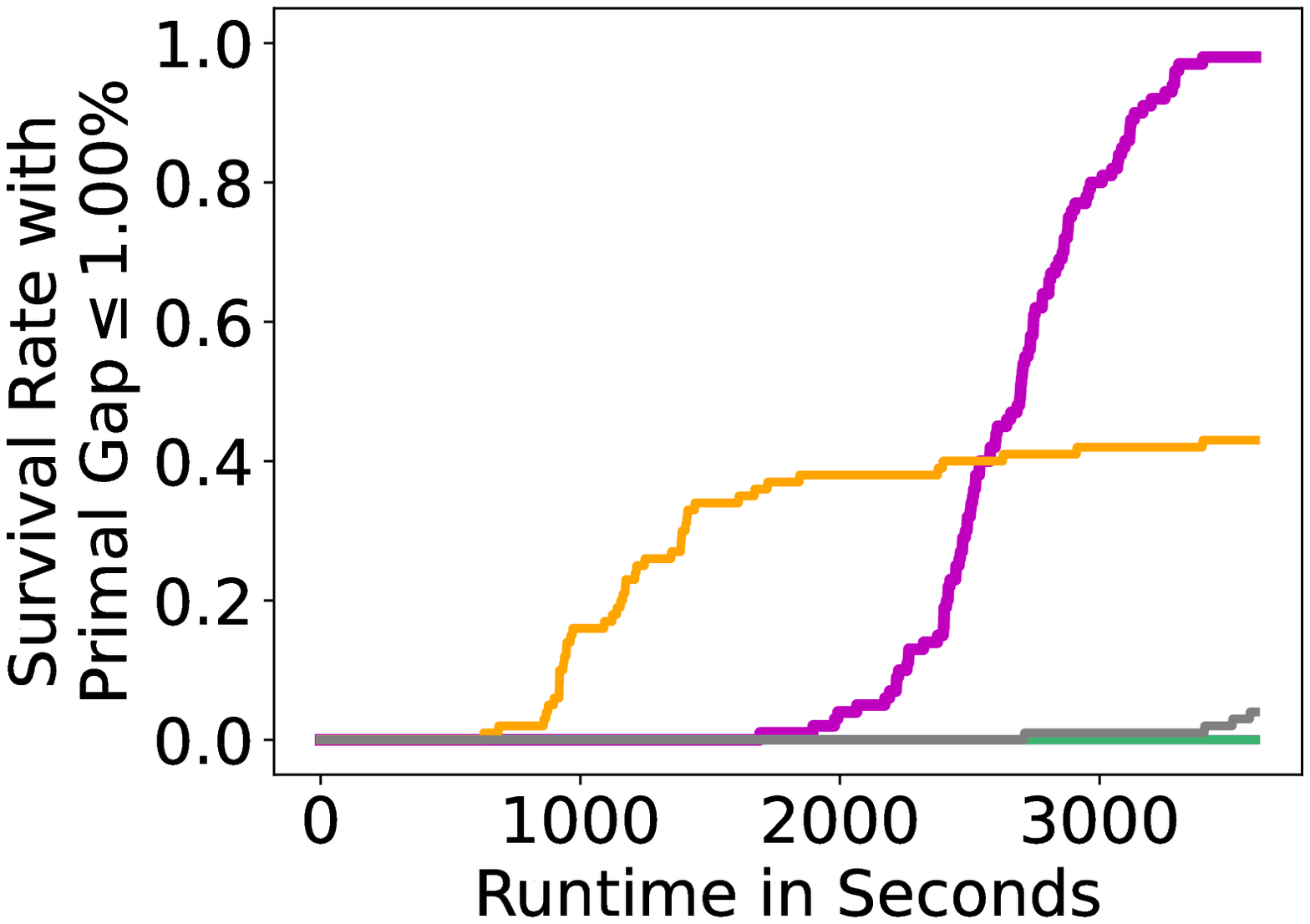}    
    }
    \subfloat[SC-S (left) and SC-L (right).]
    {
        \includegraphics[width=0.24\textwidth]{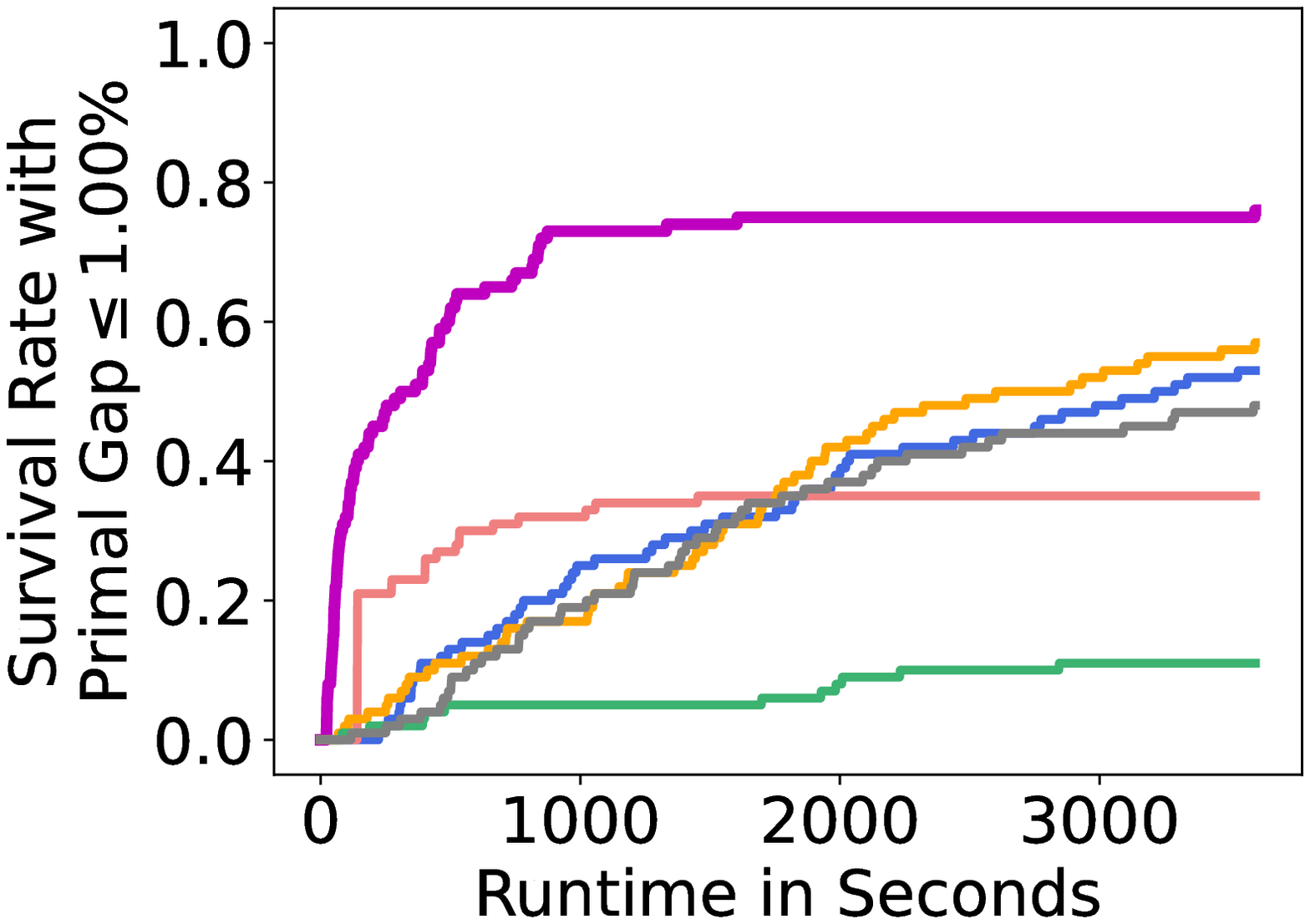}
        \includegraphics[width=0.24\textwidth]{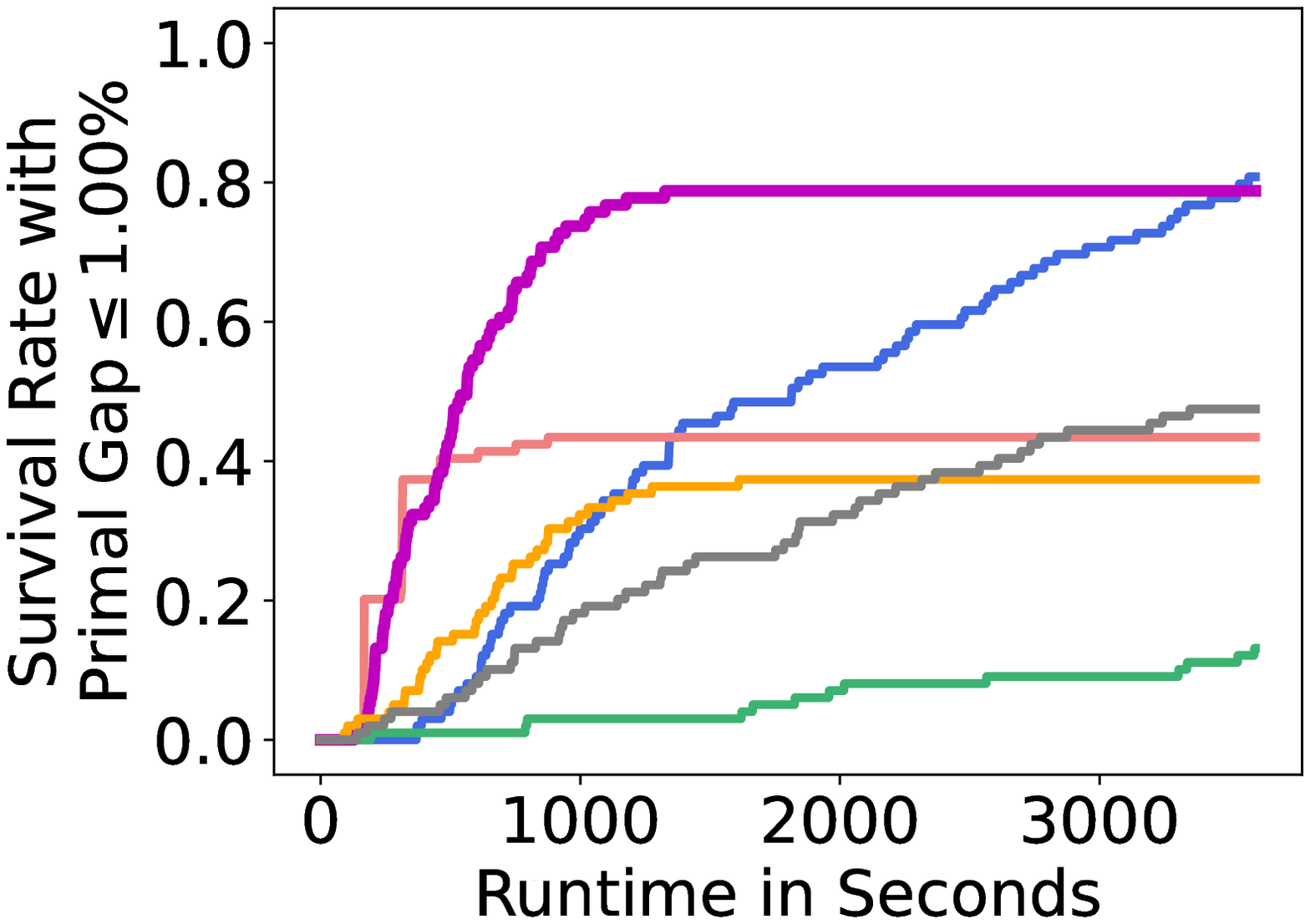}    
    }
    \caption{The survival rate (the higher the better) over 100 test instances as a function of runtime to meet primal gap threshold 1.00\%. For ML approaches, the policies are trained on only small training instances but tested on both small and large test instances.  \label{res::survival}}
\end{figure*}

\begin{figure*}[btp]
    \centering
    \includegraphics[width=0.8\textwidth]{figure_all/legend_ML_horizontal_timeVSobj_3601.eps}

    \subfloat[MVC-S]
    {
        \includegraphics[width=0.23\textwidth]{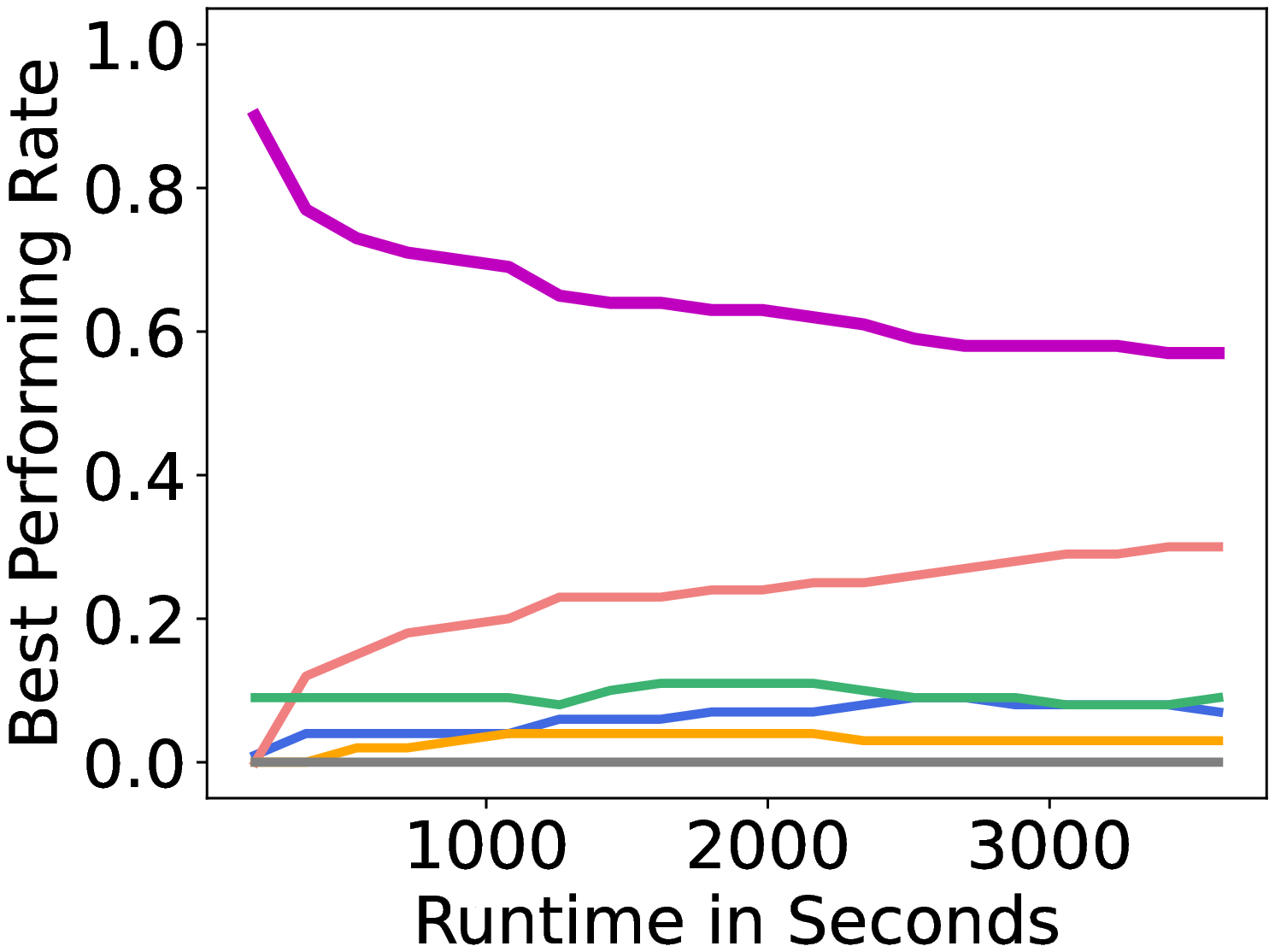}
    }
    \subfloat[MIS-S]
    {
        \includegraphics[width=0.23\textwidth]{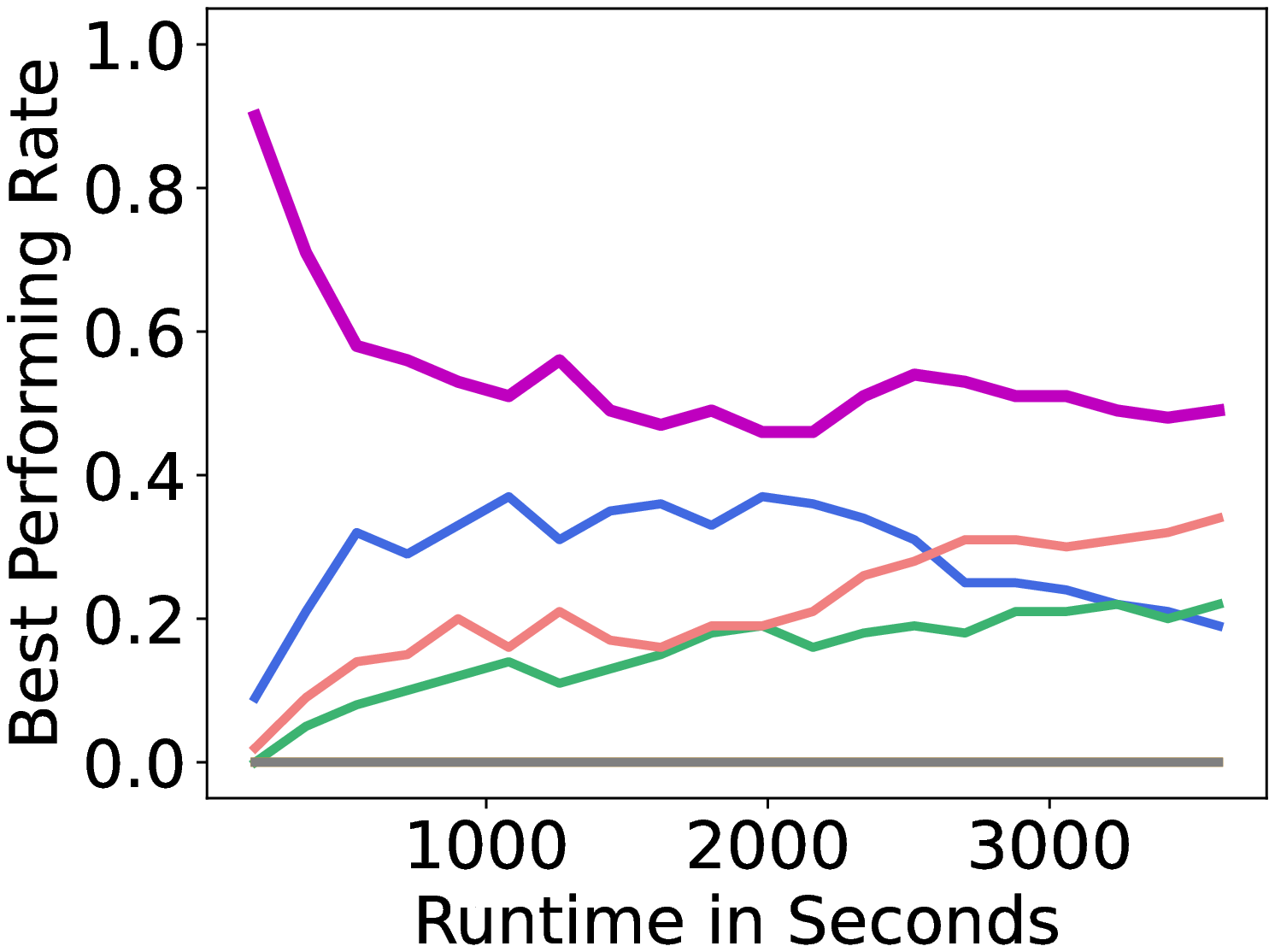}
    }
    \subfloat[CA-S]
    {
        \includegraphics[width=0.23\textwidth]{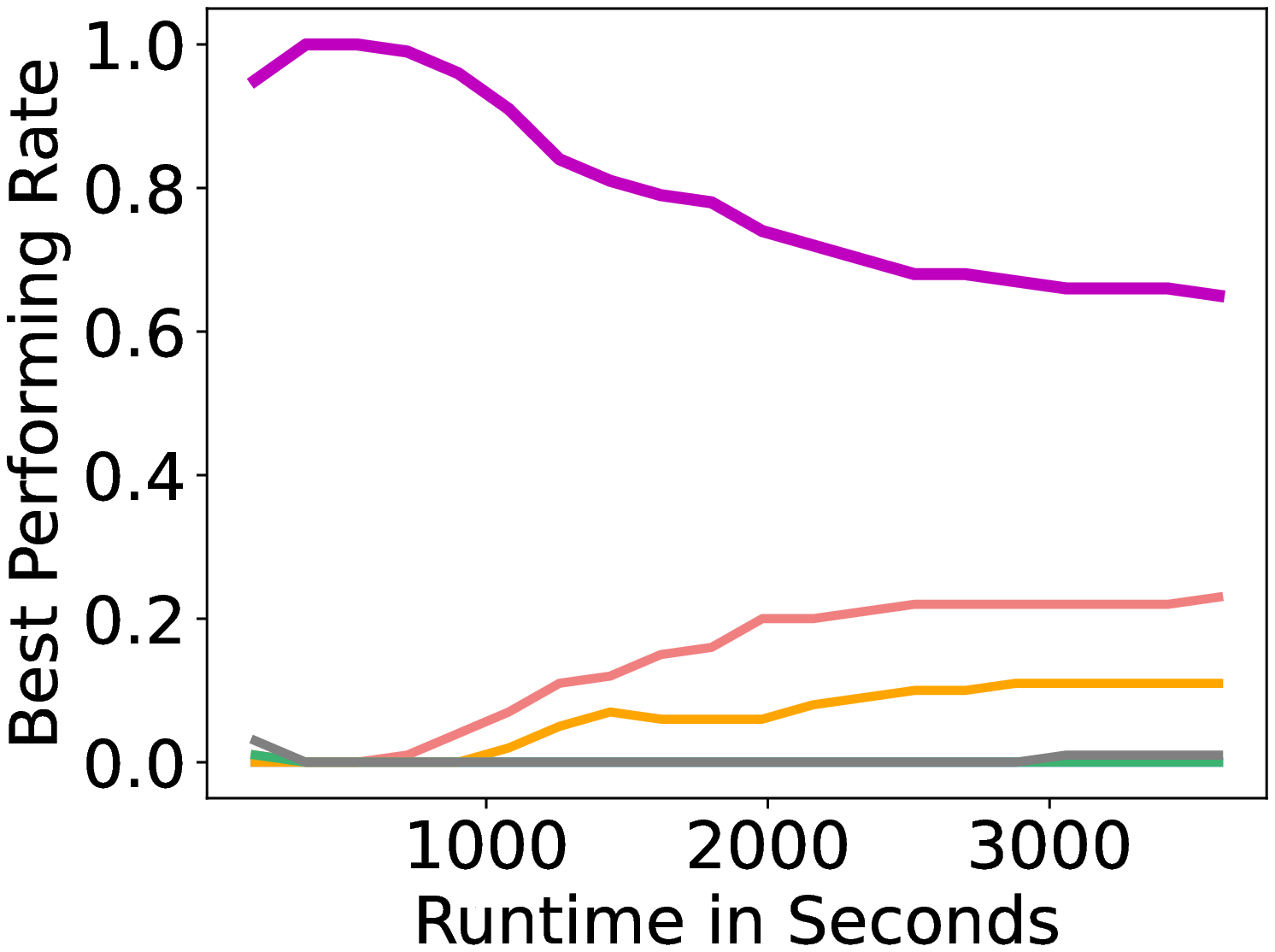}
    }
    \subfloat[SC-S]
    {
        \includegraphics[width=0.23\textwidth]{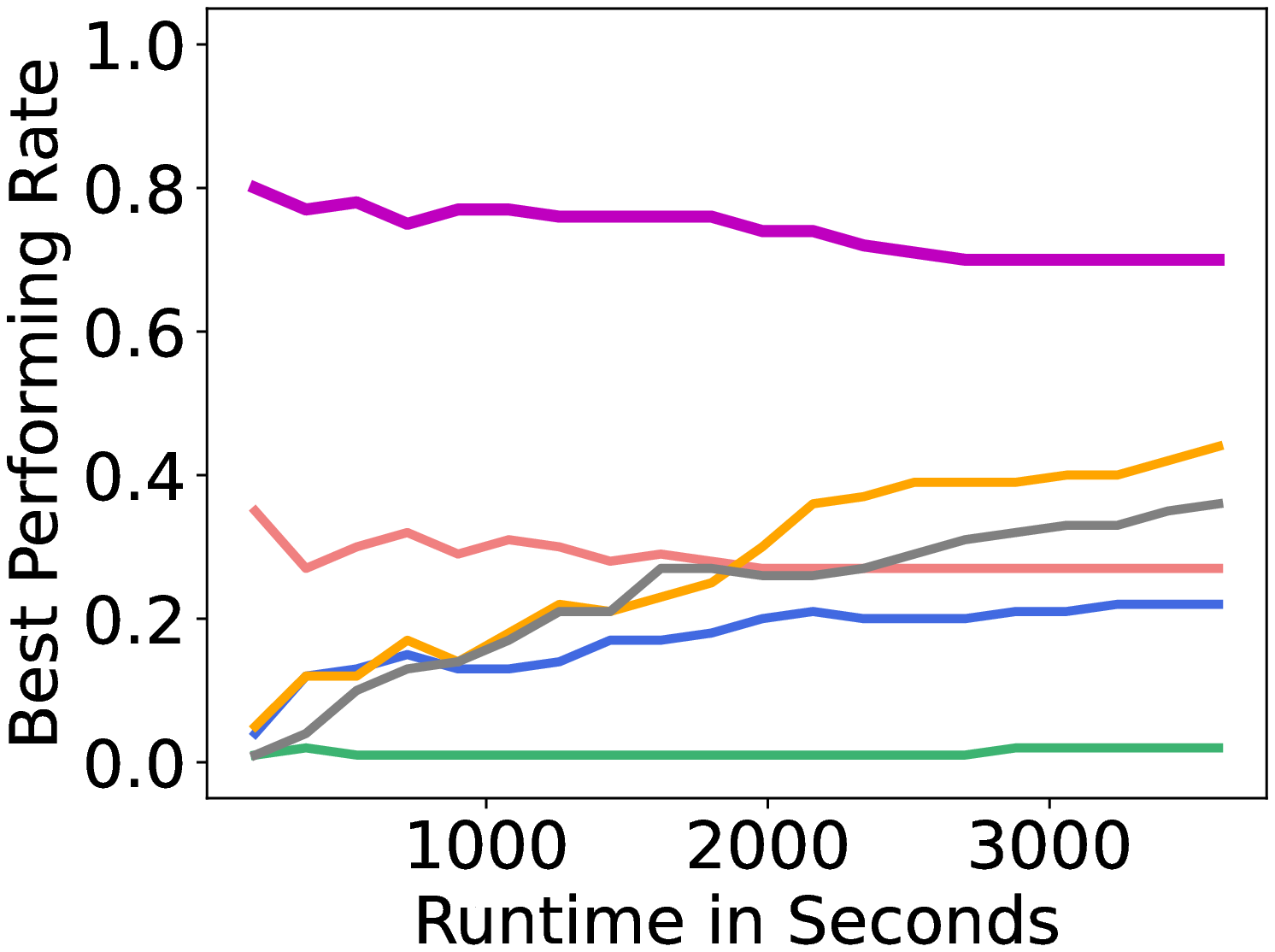}
    }
    \caption{The best performing rate (the higher the better) as a function of runtime on 100 small instances (see Appendix for results on large instances).  The sum of the
best performing rates at a given runtime might sum up greater than 1 since ties are counted multiple times.\label{res::winRate}}
\end{figure*}

\paragraph{Hyperparameters}
We conduct experiments on 2.5GHz Intel Xeon Platinum 8259CL CPUs with 32 GB memory. Trainings are done on a NVIDIA A100 GPU with 40 GB memory. All experiments use the hyperparameters described below unless stated otherwise. 
We use SCIP (v8.0.1) \cite{BestuzhevaEtal2021OO} to solve the sub-ILP in every iteration of LNS. 
To run LNS, we find an initial solution by running SCIP for 10 seconds. We set the time limit to 60 minutes to solve each instance and 2 minutes for solving the sub-ILP in every LNS iteration.
All approaches require a neighborhood size $k^t$ in LNS, except for BnB and \RL ($k^t$ in \RL is defined implicitly by how the policy is used). For \LBRELAX, \DM and \CL, the initial neighborhood size $k^0$ is set to  $100, 3000, 1000$ and $150$ for MVC, MIS, CA and SC, respectively, except $k^0$ is set to $150$ for SC for \DM; for \RANDOM, it is set to $200, 3000, 1500$ and $200$ for MVC, MIS, CA and SC, respectively. All approaches use adaptive neighborhood sizes with  $\gamma=1.02$ and $\beta=0.5$, except for BnB and \RL. For \DM, when applying its learned policies, we use the sampling methods on MVC and CA instances and the greedy method on SC and MIS instances. For \CL, the greedy method is used on all instances. 
Additional details on hyperparameter tunings are provided in Appendix.

For data collection, we use different neighborhood sizes $k^0=50, 500, 200$ and $50$ for MVC, MIS, CA and SC, respectively, which we justify in Section \ref{sec::expresults}. We set $\gamma=1$ and run LNS with LB until no new incumbent solution found. The runtime limit for solving LB in every iteration is set to 1 hour. 
For training, we use the Adam optimizer \cite{kingma2015adam} with learning rate $10^{-3}$. We use a batch size of 32 and train for 30 epochs (the training typically converges in less than 20 epochs and 24 hours).

\subsection{Results} \label{sec::expresults}

Figure \ref{res::gap} shows the primal gap as a function of runtime.  Table \ref{res::smalltable60}  presents the average primal gap and primal integral at 60 minutes runtime cutoff on small and large instances, respectively (see results at 15, 30 and 45 minutes runtime cutoff in Appendix). Note that we were not able to reproduce the results on CA-S and CA-L reported in \citet{wu2021learning} for \RL despite using their code and repeating training with five random seeds. \CL shows significantly better  anytime performance than all baselines on all problems, achieving the smallest average primal gap and primal integral. It also demonstrates strong generalization performance on large instances unseen during training. Figure \ref{res::survival} shows the survival rate to meet the $1.00\%$ primal gap threshold. \CL achieves the best survival rate at 60 minutes runtime cutoff on all instances, except that, on SC-L, its final survival rate is slightly worse than \RL  but it achieves the rate with much shorter runtime. On MVC-L, MIS-S and MIS-L instances, several baselines achieve the same survival rate as \CL but it always achieves the rates with the shortest runtime. Figure \ref{res::winRate} shows the best performing rate on the small test instances where \CL consistently performs best on  50\% to 100\% of the instances.  In Appendix, we present strong results in comparison with two more baselines and on one more performance metric.

\begin{figure*}[btp]
    \centering
\scriptsize

\begin{tabular}{c|rr|rr|rr|rr}
\hline

                       & \multicolumn{2}{c|}{MVC-S}  & \multicolumn{2}{c|}{MIS-S}  & \multicolumn{2}{c|}{CA-S}  & \multicolumn{2}{c}{SC-S}                                                                         \\ \hline
                                              & \multicolumn{1}{c}{NH size}             & \multicolumn{1}{c|}{Runtime}         & \multicolumn{1}{c}{NH size}             & \multicolumn{1}{c|}{Runtime}& \multicolumn{1}{c}{NH size}             & \multicolumn{1}{c|}{Runtime}& \multicolumn{1}{c}{NH size}             & \multicolumn{1}{c}{Runtime} \\ \hline

LB  &100 & 3600$\pm$0  &  3,000 & 3600$\pm$0 & 1,000 & 3600$\pm$0& 100 &3600$\pm$0\\ 
LB (data collection) & 50 & 3600$\pm$0 & 500 &3600$\pm$0 &   200 & 3600$\pm$0& 50 &3600$\pm$0\\ 
\DM & 100 & 2.1$\pm$0.1& 3,000 & 1.3$\pm$0.2 & 1,000 & 20.8$\pm$13.1 & 150 &120.9$\pm$1.3 \\ 
\CL & 100 & 2.2$\pm$0.1& 3,000 & 1.3$\pm$0.1 & 1,000 & 25.1$\pm$15.3 & 100 & 50.1$\pm$10.4\\ \hline
\end{tabular}
    \includegraphics[width=0.50\textwidth]{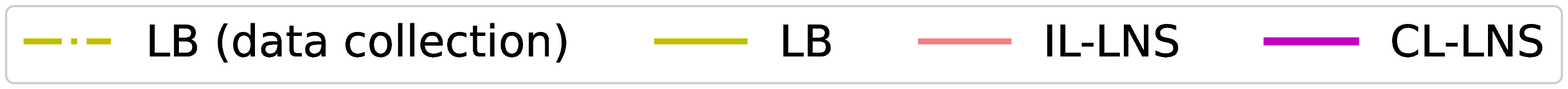}

    \subfloat[MVC-S]
    {
        \includegraphics[height=2.8cm]{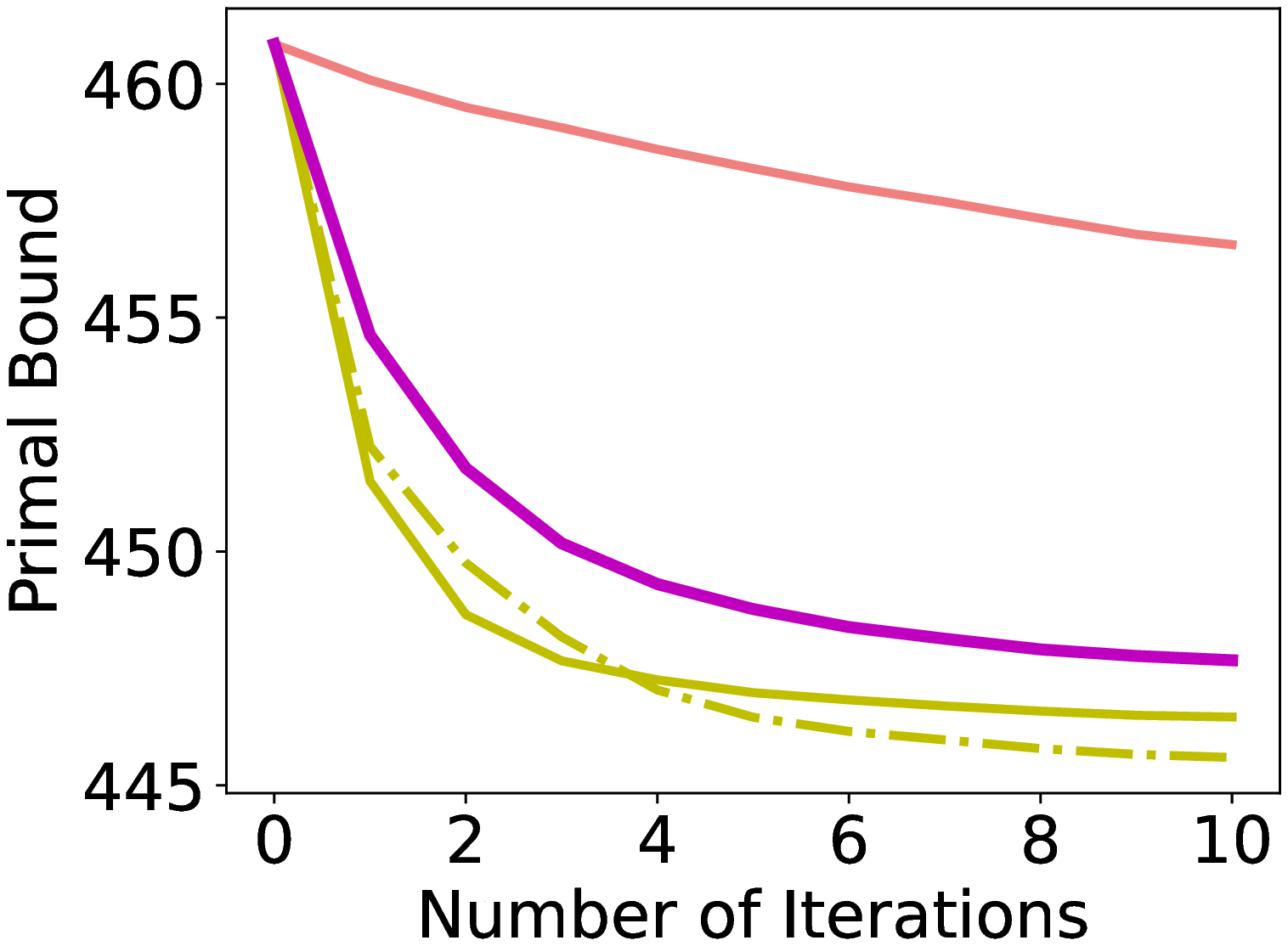}
    }
    \subfloat[MIS-S]
    {
        \includegraphics[height=2.8cm]{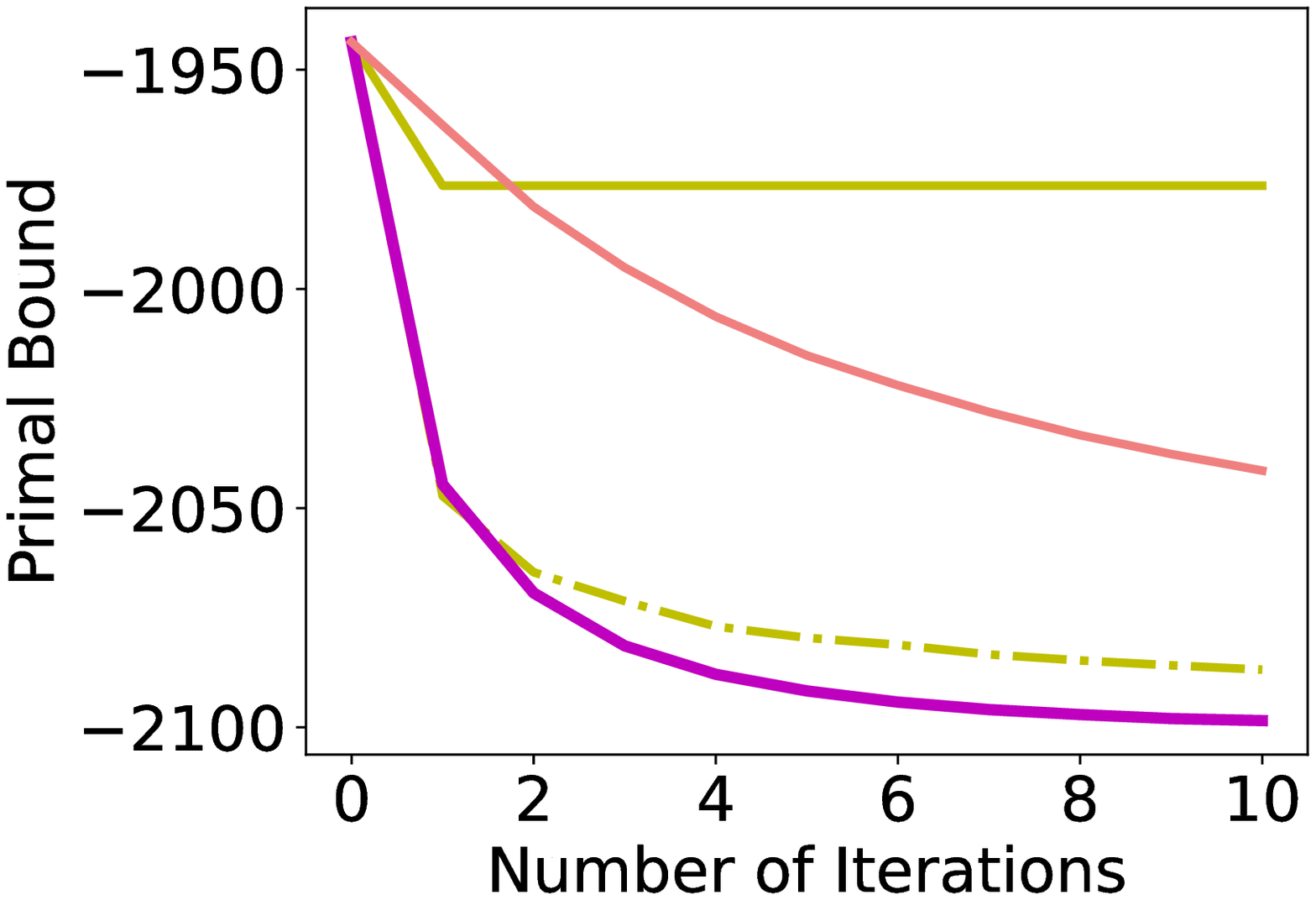}    
    }
    \subfloat[CA-S]
    {
        \includegraphics[height=2.8cm]{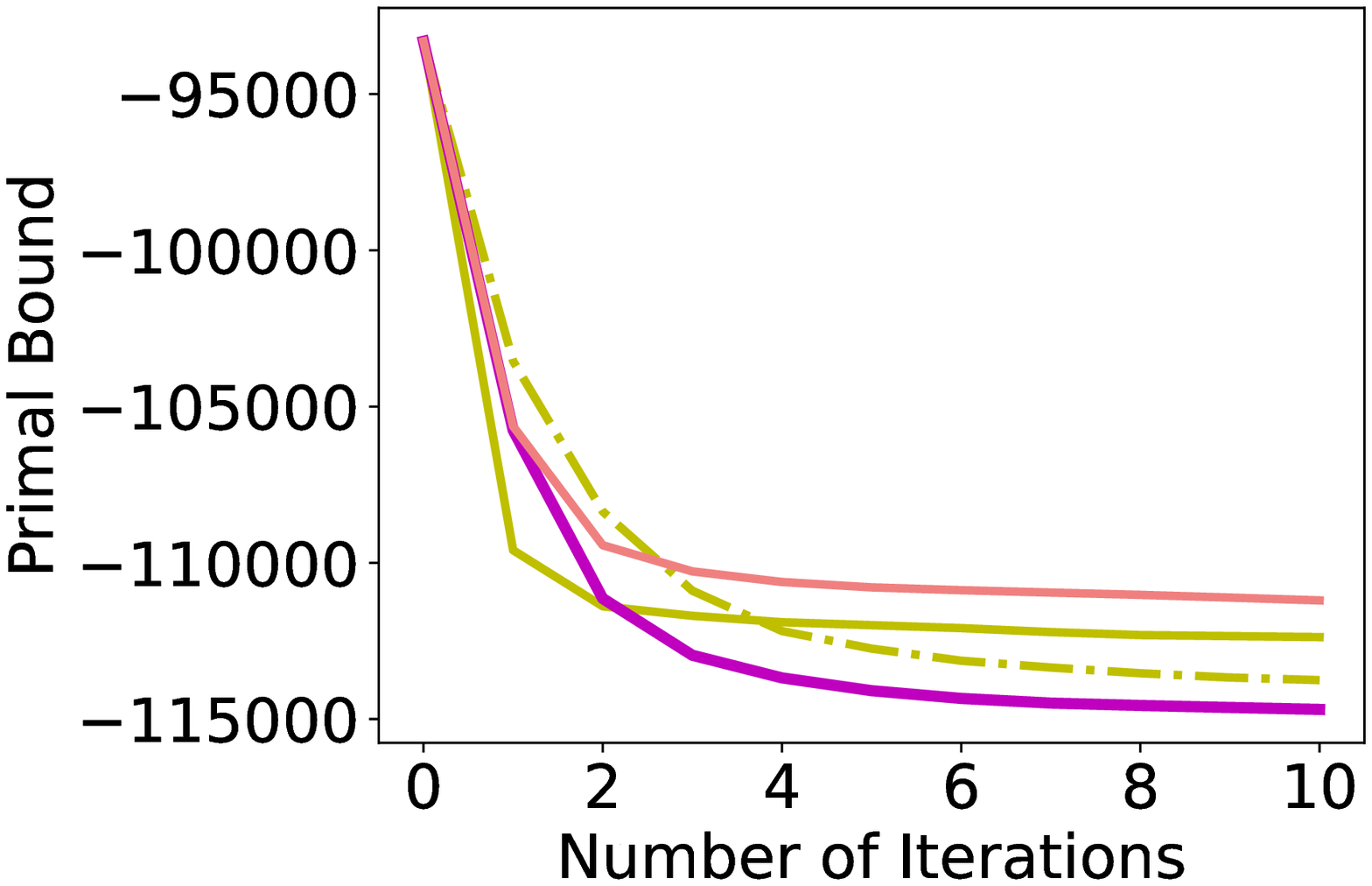}
    }
    \subfloat[SC-S]
    {
        \includegraphics[height=2.8cm]{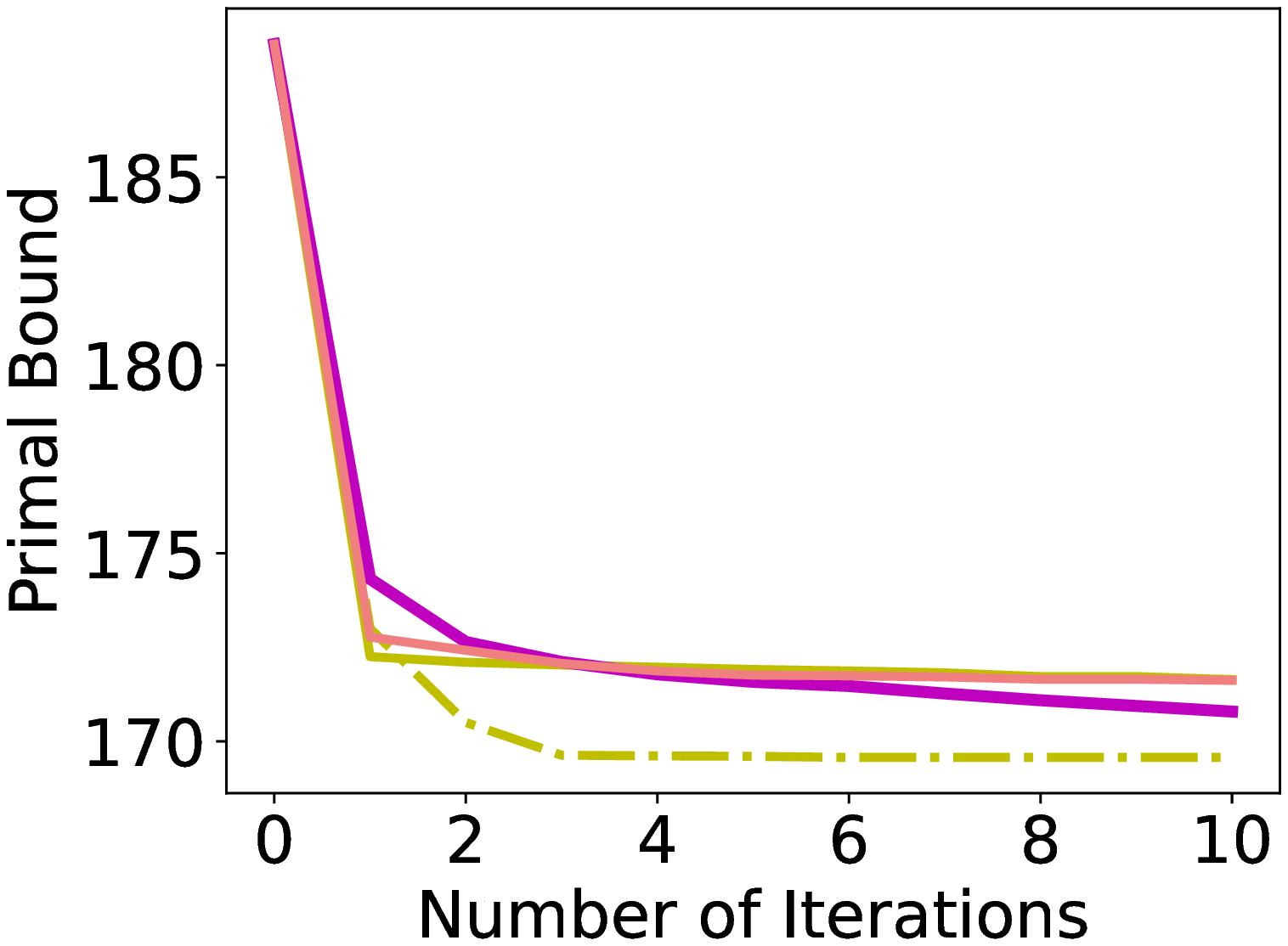}
    }
    \caption{The primal bound (the lower the better) as a function of number of iterations, averaged over 100 small test instances. LB and LB (data collection) are LNS with LB using the neighborhood sizes fune-tunded for \CL and for data collection, respectively. The table shows the neighborhood size (NH size) and the average runtime in seconds  (with standard deviations) per iteration for each approach.   \label{res::boundperIteration}}
\end{figure*}

\paragraph{Comparison with LB (the Expert)}

Both \DM and \CL learn to imitate LB. On the small test instances, we run LB with two different neighborhood sizes, one that is fine-tuned in data collection and the other the same as \CL, for 10 iterations and compare its per iteration performance with \DM and \CL.  This allows us to compare the quality of the learned policies to the expert independently of their speed. The runtime limit per iteration for LB is set to 1 hour. Figure \ref{res::boundperIteration} shows the primal bound as a function of number of iterations. The table in the figure summarizes the neighborhood sizes and the average runtime per iteration. For LB, the result shows that the neighborhood size affects the overall performance. Intuitively, using a larger neighborhood size in LB allows LNS to find better incumbent solutions due to being able to explore larger neighborhoods. However, in practice, LB becomes less efficient in finding good incumbent solutions as the neighborhood size increases, sometimes even performs worse than using a smaller neighborhood size (the one for data collection).  The neighborhood size for data collection is fine-tuned on validation sets to achieve the best primal bound upon convergences, allowing the ML models to observe demonstrations that lead to as good primal bounds as possible in training. However, when using the ML models in testing, we have the incentive to use a larger neighborhood size and fine tune it since we no longer suffer from the bottleneck of LB.  We therefore fine tune the neighborhood sizes for \DM and \CL separately on validation sets. \CL has a strong per-iteration performance that is consistently better than \DM. With the fine-tuned neighborhood size, \CL even outperforms the expert that it learns from (LB for data collection) on MIS-S and CA-S.

\paragraph{Ablation Study}

\begin{figure}[htbp]
    \centering
    \includegraphics[width=0.5\textwidth]{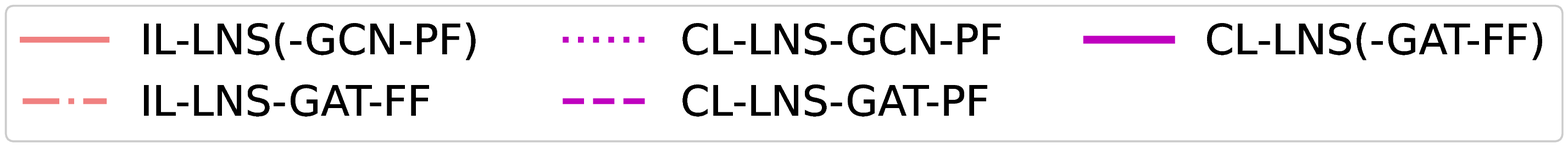}

    \subfloat[MVC-S]
    {
        \includegraphics[width=0.24\textwidth]{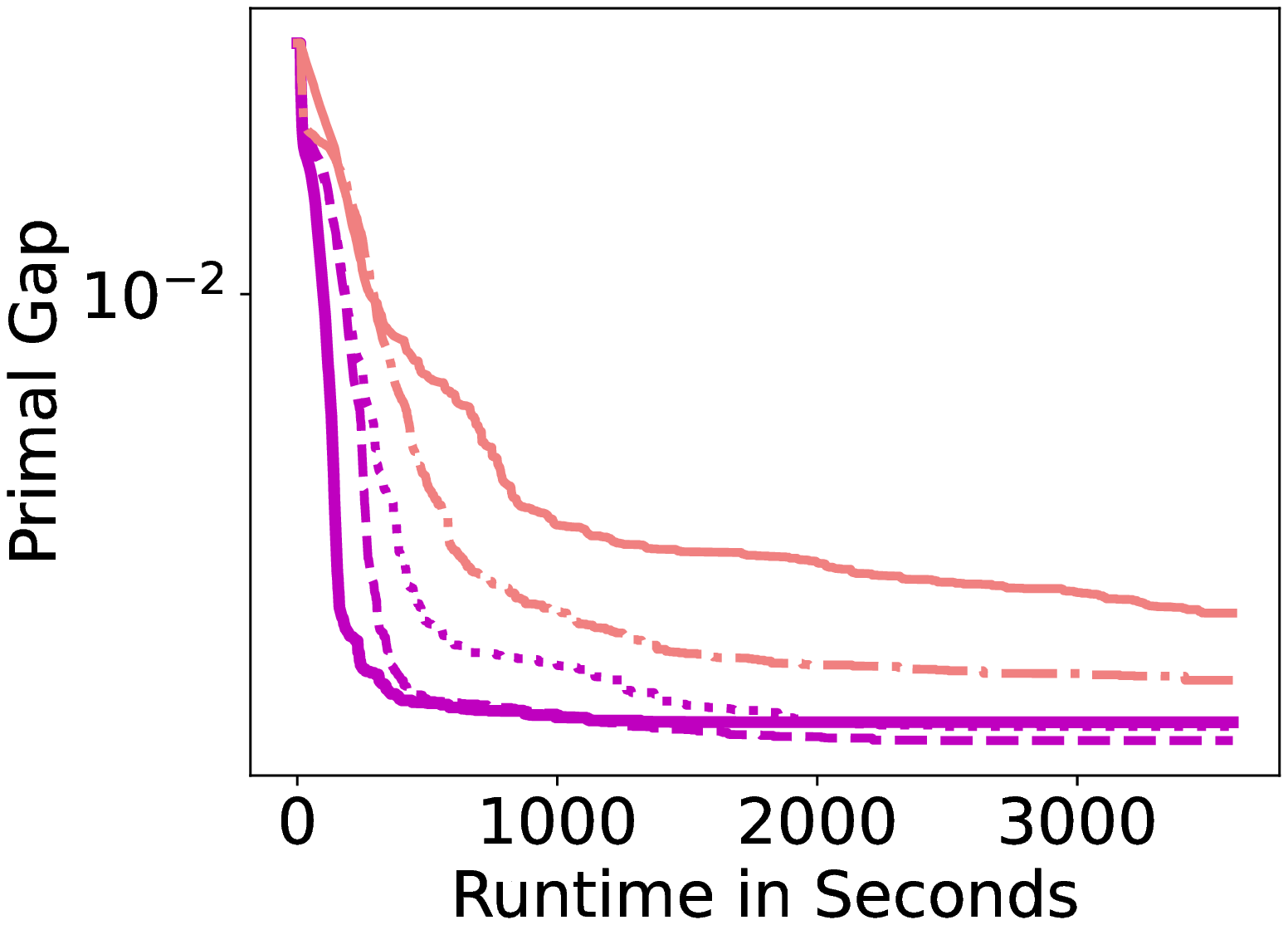}
    }
    \subfloat[CA-S]
    {
        \includegraphics[width=0.24\textwidth]{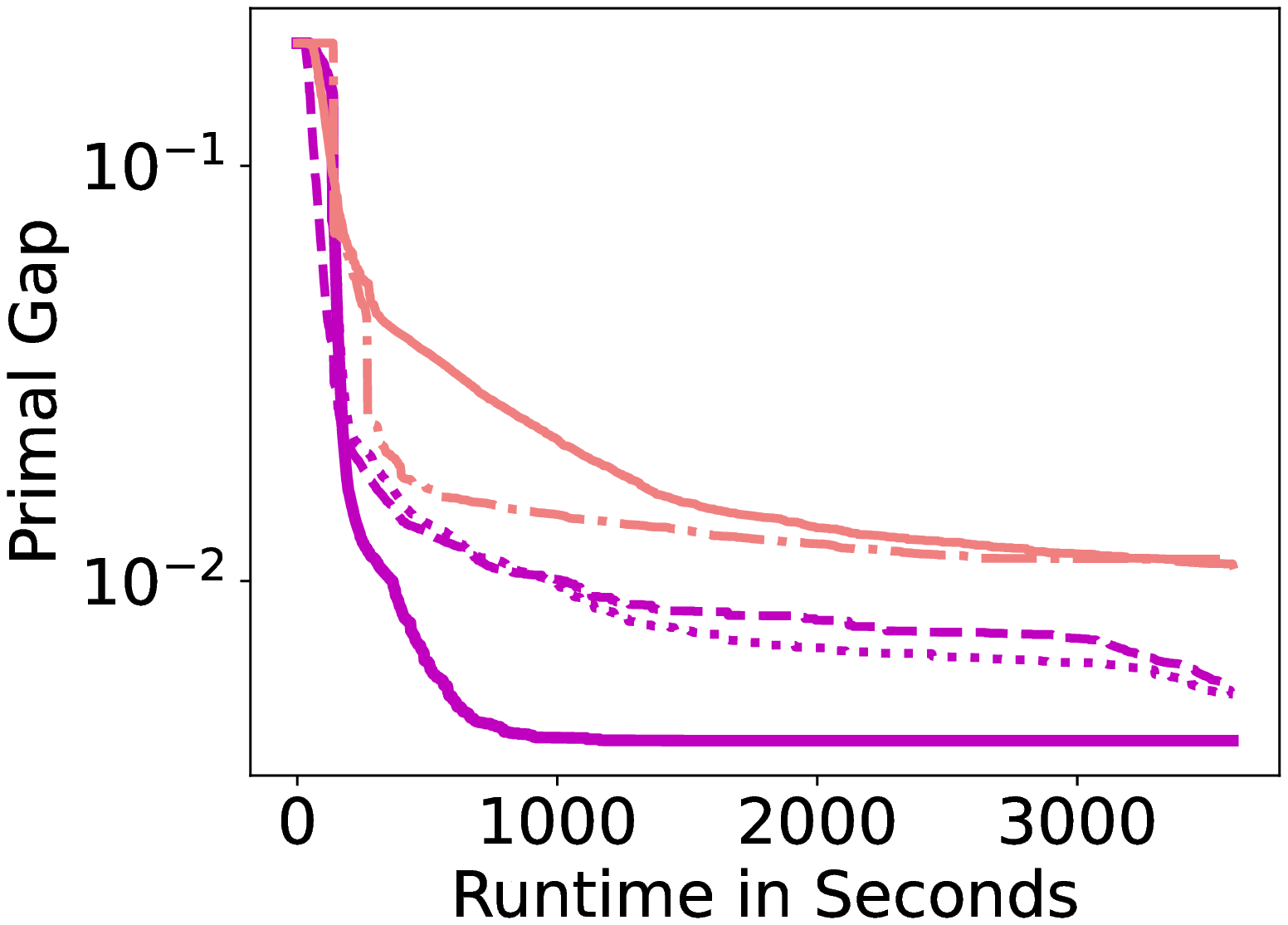}    
    }
   
    \caption{Ablation study: The primal gap (the lower the better) as a function of time, averaged over 100 small test instances.\label{res::ablation_gap}} 
\end{figure}

We evaluate how contrastive learning  and  two enhancements  contribute to \CL's performance. Compared to \DM,  \CL uses (1) addition features from \citet{khalil2016learning} and (2) GAT instead of GCN. We denote by ``FF'' the full feature set  used in \CL and ``PF'' the partial feature set in \DM. In addition to \DM and \CL, we evaluate the performance of \DM with FF and GAT (denoted by IL-LNS-GAT-FF), \CL with GCN and PF (denoted by CL-LNS-GCN-PF) as well as \CL with GAT and PF (denoted by CL-LNS-GAT-PF) on MVC-S and CA-S. Figure \ref{res::ablation_gap} shows the primal gap as a function of runtime. Table \ref{res::ablation_table} presents the primal gap and primal integral at 60 minutes runtime cutoff. The result shows that IL-LNS-GAT-FF, imitation learning with the two enhancements, still performs worse than CL-LNS-GCN-PF without any enhancements.  CL-LNS-GCN-PF and CL-LNS-GAT-PF perform similarly in terms of the primal gaps but CL-LNS-GAT-PF has better primal integrals, showing the benefit of replacing GCN with GAT. On MVC-S, three variants of \CL have similar average primal gaps and on CA-S, \CL has better average primal gap than the other two variants. But adding the two enhancement helps improve the primal integral, leading to the overall best performance of \CL on both MVC-S and CA-S. 

\begin{table}[htbp]

\centering
\caption{Ablation study: Primal gap (PG) (in percent) and primal integral (PI) at 60 minutes runtime cutoff, averaged over 100 small test instances and their standard deviations. 
``$\downarrow$'' means the lower the better. \label{res::ablation_table}}

\scriptsize
\begin{tabular}{c|rr|rr}
\hline
                       & \multicolumn{1}{c}{PG (\%) $\downarrow$}         & \multicolumn{1}{c|}{PI $\downarrow$}     & \multicolumn{1}{c}{PG (\%) $\downarrow$}         & \multicolumn{1}{c}{PI $\downarrow$} \\ \hline

                       & \multicolumn{2}{c|}{MVC-S}                                                                        & \multicolumn{2}{c}{CA-S}                                                                         \\ \hline
IL-LNS(-GCN-PF)        & {0.29$\pm$0.23} & {19.2$\pm$10.2} &  1.09$\pm$0.51 & 90.0$\pm$20.8            \\ 
IL-LNS-GAT-FF       & 0.24$\pm$0.17 & 15.3$\pm$7.3 & 1.13$\pm$0.63 & 78.9$\pm$22.7 \\ \hline
CL-LNS-GCN-PF       & 0.17$\pm$0.10 & 11.4$\pm$8,8 & 0.75$\pm$0.40& 57.9$\pm$21.2\\
CL-LNS-GAT-PF       & {\bf0.16$\pm$0.09} & 10.1$\pm$0.6 & 0.76$\pm$0.39 & 53.8$\pm$22.1 \\
{\bf CL-LNS(-GAT-FF)}      & { 0.17$\pm$0.09} & {\bf 8.7$\pm$6.7} &  {\bf 0.65$\pm$0.32} & {\bf50.7$\pm$22.7} \\ \hline

\end{tabular}
\end{table}

\section{Conclusion}

We proposed \CL, that uses a contrastive loss to learn efficient and effective destroy heuristics in LNS for ILPs. We presented a novel data collection process tailored for \CL and used GAT with a richer set of features to further improve its performance. Empirically, \CL significantly outperformed state-of-the-art approaches on four ILP benchmarks w.r.t. to the primal gap, the primal integral, the best performing rate and the survival rate. \CL achieved good generalization performance on out-of-distribution instances. 
It is future work to learn policies that can generalize across problem domains. \CL does not guarantee optimality and it is also interesting future work to integrate it in BnB for which many other learning techniques are developed. Our approach is closely related to and could be useful for many problems of identifying substructures in combinatorial searches, for example, identifying backdoor variables in ILPs \cite{ferber2022learning} and selecting neighborhoods in LNS for other COPs.


\bibliography{example_paper}
\bibliographystyle{icml2023}

\newpage
\appendix
\onecolumn
\section*{\LARGE Appendix}

\section{Additional Related Work}

\subsection{LNS-based primal heuristics in BnB}

LNS-based primal heuristics is a family of primal heuristics in BnB and have been studied extensively in past decades. With the same purpose of improving primal bounds, the main differences between the LNS-based primal heuristics in BnB and LNS for ILPs are: (1) LNS-based primal heuristics are executed periodically at different search tree nodes during the search and the execution schedule is itself dynamic, because they are often more expensive to run than the other primal heuristics in BnB; (2) the destroy heuristics in LNS-based primal heuristics are often designed to use information specific to BnB, such as the dual bound and the LP relaxation at a search tree node, and they are not directly applicable in LNS for ILPs in our setting. 

Next, we briefly summarize the destroy heuristics in LNS-based primal heuristics:
\begin{itemize}
    \item Crossover heuristics \cite{rothberg2007evolutionary}: it destroys variables that have different values in a set of selected known solutions (typically two). The Mutation heuristics \cite{rothberg2007evolutionary} destroys a random subset of variables. 
\item Relaxation Induced Neighborhood Search (RINS) \cite{danna2005exploring}: it destroys variables whose values disagree in the solution of the LP relaxation at the search tree node and the incumbent solution. 

\item Relaxation Enforced Neighborhood Search (RENS) \cite{berthold2014rens}: it restricts the neighborhood to be the feasible roundings of the LP relaxation at the current search tree node. 
\item Local Branching (LB)\cite{fischetti2003local}: it restricts the neighborhood to a ball around the current incumbent solution. 
\item Distance Induced Neighborhood Search (DINS) \cite{ghosh2007dins}: it takes the intersection of the neighborhoods of the Crossover, Local Branching and Relaxation Induced Neighborhood Search heuristics. 
\item Graph-Induced Neighborhood Search (GINS) \cite{maher2017scip}: it destroys the breadth-first-search neighborhood of a variable in the bipartite graph representation of the ILP. 
\end{itemize}
Recently, an adaptive LNS primal heuristic  \cite{hendel2022adaptive} has been proposed to combine the power of these heuristics, where it essentially solves a multi armed bandit problem to choose which heuristic to apply.

\subsection{Learning to Solve Other COPs}

ML has been applied to solve a number of COPs, including traveling salesman problems \cite{xin2021neurolkh,zheng2021combining}, vehicle routing \cite{kool2018attention}, boolean satisfiability \cite{selsam2018learning,amizadeh2018learning} and general graph optimization problems \cite{khalil2017learning,li2018combinatorial}.

\section{Network Architecture}

We give full details of the GAT architecture described in Section \ref{sec::policyNetwork}.
The policy takes as input the state $\bs^t$ and output a score vector $\pi_{\btheta}(\bs^t)\in [0,1]^n$, one score per variable.
 We use  2-layer MLPs with 64 hidden units per layer and ReLU as the activation function to map each node feature and edge feature to $\mathbb{R}^d$ where $d=64$. 

Let $\bfv_j, \bfc_i, \bfe_{i,j}\in\mathbb{R}^{d}$ be the embeddings of the $i$-th variable, $j$-th constraint and the edge connecting them output by the embedding layers. We perform two rounds of message passing through the GAT.  In the first round, each constraint node $\bfc_i$ attends to its neighbors $\calN_i$ using an attention stucture with $H=8$ attention heads: 
$$
\bfc'_i = \frac{1}{H}\sum_{i=1}^{H}\left(\alpha_{ii,1}^{(h)}\btheta_{c,1}^{(h)} \bfc_i+\sum_{j\in\calN_i}\alpha_{ij,1}^{(h)}\btheta_{v,1}^{(h)}\bfv_j\right)
$$
where $\btheta_{c,1}^{(h)}\in \mathbb{R}^{d\times d}$ and $\btheta_{v,1}^{(h)}\in \mathbb{R}^{d\times d}$ are learnable weights. The updated constraint embeddings $\bfc'_i$ are averaged across $H$ attention heads using attention weights \cite{brody2021attentive}
$$
\alpha^{(h)}_{ij,1} = \frac{\exp(\bw_1^{\mathsf{T}}\rho([\btheta_{c,1}^{(h)} \bfc_i,\btheta_{v,1}^{(h)}\bfv_j,\btheta_{e,1}^{(h)}\bfe_{i,j}]))}{\sum_{k\in\calN_i}\exp(\bw_1^{\mathsf{T}}\rho([\btheta_{c,1}^{(h)} \bfc_i,\btheta_{v,1}^{(h)}\bfv_k,\btheta_{e,1}^{(h)}\bfe_{i,k}]))}
$$ 
where the attention coefficients $\bw_1\in\mathbb{R}^{3d}$ and $\btheta_{e,1}^{(h)}\in \mathbb{R}^{d\times d}$ are both learnable weights and $\rho(\cdot)$ refers to the LeakyReLU activation function with negative slope $0.2$. 
In the second round, similary, each variable node attends to its neighbors to get  updated variable node embeddings 
$$\bfv'_j = \frac{1}{H}\sum_{i=1}^{H}\left(\alpha_{jj,2}^{(h)}\btheta_{c,2}^{(h)} \bfc'_i+\sum_{j\in\calN_i}\alpha_{ji,2}^{(h)}\btheta_{v,2}^{(h)}\bfv_j\right)$$ 
with attention weights 
$$ \alpha^{(h)}_{ji,2} = \frac{\exp(\bw_2^{\mathsf{T}}\rho([\btheta_{c,2}^{(h)} \bfc'_i,\btheta_{v,2}^{(h)}\bfv_j,\btheta_{e,2}^{(h)}\bfe_{i,j}]))}{\sum_{k\in\calN_j}\exp(\bw_2^{\mathsf{T}}\rho([\btheta_{c,2}^{(h)} \bfc'_i,\btheta_{v,2}^{(h)}\bfv_j,\btheta_{e,2}^{(h)}\bfe_{i,k}]))}$$ where $\bw_2\in\mathbb{R}^{3d}$ and $\btheta_{c,2}^{(h)},\mathbf{\btheta}_{v,2}^{(h)},\btheta_{e,2}^{(h)}\in \mathbb{R}^{d\times d}$ are learnable weights. 
After the two rounds of message passing, the final representations of variables $\bfv'$ are passed through a 2-layer MLP with 64 hidden units per layer to obtain a scalar value for each variable. 
Finally, we apply the sigmoid  function to get a score  between 0 and 1.

\section{Additional Details of Instance Generation}
We present the ILP formulations for the minimum vertex cover (MVC), maximum independent set (MIS), set covering (SC) and combinatorial auction (CA) problems. 
\subsection{MVC}
In an MVC instance, we are given an undirected graph $G=(V,E)$. The goal is to select the smallest subset of nodes such that at least one end point of every edge in the graph is selected: 

\begin{eqnarray*}
&\min \sum_{v\in V} x_v \\
\textrm{s.t.} &x_u+x_v\geq 1, \,\forall (u,v)\in E, \\
 & x_v\in\{0,1\}, \, \forall v\in V.
\end{eqnarray*}

\subsection{MIS}
In an MIS instance, we are given an undirected graph $G=(V,E)$. The goal is to select the largest subset of nodes such that no two nodes in the subsets are connected by an edge in $G$:
\begin{eqnarray*}
&\min -\sum_{v\in V} x_v \\
\textrm{s.t.} &x_u+x_v\leq 1, \,\forall (u,v)\in E, \\
 & x_v\in\{0,1\}, \, \forall v\in V.
\end{eqnarray*}
\subsection{SC}
In an SC instance, we are given $m$ elements and a collection $S$ of $n$ sets whose union is the set of all elements. The goal is to select a minimum number of sets from $S$ such that the union of the selected set is still the set of all elements:

\begin{eqnarray*}
&\min \sum_{s\in S} x_s \\
\textrm{s.t.} &\sum_{s\in S: i\in s} x_s \geq 1, \,\forall i\in[m], \\
 & x_s\in\{0,1\}, \, \forall s\in S.
\end{eqnarray*}

\subsection{CA}
In a CA instance, we are given $n$ bids $\{(B_i,p_i):i\in[n]\}$ for $m$ items, where  $B_i$ is a subset of items and $p_i$ is its associated bidding price. The objective is to allocate items to bids such that the total revenue is maximized:
\begin{eqnarray*}
&\min -\sum_{i\in[n]} p_ix_i \\
\textrm{s.t.} &\sum_{i: j\in B_i} x_i \leq 1, \,\forall j\in[m], \\
 & x_i\in\{0,1\}, \, \forall i\in[n].
\end{eqnarray*}

\section{Additional Details on Hyperparameter Tuning}

For \RL, we use all the hyperparameters provided in their code \cite{wu2021learning} in our experiments. For the other LNS methods, all hyperparameters used in experiments are fine-tuned on the validation set and the hyperparameter tunings are described in the following. 

For $\beta$ which upper bounds the neighborhood size, we tried values from $\{0.25, 0.5, 0.6, 0.7\}$. $\beta=0.25$ is the worst for all approaches, resulting in the highest gap. For \LBRELAX, \DM and \CL, all values perform similarly (because they select effective neighborhoods early in the search and their neighborhood sizes either do not reach the upper bound or they already converge to good solutions before reaching it). For \RANDOM and \GRAPH, $\beta=0.5$ is the best for them. So we set $\beta=0.5$ consistently for all approaches.

For initial neighborhood sizes $k^0$, we observe that the best values are sensitive for approaches that need longer runtime to select variables, such as \LBRELAX, \DM and \CL, thus they need the right $k^0$ from the beginning and we fine tune it for them. For \RANDOM and \GRAPH, their runtime for selecting variables is short, and with the adaptive neighborhood size mechanism, they could very quickly find the right neighborhood size and are insensitive to $k^0$. They converge to the same primal gaps ($<1\%$ relative differences) with similar primal integrals ($<2\%$ relative differences) using different $k^0$. Despite that the differences are small, we still use the best $k^0$ for them.  

For $\gamma$ that controls the rate at which $k^t$ increases, we tried values from $\{1, 1.01, 1.02, 1.05\}$. Overall, $\gamma$ does not have a big impact on the performance if $\gamma>1$, however $\gamma=1$ is far worse than the others. 

For the runtime limit for each repair operation, we tried different limits of 0.5, 1, 2 and 5 minutes. All approaches are not sensitive to it since most repairs are finished within 20 seconds. Except for \DM on the SC instances, it selects neighborhoods that require a longer time to repair and a 2-minute runtime limit is necessary. We therefore use 2 minutes consistently.

For BnB, the aggressive mode is fune-tuned for each problem on the validation set. With the aggressive mode turned on, BnB (SCIP) does not always deliver better anytime performance than having it turned off.  Based on the validation results, the aggressive mode is turned on for MVC and SC instances and turned off for CAT and MIS instances.

For \DM, it uses the same training dataset as \CL but uses only the positive samples. We  fine tune its hyperparameters for each problem on the validation set, resulting in a different $k^0$ on the SC instance from \CL.  Also in \citet{sonnerat2021learning}, they use sampling methods to select variables when using the learned policy. For the temperature parameter $\eta$ in the sampling method, we tried values from $\{1/2, 2/3, 1\}$ and $\eta=0.5$ performs the best overall. However, in our experiment, we observe that our greedy method described in Section \ref{sec::usingthepolicy} works better for \DM on SC and MIS instances, thus, \CL is compared against the corresponding results on SC and MIS instances.

For \LBRELAX, there are three variants of it presented in \citet{huang2022local}. We present only the best of the three variants for each problem in the paper for simplicity. 

In Table \ref{table::sumHyperparameter}, we summarize all the hyperparameters with their notations and values used in our experiments. 

\begin{table*}[]
\centering
\small
\caption{Hyperparameters with their notations and values used. \label{table::sumHyperparameter}}
\begin{tabular}{lcc}
Hyperparameter & Notation              & Value \\ \hline
 Suboptimality threshold to determine positive samples & $\alpha_{\sfp}$ & 0.5\\
 Upper bound on the number of positive samples & $u_{\sfp}$ & 10 \\
 Suboptimality threshold to determine negative samples & $\alpha_{\sfn}$ & 0.05\\
 Ratio between the numbers of
positive and negative samples & $\kappa$ & 9 \\
 Feature embedding dimension & $d$              & 64    \\
                Window size of the most recent incumbent values in variable features           &  &     3  \\
   Number of attention heads in the GAT  & $H$    &  8 \\
   Temperature parameter in the contrastive loss & $\tau$ & 0.07 \\
    Rate at which $k^t$ increases & $\gamma$ & 1.02 \\
   Upper bound on $k^t$ as a fraction of number of variables & $\beta$ & 0.5 \\
    Temperature parameter for sampling variables in \DM & $\eta$ & 0.5 \\
   Initial neighborhood size &$k^0$  & Fine-tuned for each case \\
    Runtime for finding initial solution & & 10 seconds \\
    Runtime limit for each reoptimization & & 2 minutes \\
    Learning rate (\CL and \DM) & & $10^{-3}$   \\
    Batch size (\CL and \DM) & &32 \\
    Number of training epochs (\CL and \DM) & &30\\
    \hline
\end{tabular}
\end{table*}

\section{Additional Experimental Results}

In this section, we add two more baselines and evaluate all approaches on one more metric. We show that \CL outperforms all approaches in terms of all metrics.

We establish two addtional baselines:
\begin{itemize}
    \item LB: LNS which selects the neighborhood with the LB heuristics. We set the time limit to 10 minutes for solving the LB ILP in each iteration;
    \item \GRAPH: LNS which selects the neighborhood based on the bipartite graph representation of the ILP similar to GINS \cite{maher2017scip}. A bipartite graph representation consists of nodes representing the variables and constraints on two sides, respectively, with an edge connecting a variable and a constraint if a variable has a non-zero coefficient in the constraint.
    It runs a breadth-first search starting from a random variable node in the bipartite graph and selects the first $k^t$ variable nodes expanded. 
\end{itemize}

\begin{figure*}[btp]
    \centering
    \includegraphics[width=\textwidth]{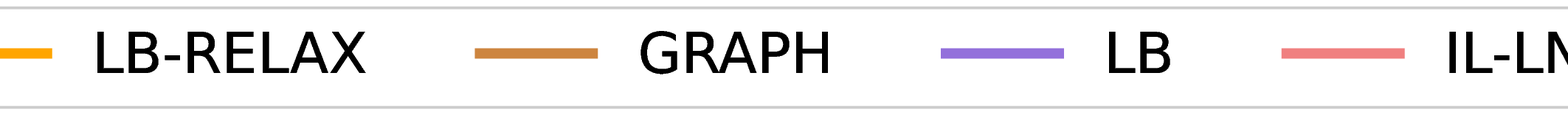}

    \subfloat[MVC-S (left) and MVC-L (right).]
    {
        \includegraphics[width=0.24\textwidth]{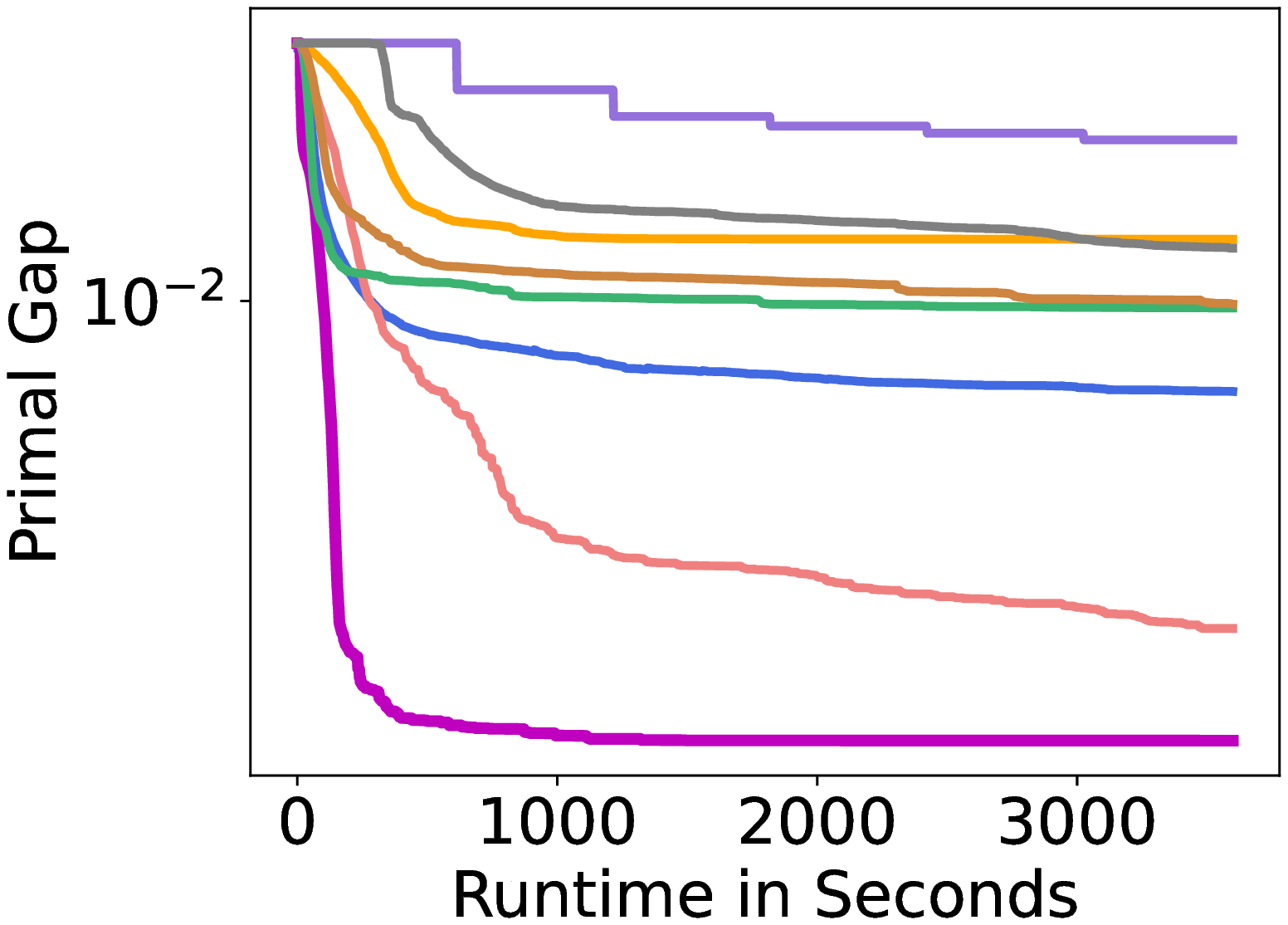}
        \includegraphics[width=0.24\textwidth]{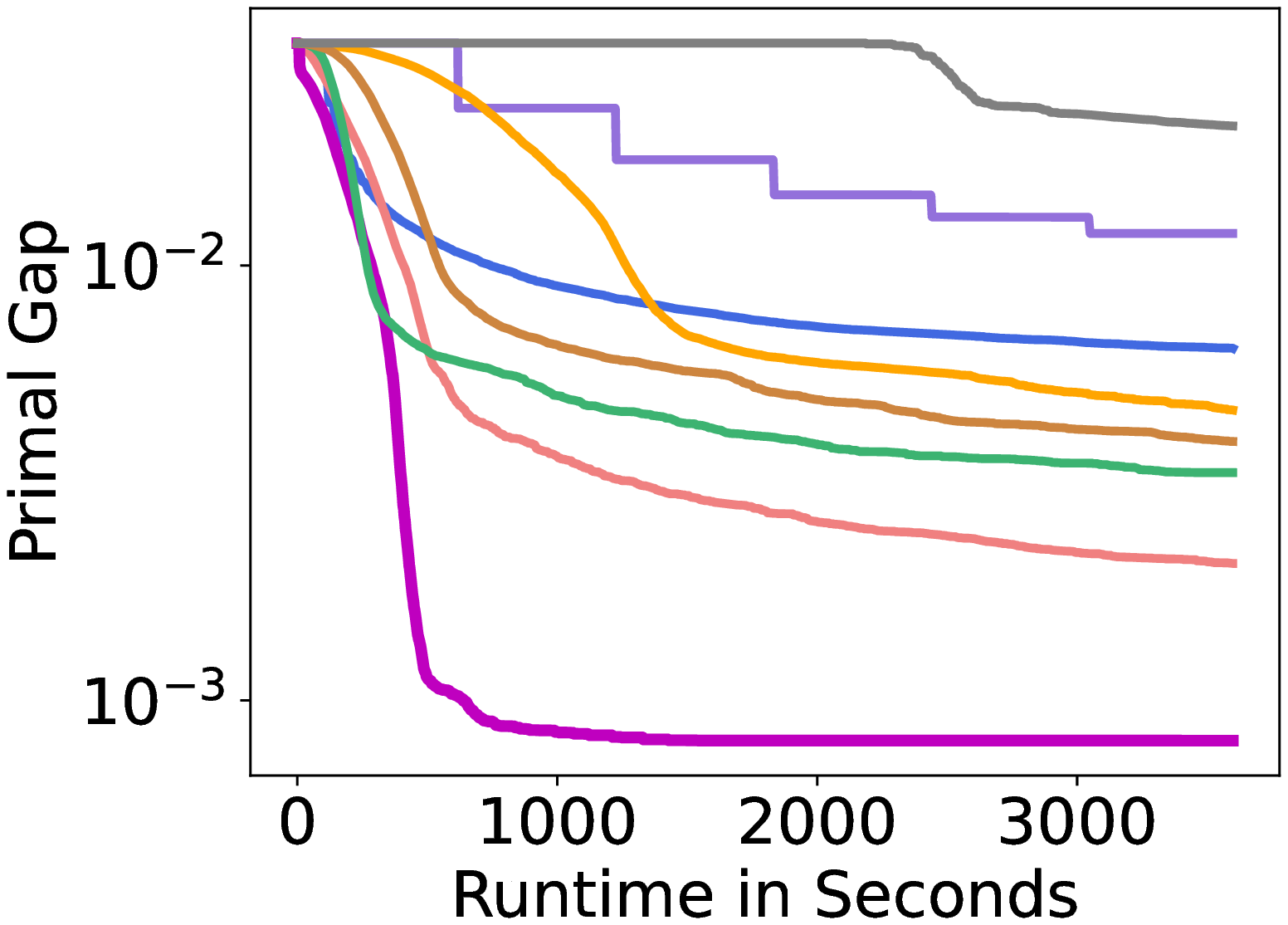}    
    }
    \subfloat[MIS-S (left) and MIS-L (right).]
    {
        \includegraphics[width=0.24\textwidth]{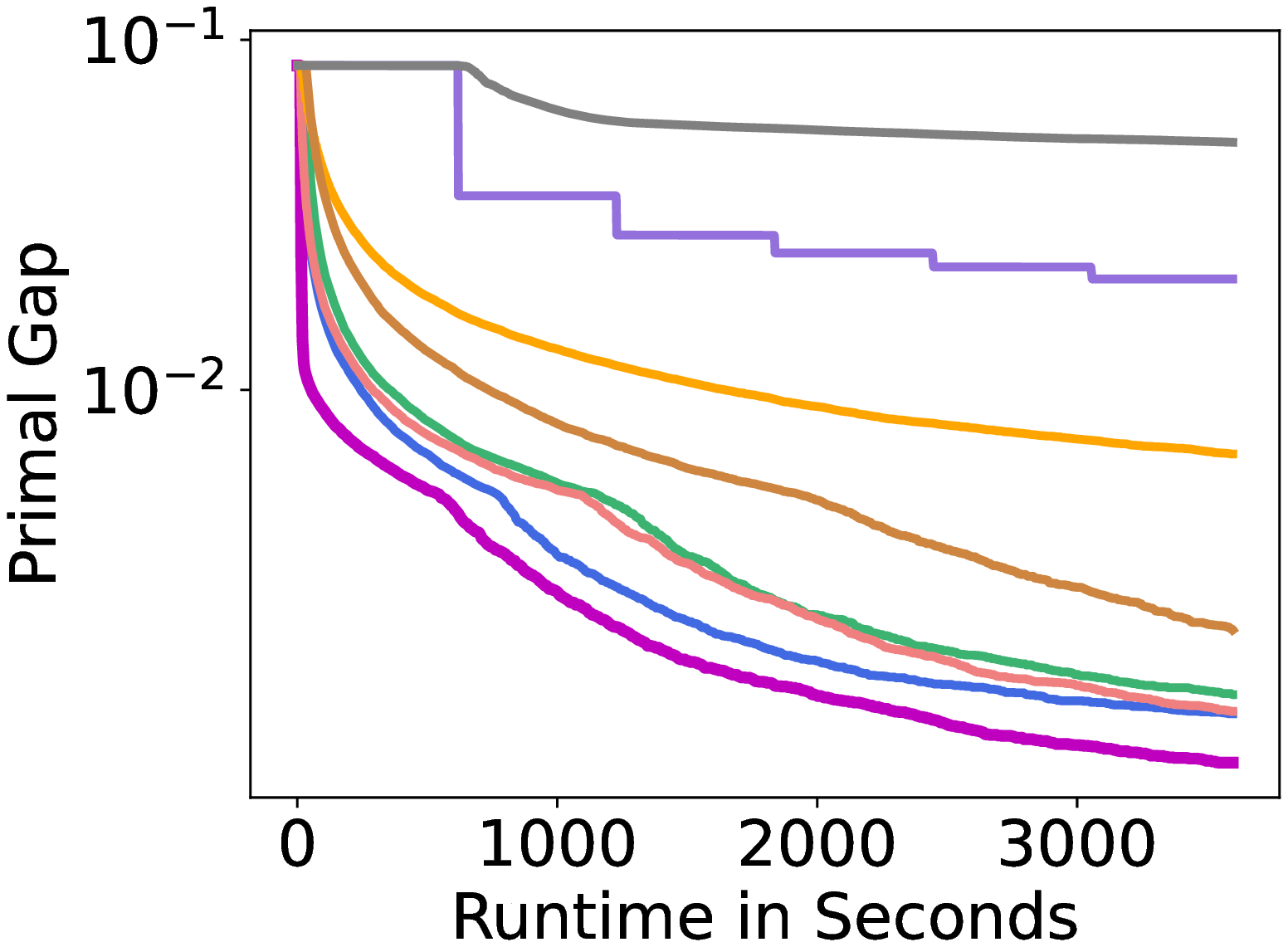}
        \includegraphics[width=0.24\textwidth]{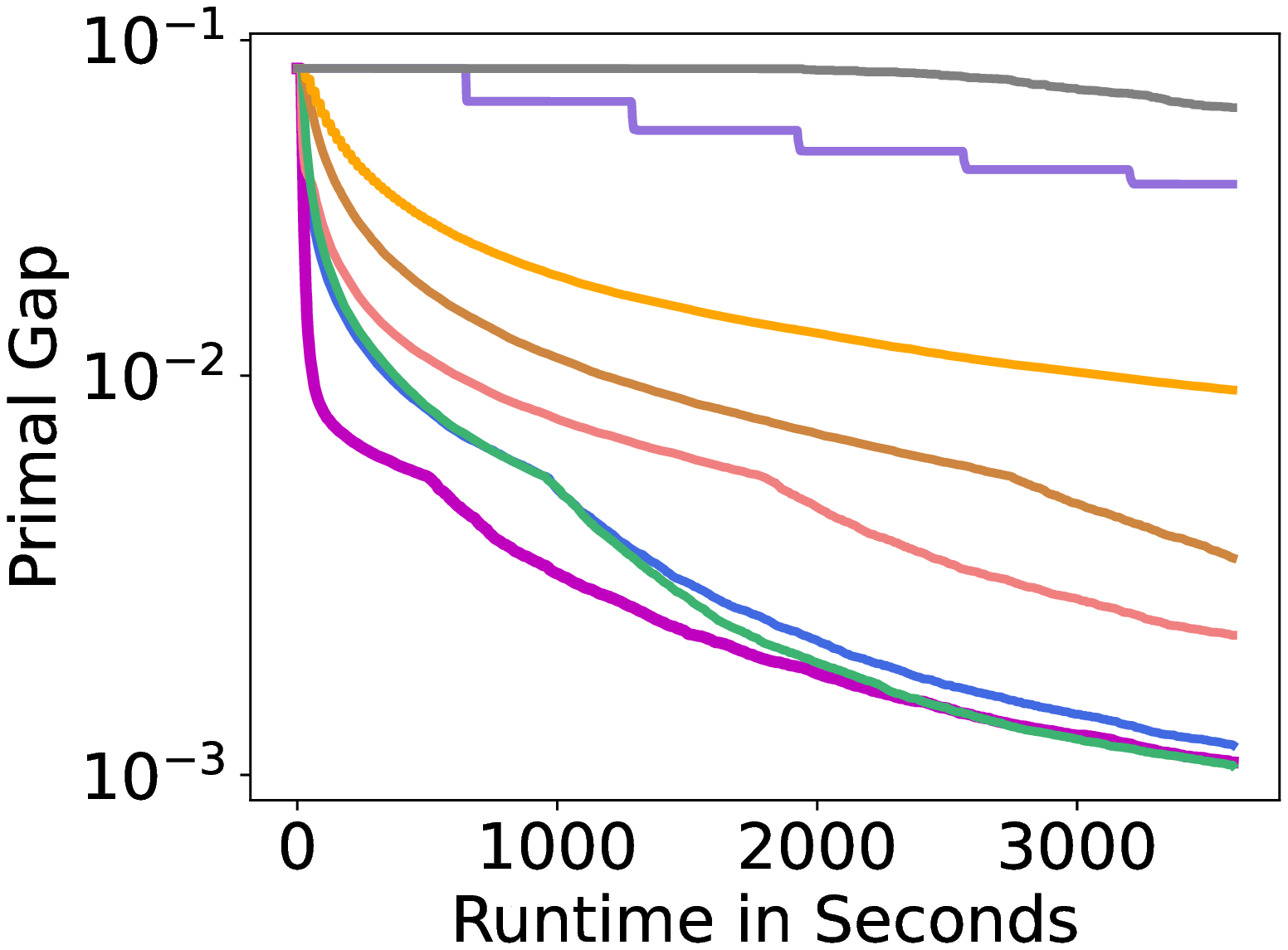}    
    }\\
    \subfloat[CA-S (left) and CA-L (right).]
    {
        \includegraphics[width=0.24\textwidth]{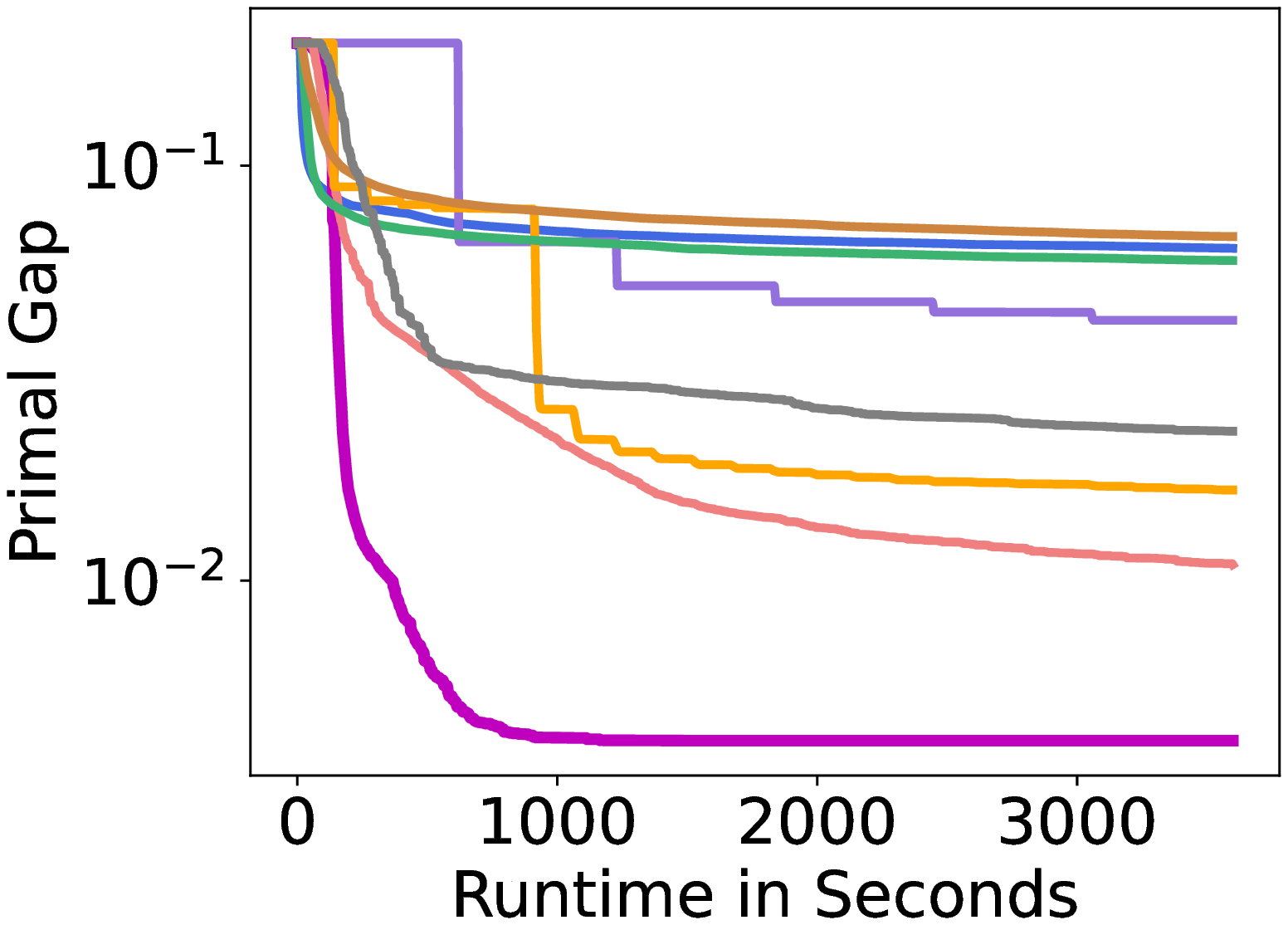}
        \includegraphics[width=0.24\textwidth]{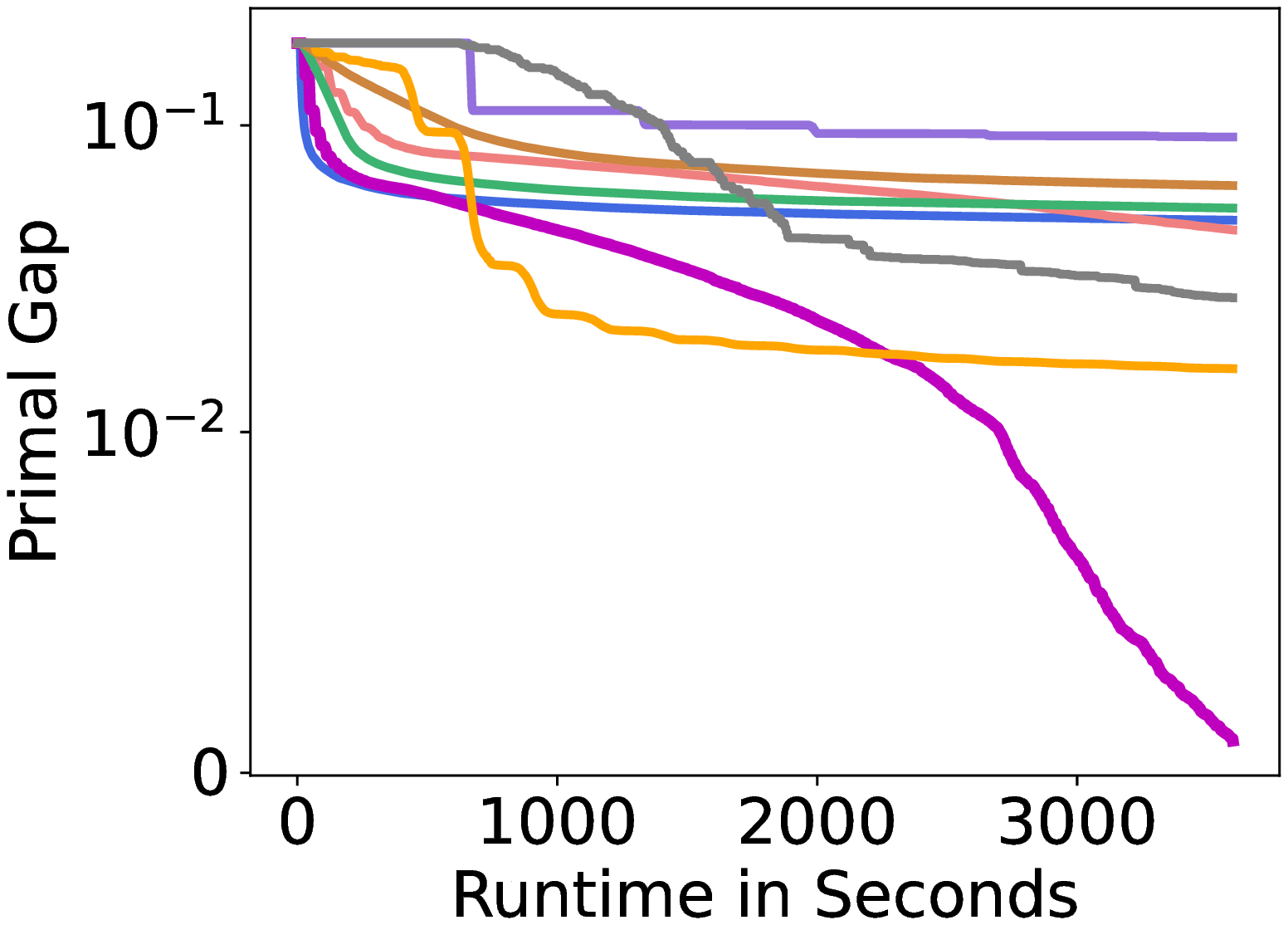}    
    }
    \subfloat[SC-S (left) and SC-L (right).]
    {
        \includegraphics[width=0.24\textwidth]{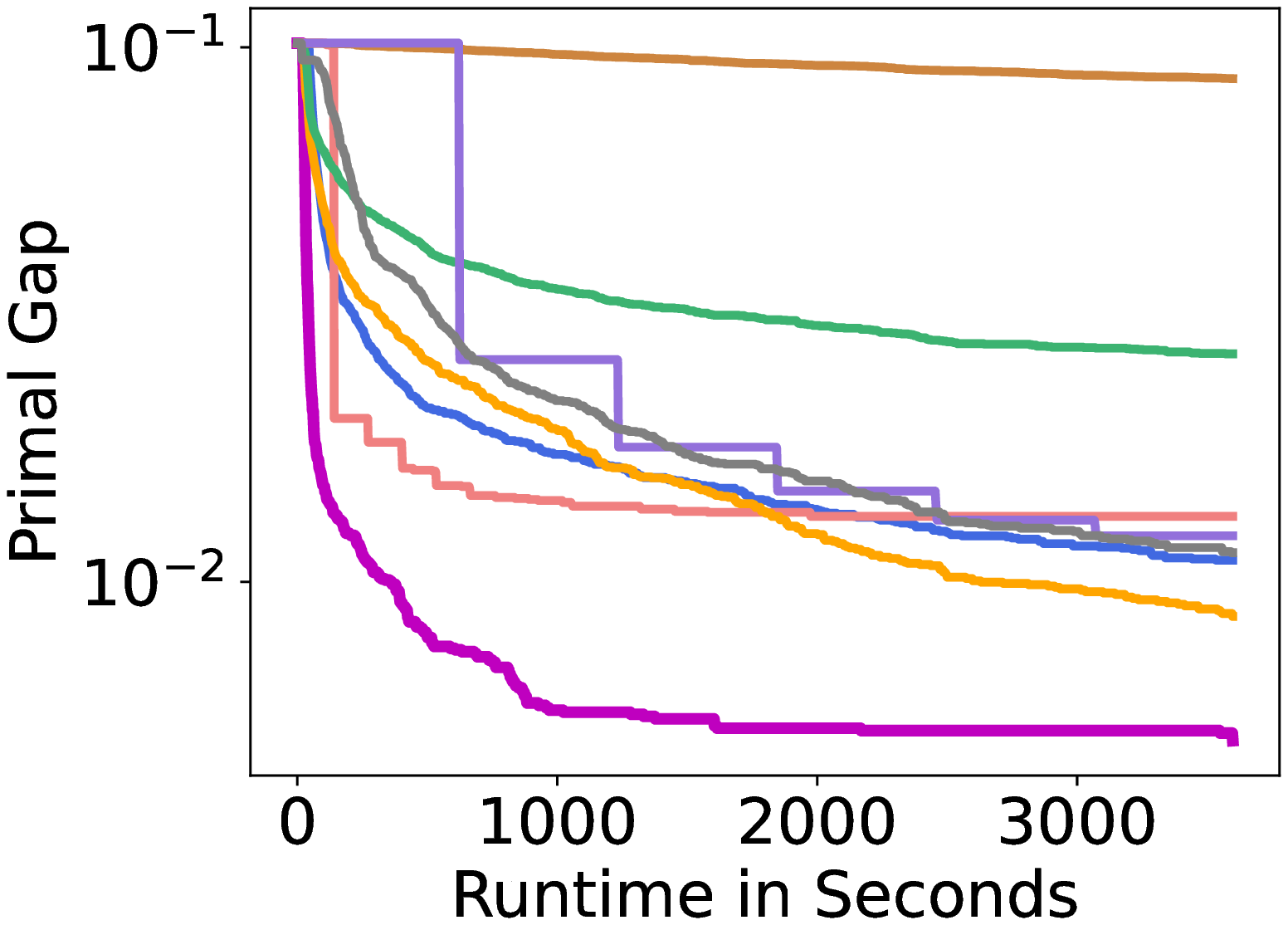}
        \includegraphics[width=0.24\textwidth]{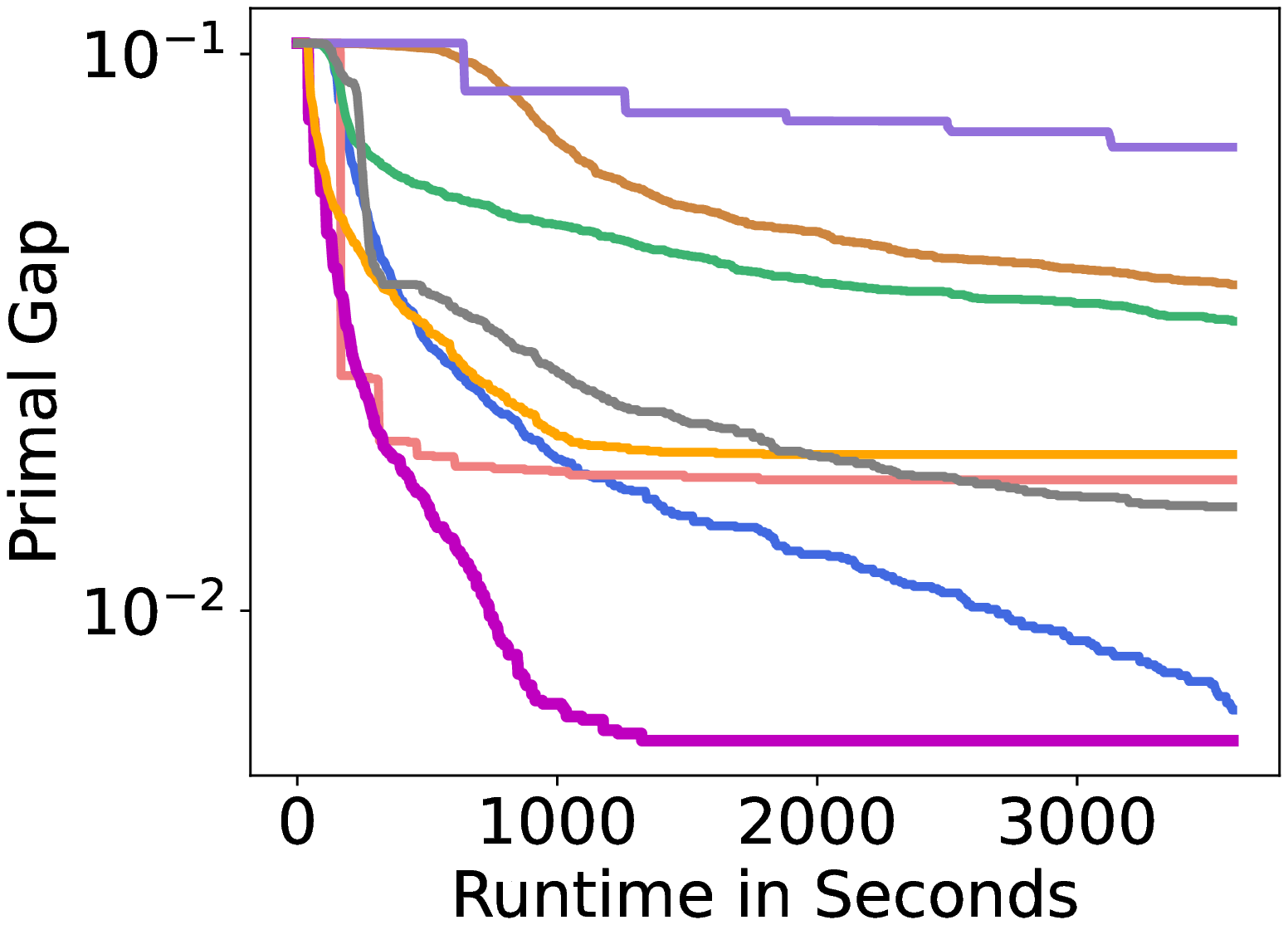}    
    }
    \caption{The primal gap (the lower the better) as a function of time, averaged over 100 instances. For ML approaches, the policies are trained on only small training instances but tested on both small and large test instances.\label{res_all::gap}}
\end{figure*}

\begin{figure*}[btp]
    \centering
    \includegraphics[width=\textwidth]{figure_all/legend_ML_horizontal_timeVSobj_3600.eps}

    \subfloat[MVC-S (left) and MVC-L (right).]
    {
        \includegraphics[width=0.24\textwidth]{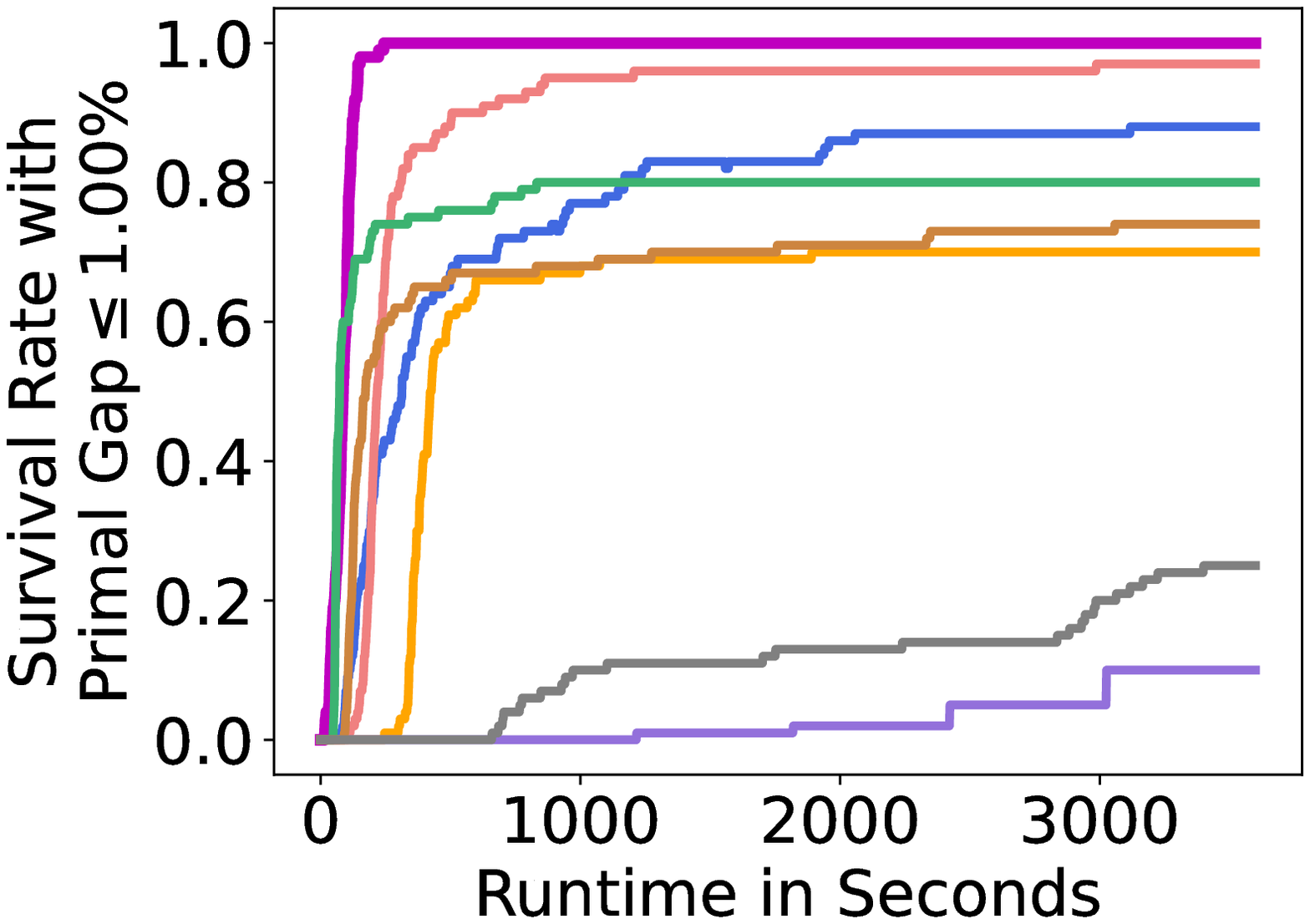}
        \includegraphics[width=0.24\textwidth]{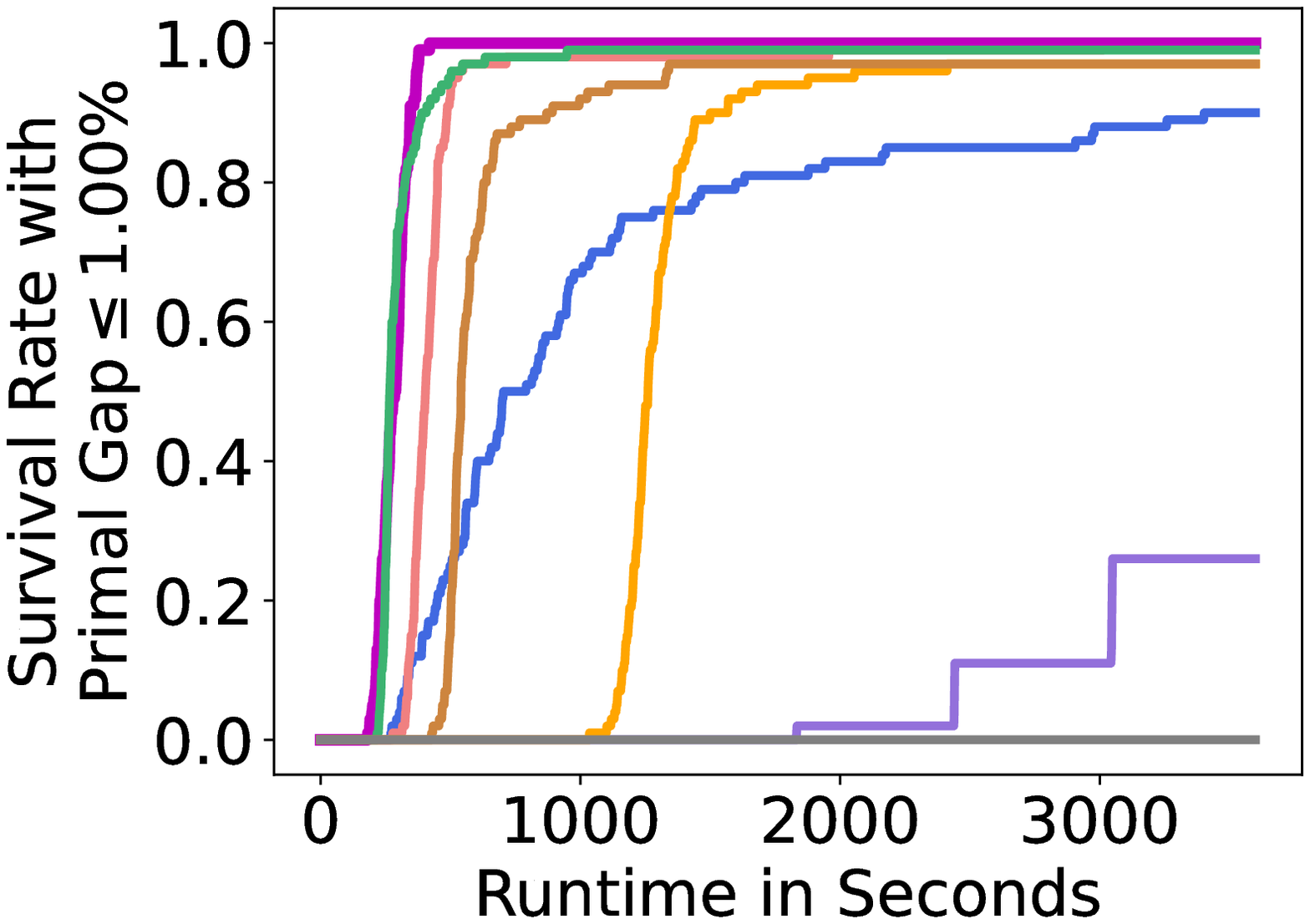}    
    }
    \subfloat[MIS-S (left) and MIS-L (right).]
    {
        \includegraphics[width=0.24\textwidth]{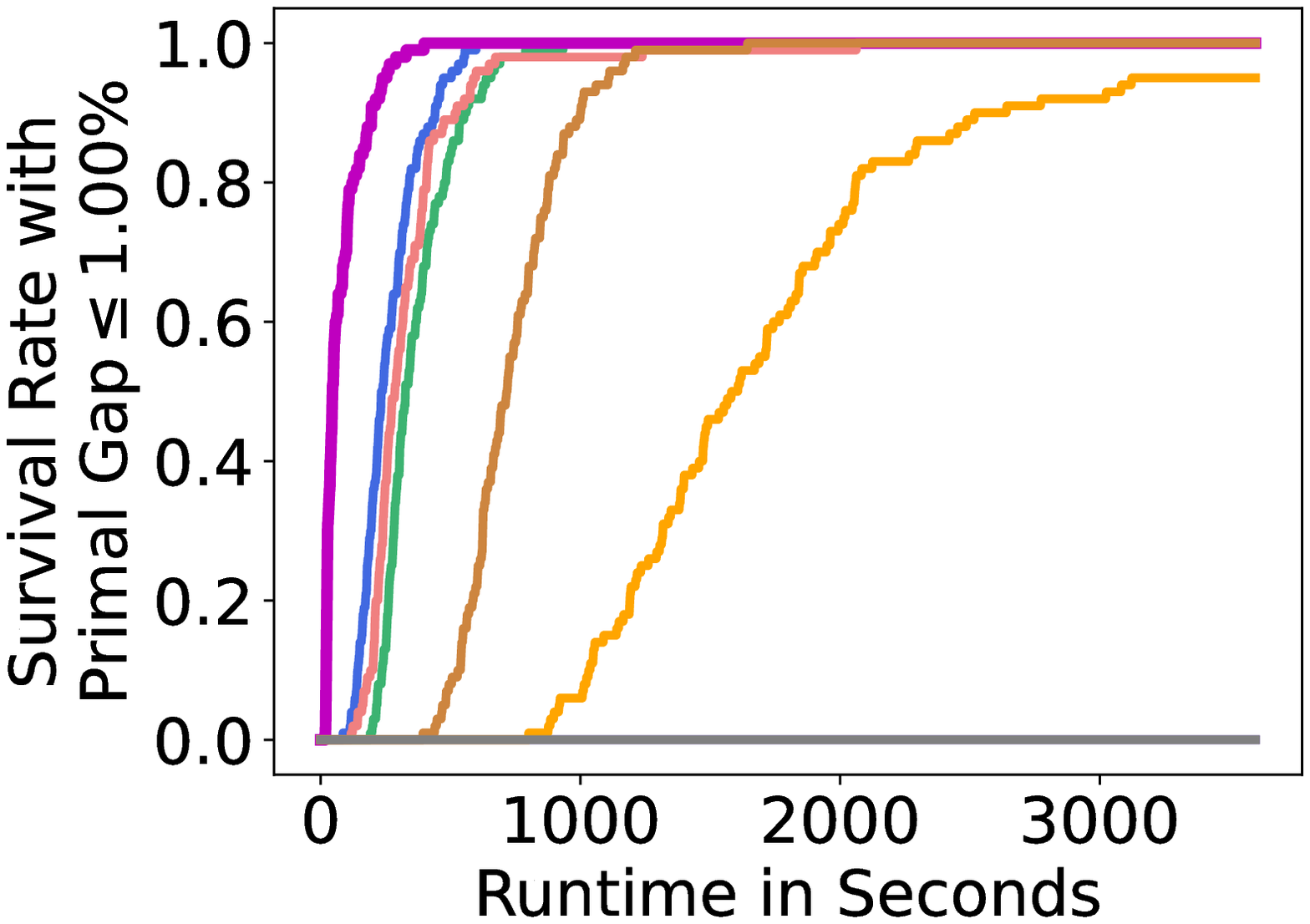}
        \includegraphics[width=0.24\textwidth]{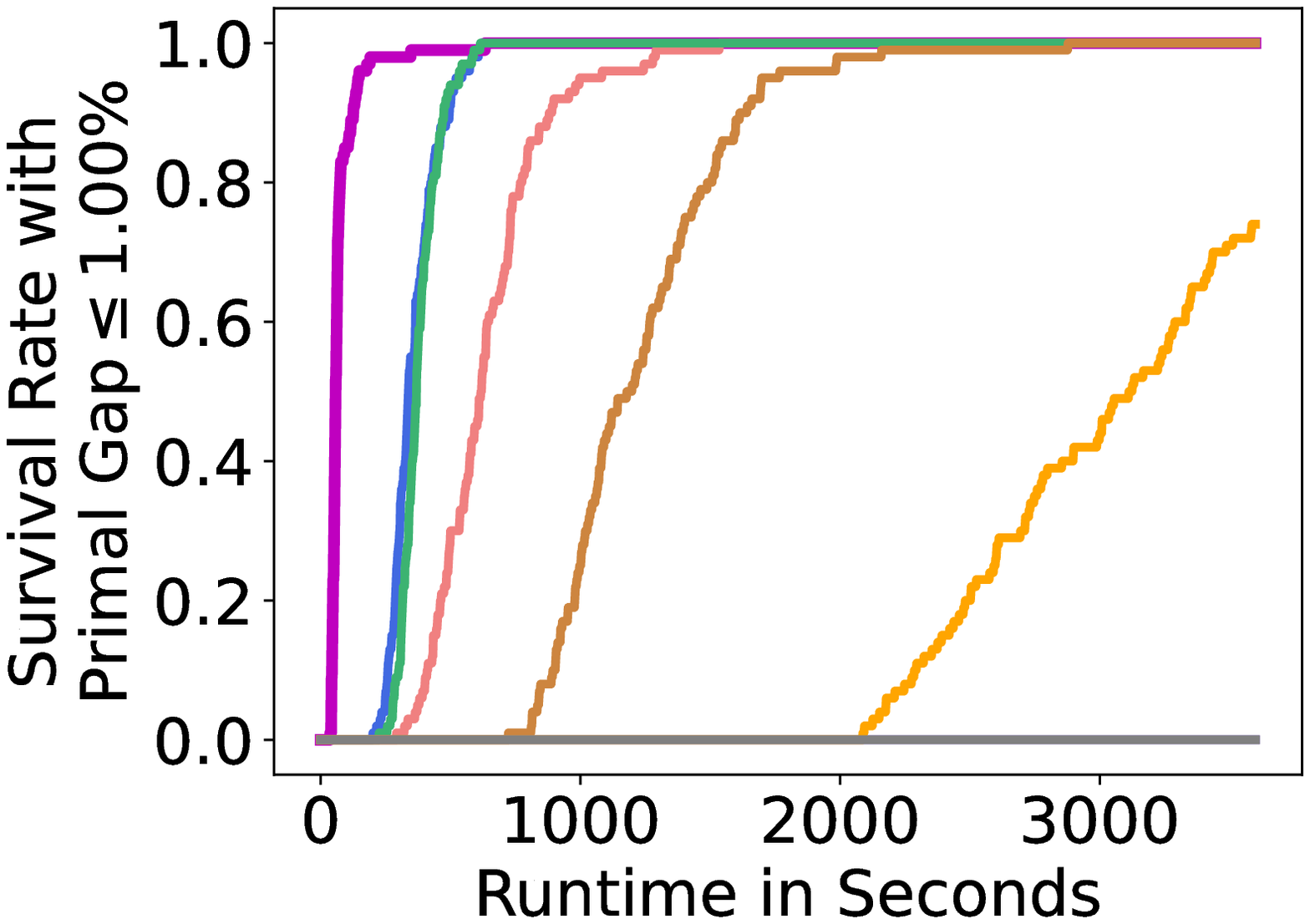}    
    }\\
    \subfloat[CA-S (left) and CA-L (right).]
    {
        \includegraphics[width=0.24\textwidth]{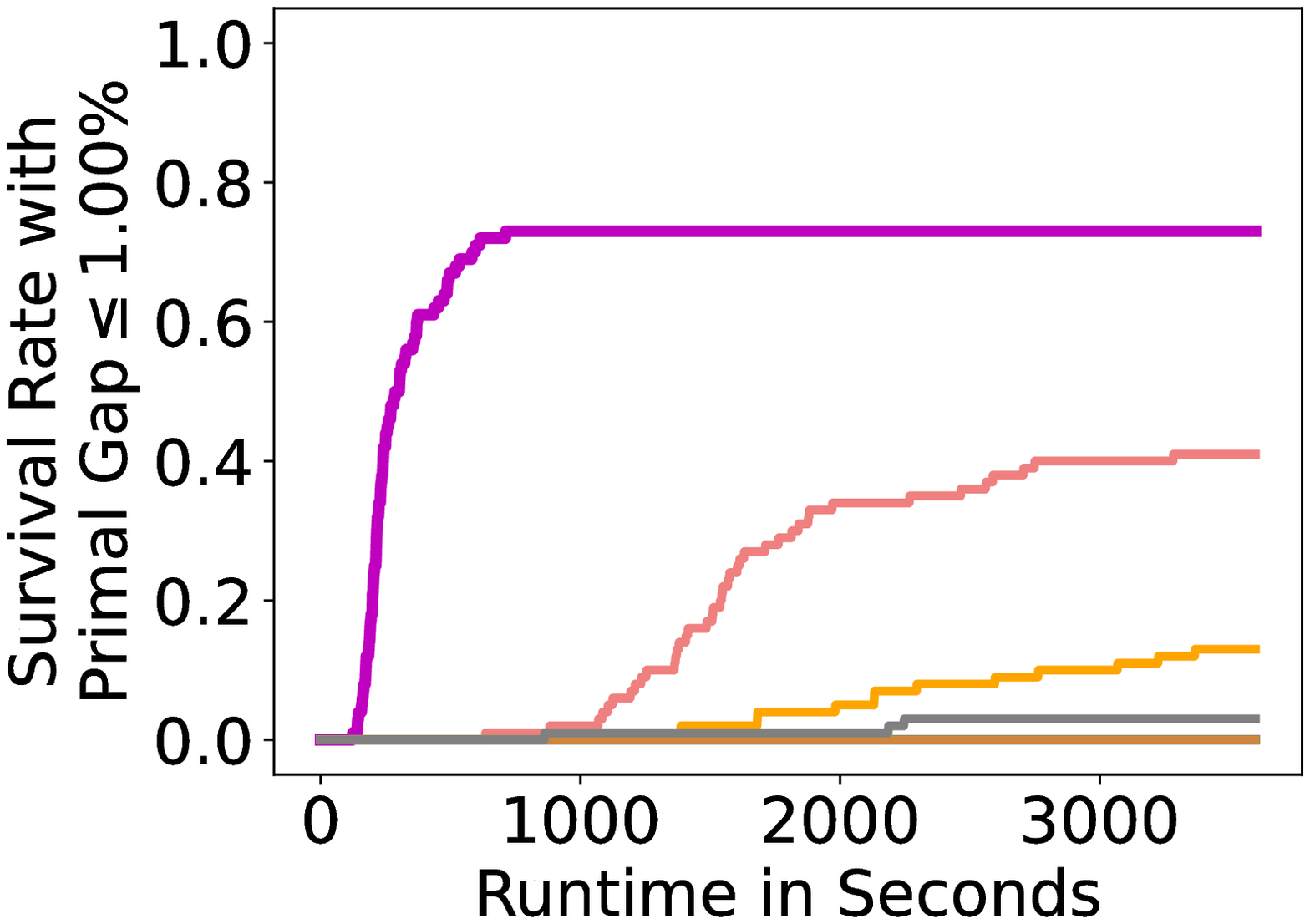}
        \includegraphics[width=0.24\textwidth]{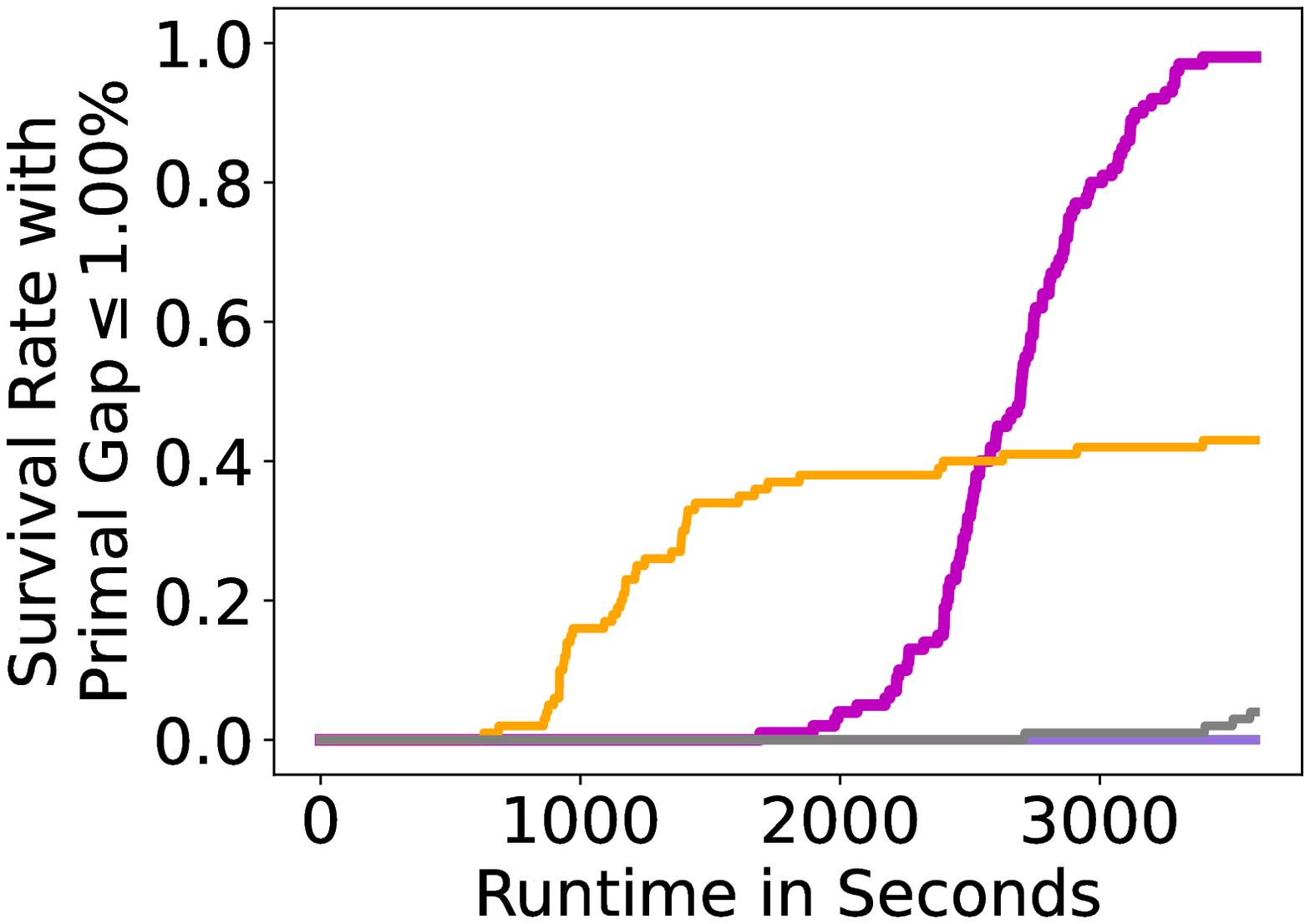}    
    }
    \subfloat[SC-S (left) and SC-L (right).]
    {
        \includegraphics[width=0.24\textwidth]{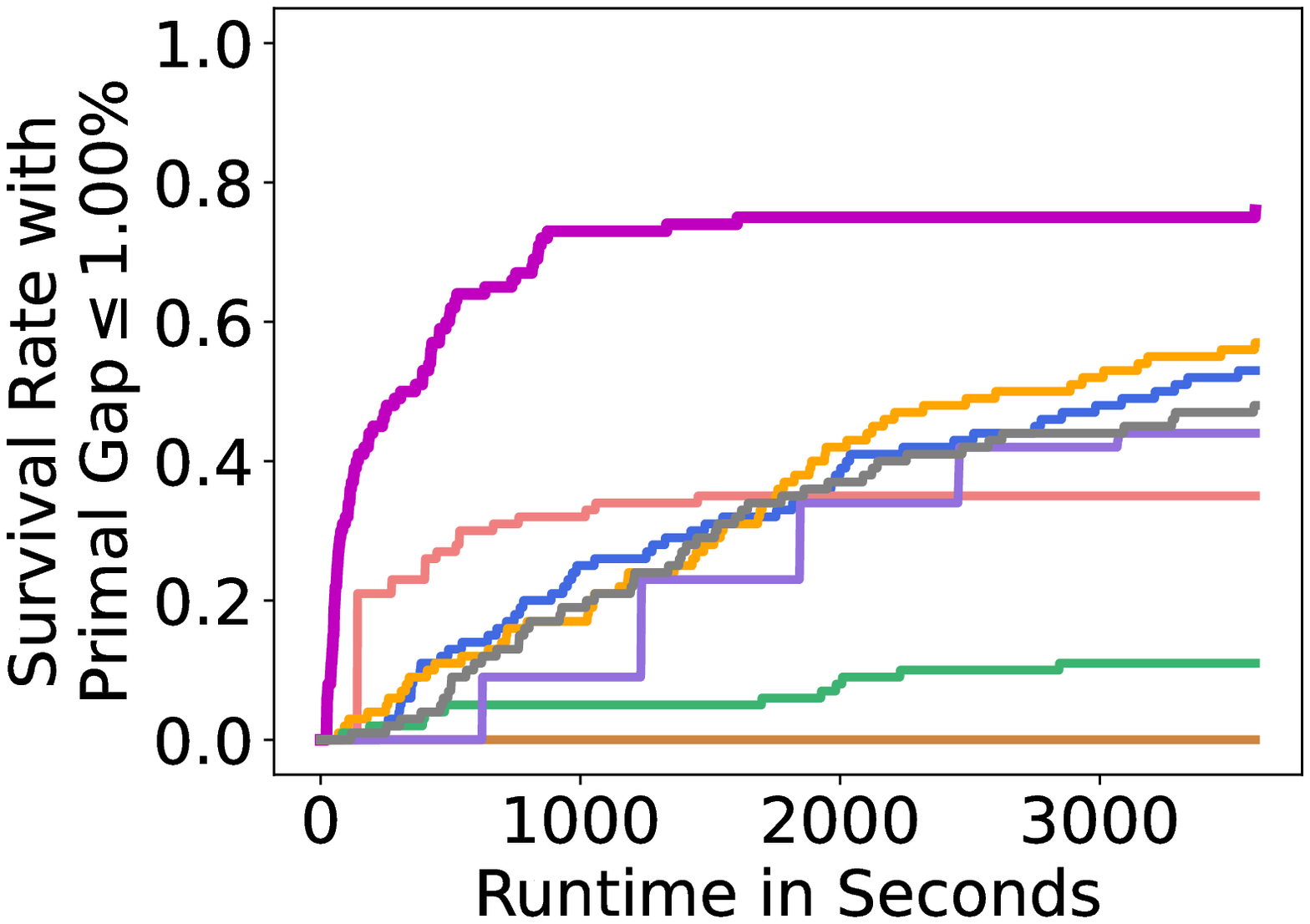}
        \includegraphics[width=0.24\textwidth]{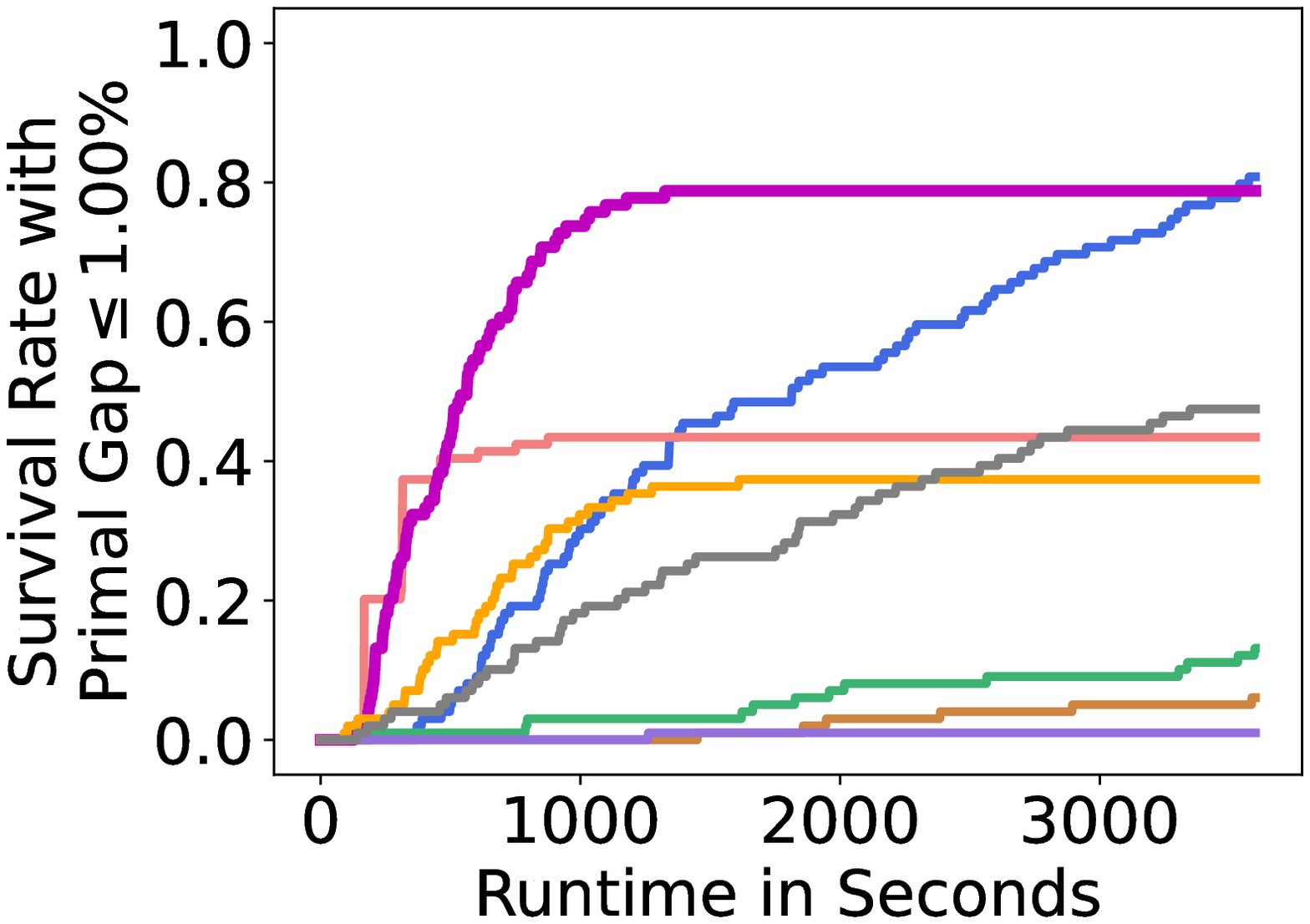}    
    }
    \caption{The survival rate (the higher the better) over 100 instances as a function of time to meet primal gap threshold 1.00\%. For ML approaches, the policies are trained on only small training instances but tested on both small and large test instances.\label{res_all::survival}}
\end{figure*}

\begin{figure*}[tbp]
    \centering
    \includegraphics[width=\textwidth]{figure_all/legend_ML_horizontal_timeVSobj_3600.eps}

    \subfloat[MVC-S (left) and MVC-L (right).]
    {
        \includegraphics[width=0.24\textwidth]{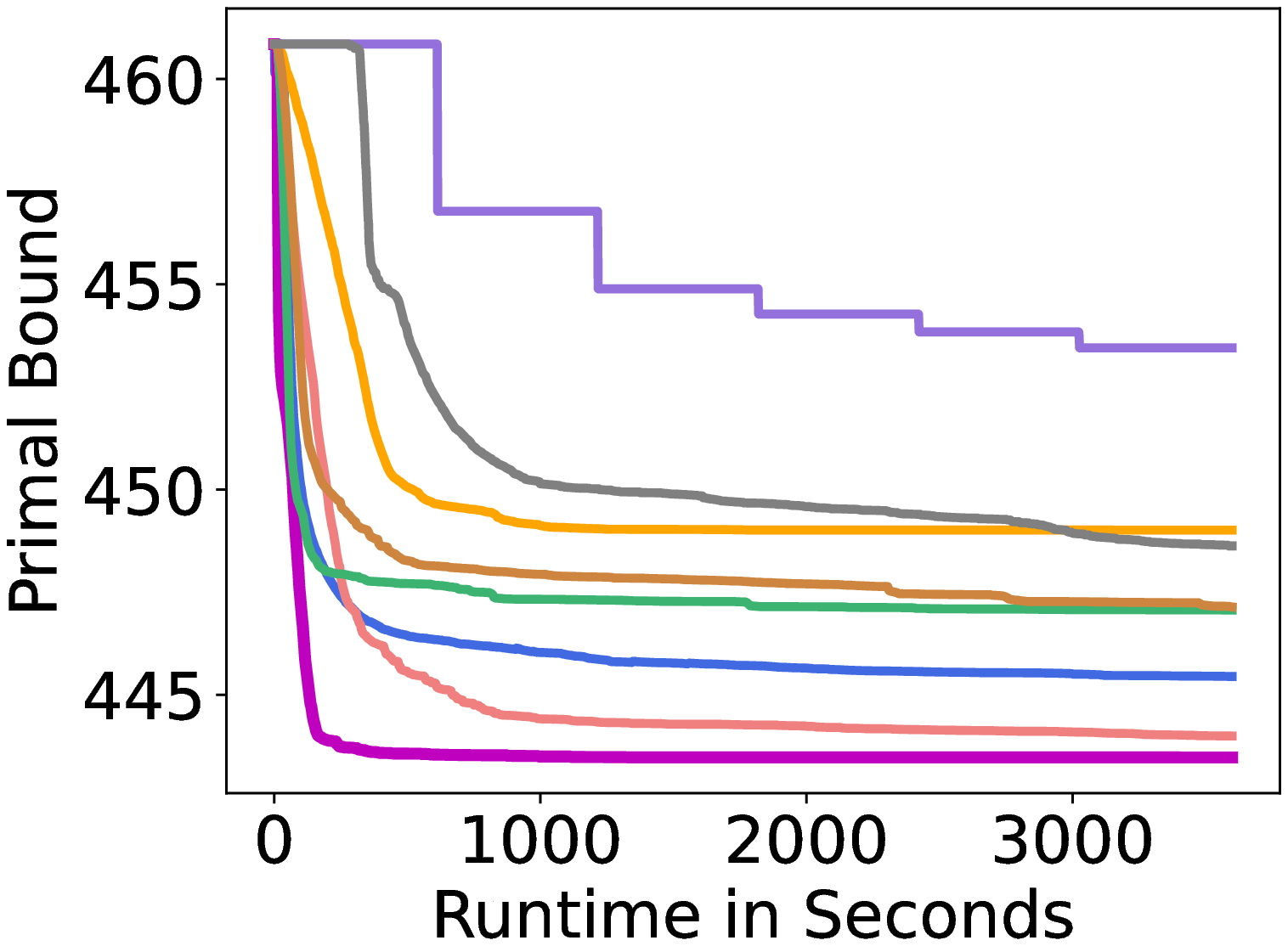}
        \includegraphics[width=0.24\textwidth]{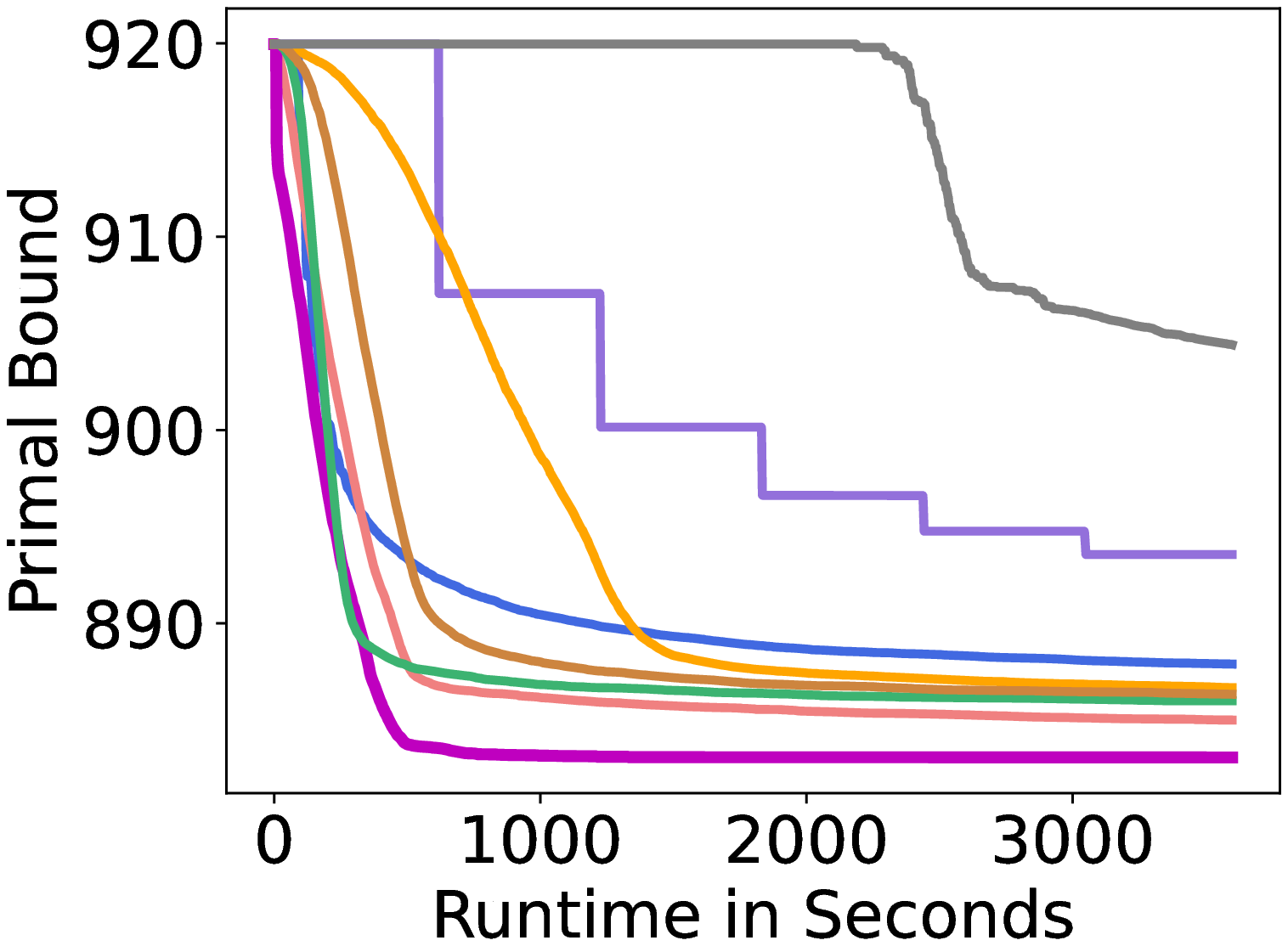}    
    }
    \subfloat[MIS-S (left) and MIS-L (right).]
    {
        \includegraphics[width=0.24\textwidth]{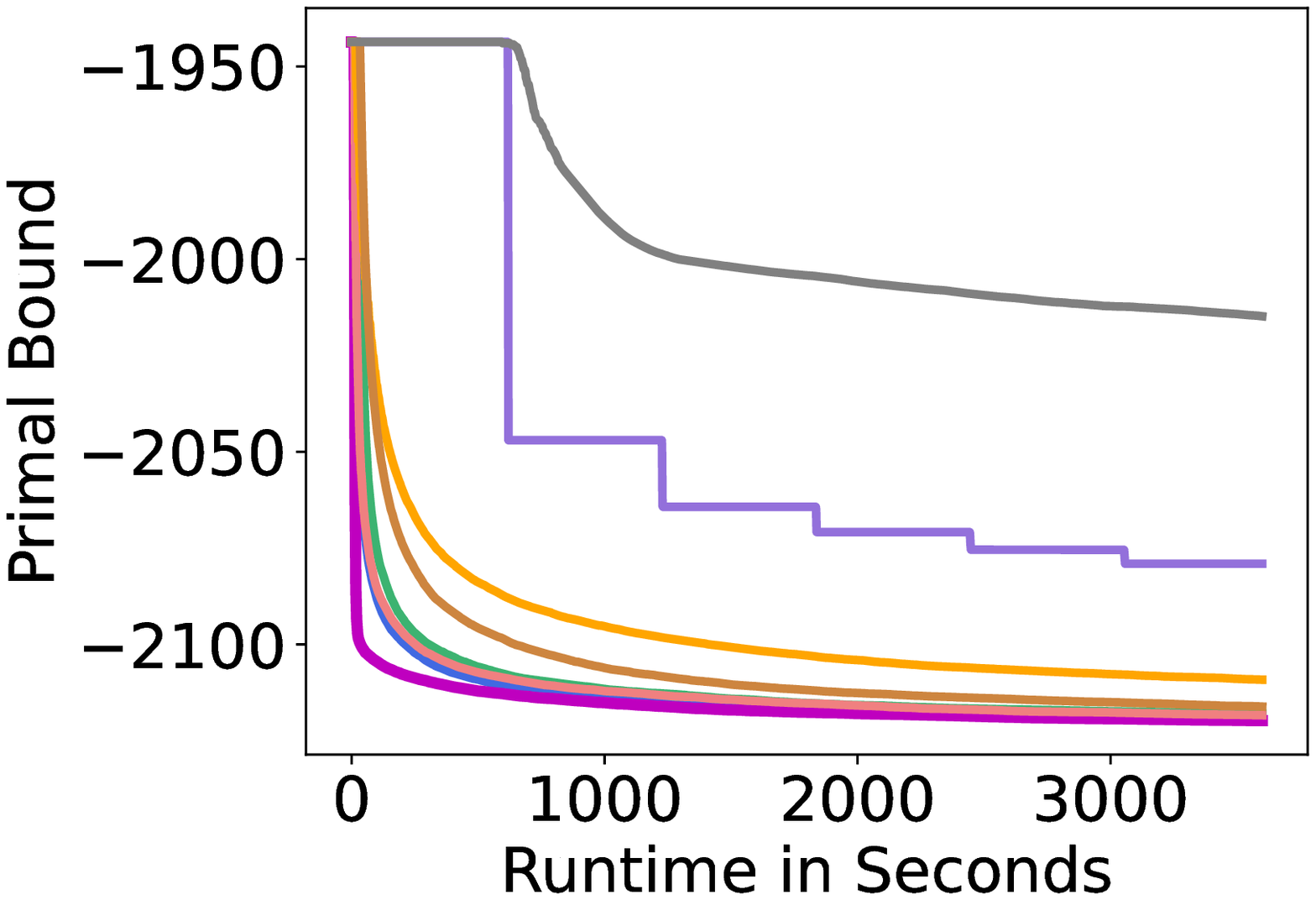}
        \includegraphics[width=0.24\textwidth]{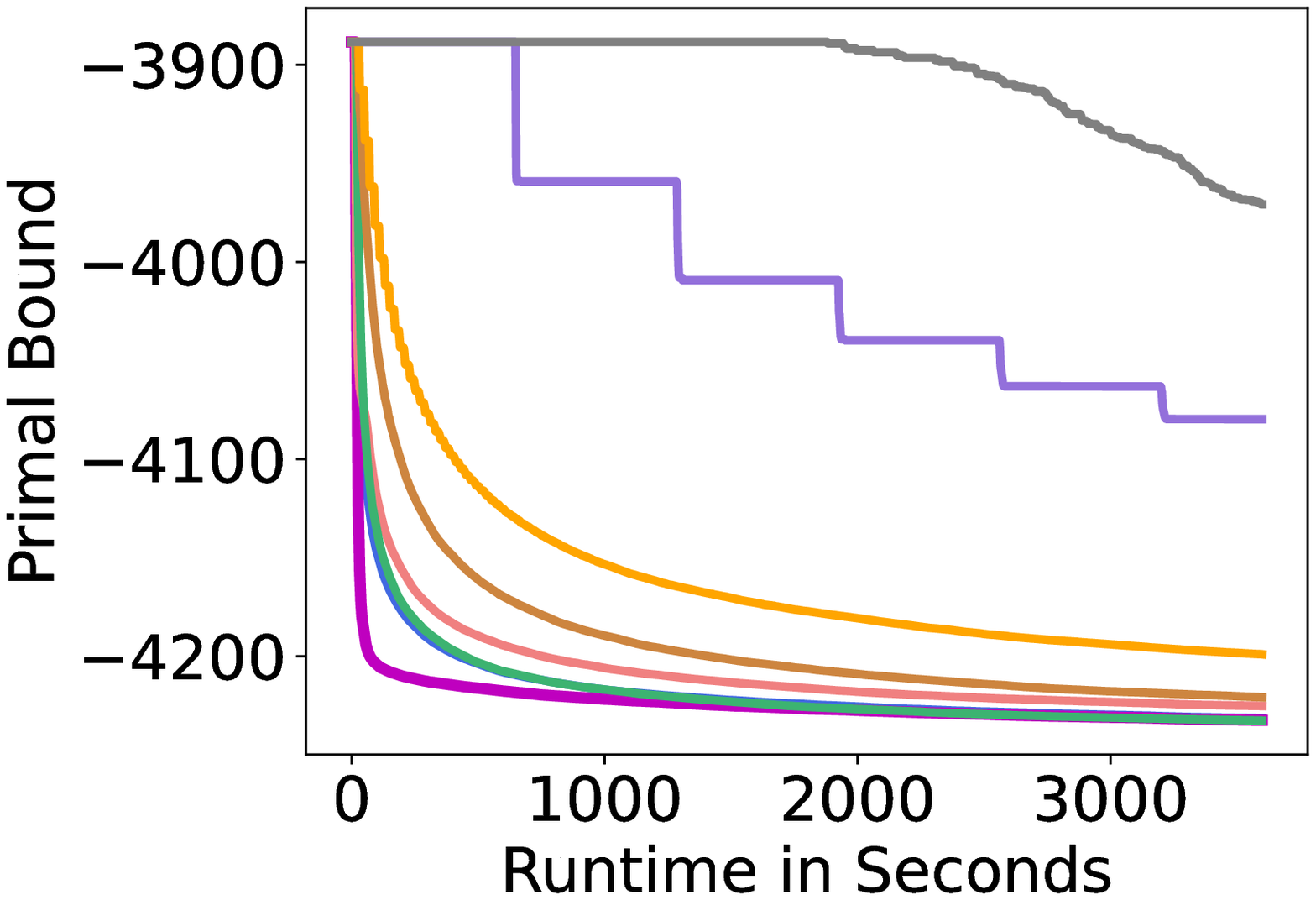}    
    }\\
    \subfloat[CA-S (left) and CA-L (right).]
    {
        \includegraphics[width=0.24\textwidth]{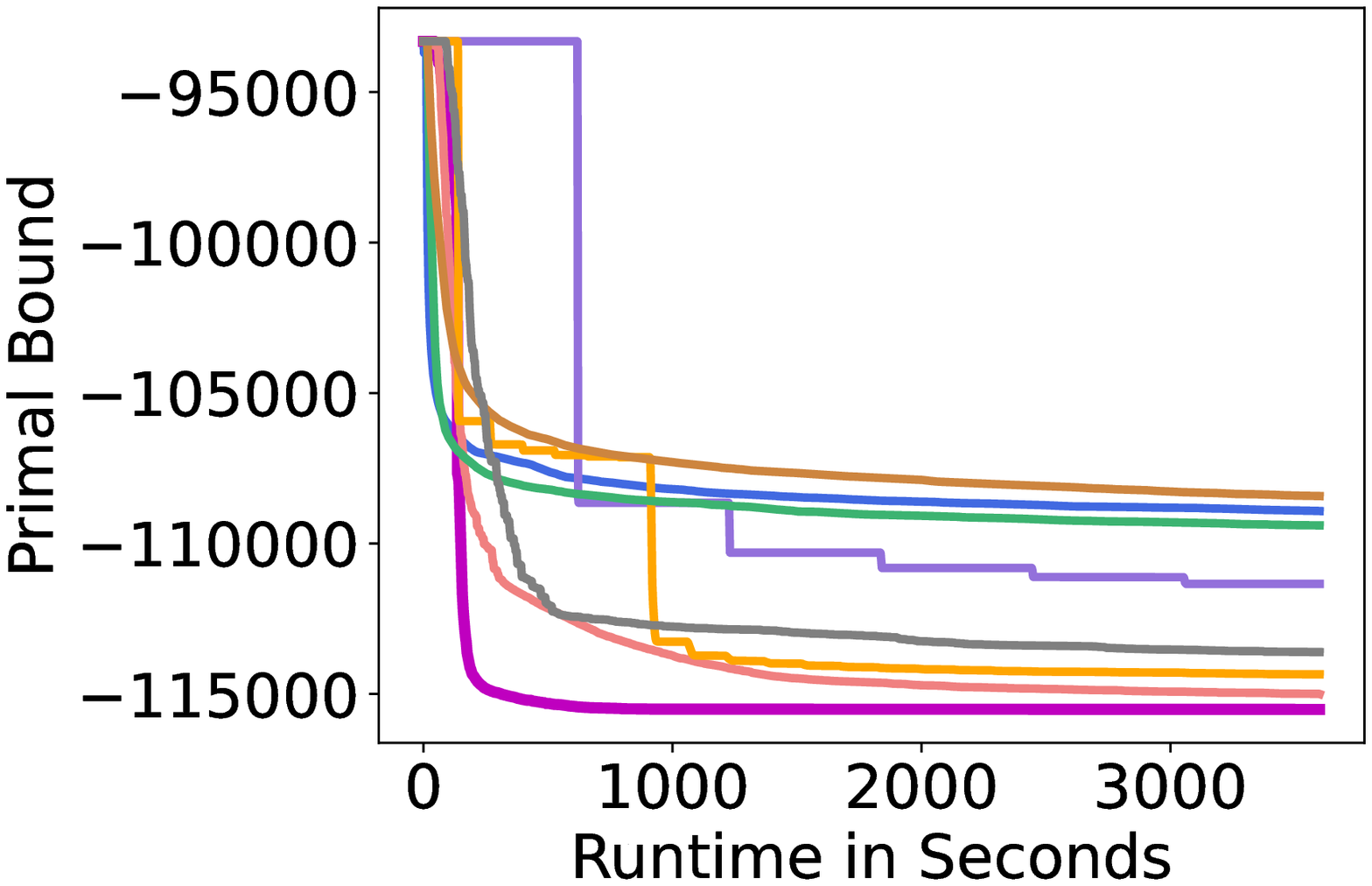}
        \includegraphics[width=0.24\textwidth]{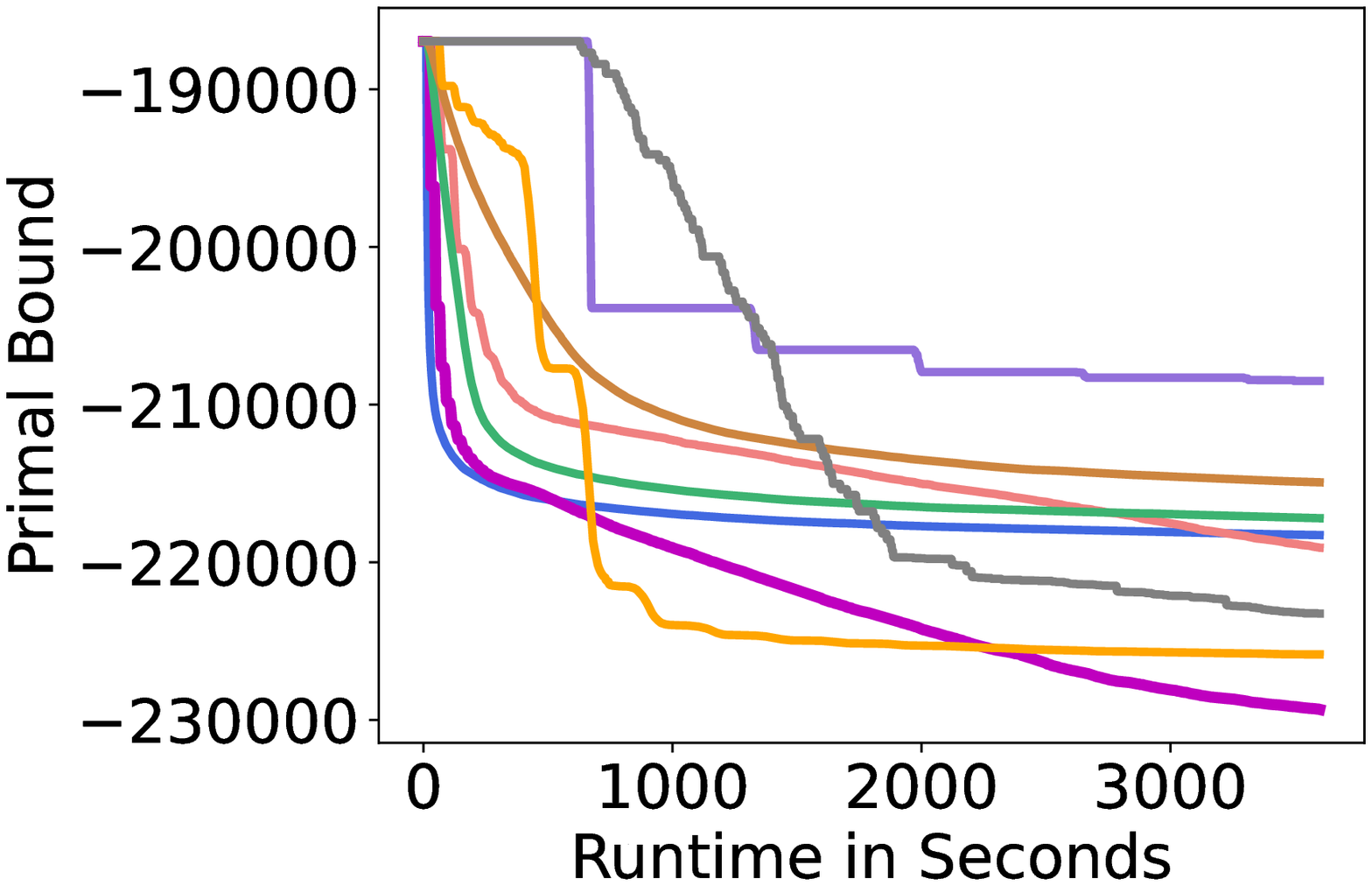}    
    }
    \subfloat[SC-S (left) and SC-L (right).]
    {
        \includegraphics[width=0.24\textwidth]{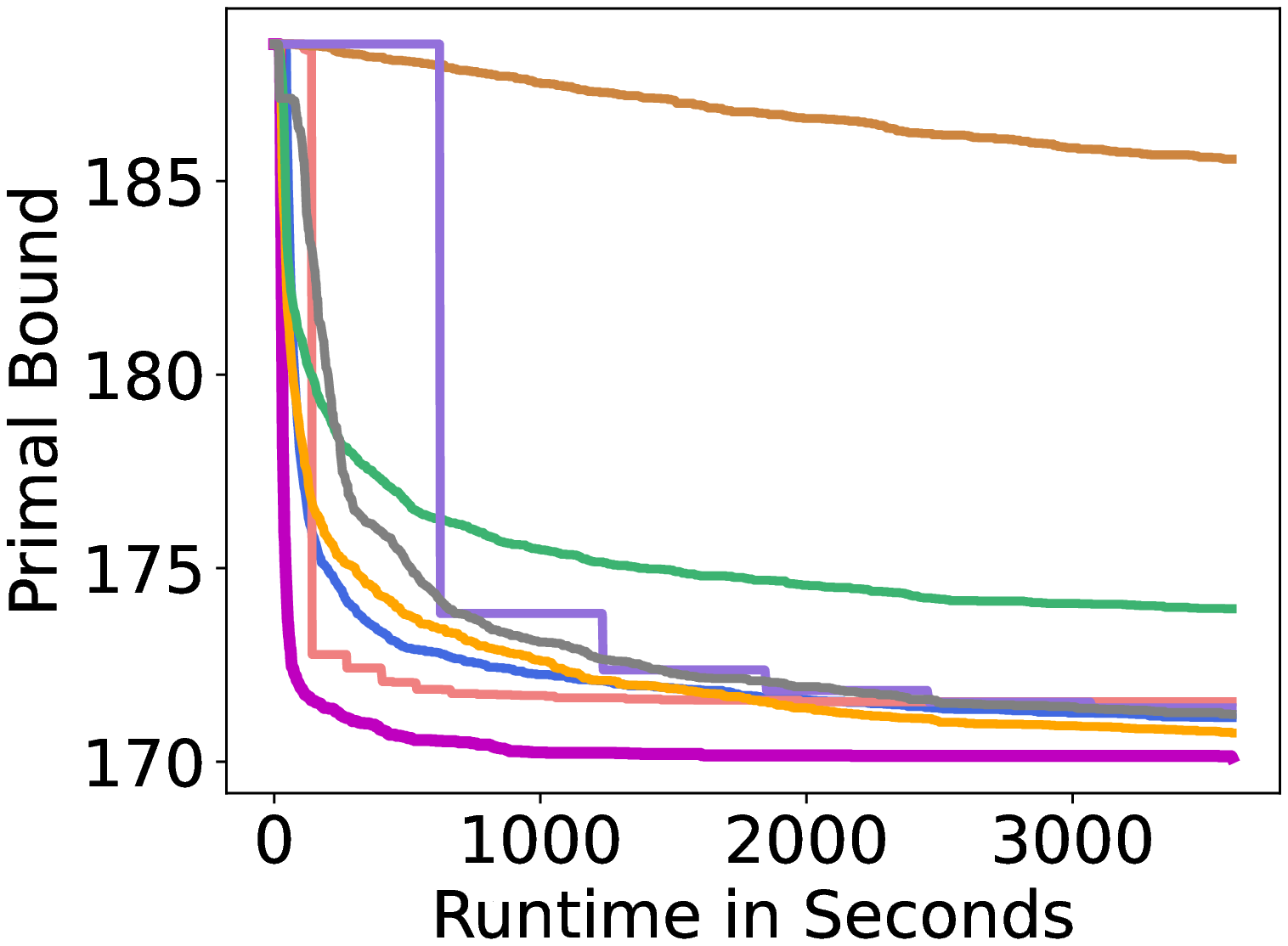}
        \includegraphics[width=0.24\textwidth]{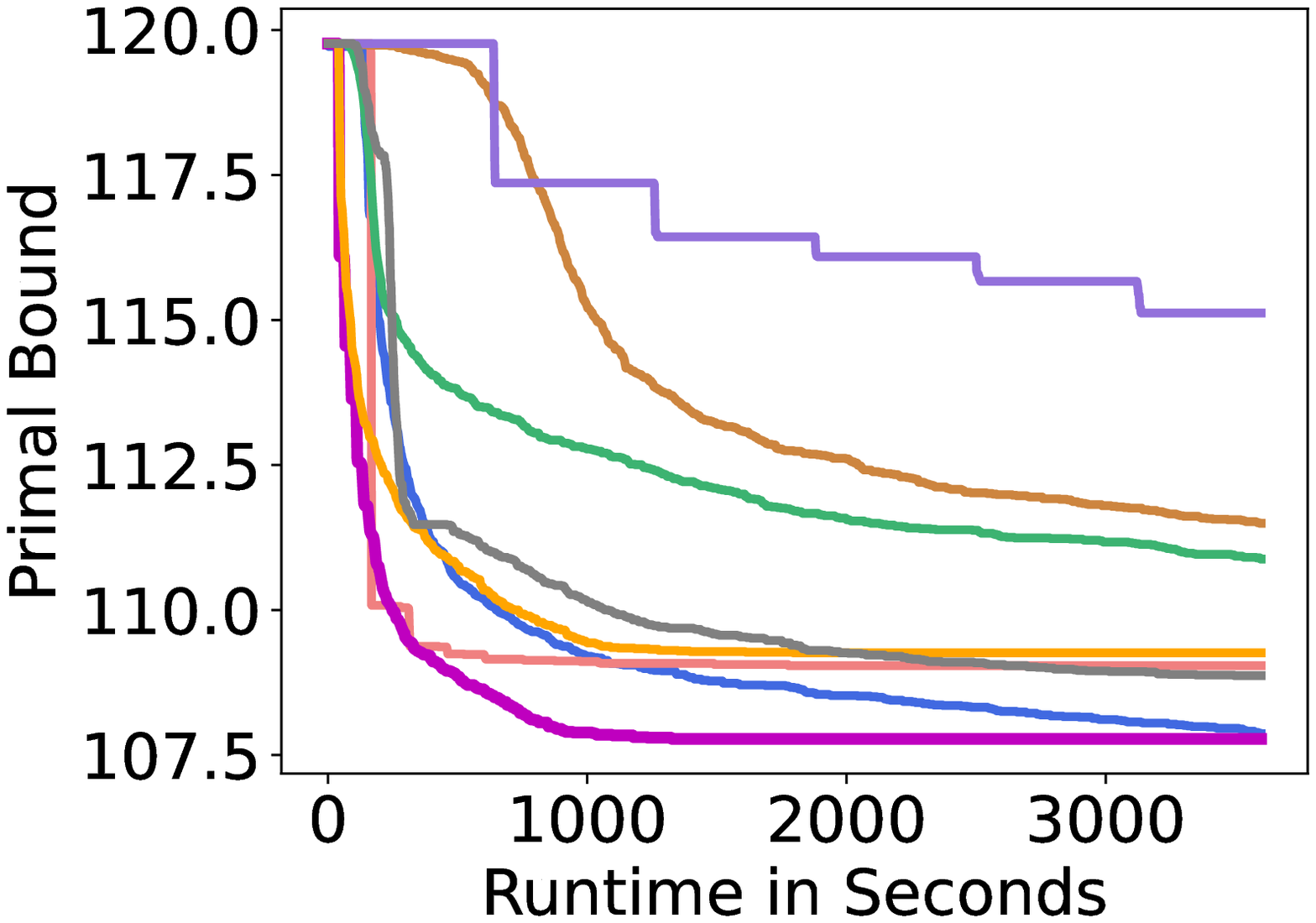}    
    }
    \caption{The primal bound (the lower the better) as a function of time, averaged over 100 instances. For ML approaches, the policies are trained on only small training instances but tested on both small and large test instances.\label{res_all::bound}}
\end{figure*}

Figure \ref{res_all::gap} shows the full results on the primal gap as a function of runtime. 
Figure \ref{res_all::survival} shows the full results on the survival rate as a function of runtime. 
Figure \ref{res_all::bound} shows the full results on the primal bound as a function of runtime. Tables \ref{res::smalltable15}, \ref{res::smalltable30}, \ref{res::smalltable45} and \ref{res::smalltable60_full} present the average primal bound, primal gap and primal integral at 15, 30, 45 and 60 minutes runtime cutoff, respectively, on the small instances.
Tables \ref{res::bigtable15}, \ref{res::bigtable30}, \ref{res::bigtable45} and \ref{res::bigtable60_full} present the average primal bound, primal gap and primal integral at 15, 30, 45 and 60 minutes runtime cutoff, respectively, on the large instances.

Next, we  evaluate the performance with one additional metric:
    The \textit{gap to virtual best} at time $q$ for an approach is the normalized difference between its best primal bound found up to time $q$ and the best primal bound found up to time $q$ by any approach in the portfolio.

\begin{figure*}[btp]
    \centering
    \includegraphics[width=\textwidth]{figure_all/legend_ML_horizontal_timeVSobj_3600.eps}

    \subfloat[MVC-S (left) and MVC-L (right).]
    {
        \includegraphics[width=0.24\textwidth]{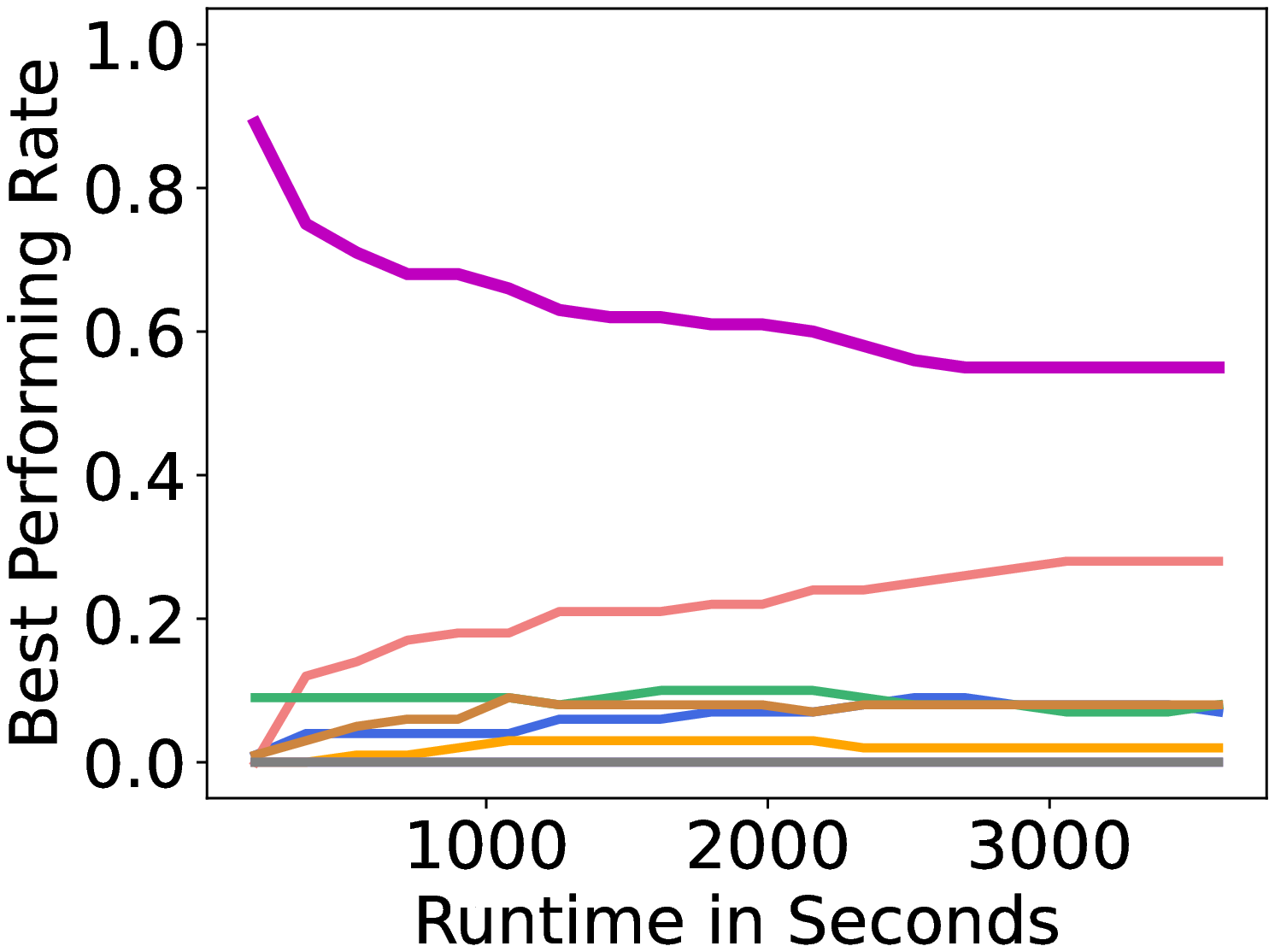}
        \includegraphics[width=0.24\textwidth]{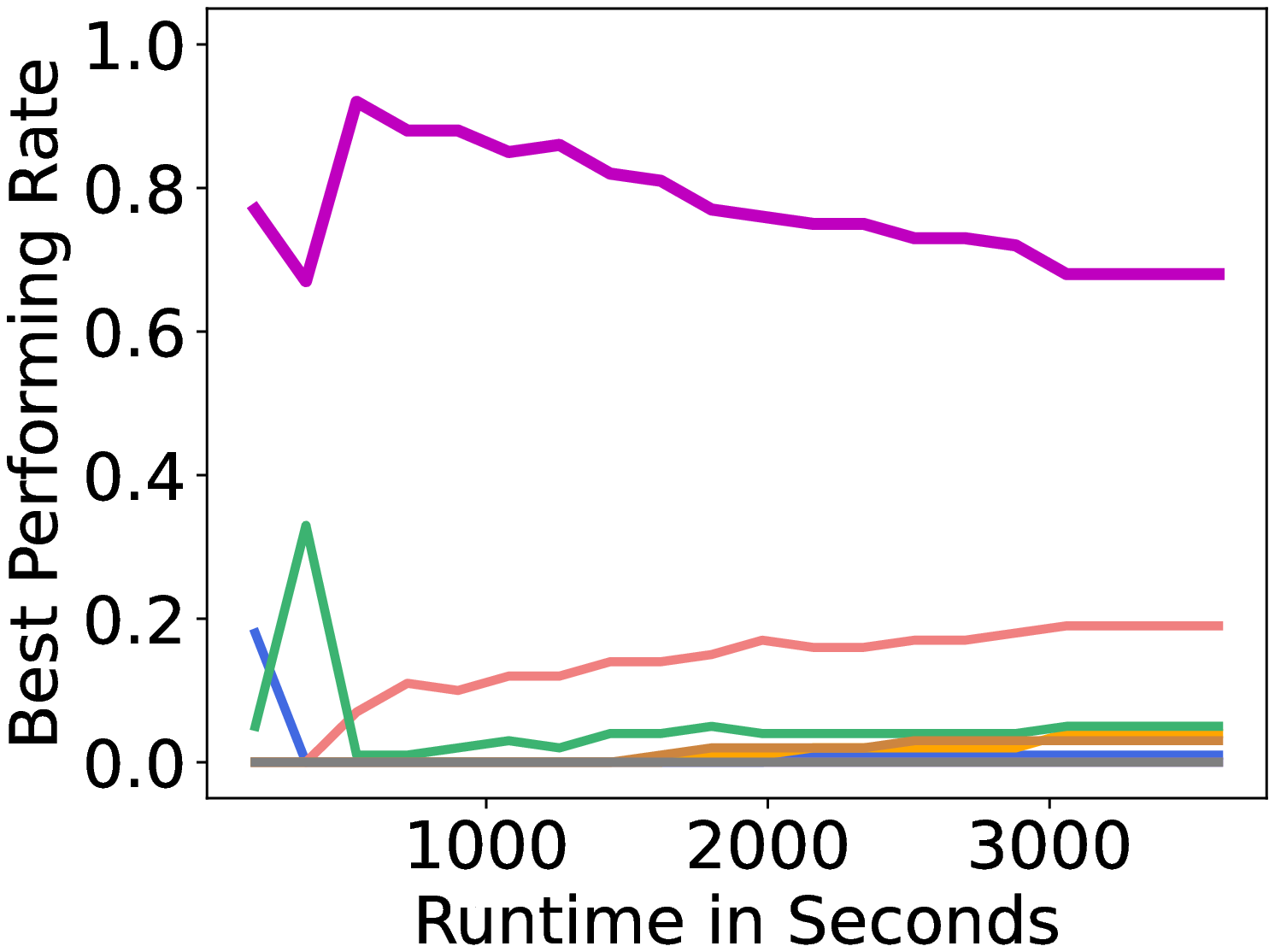}    
    }
    \subfloat[MIS-S (left) and MIS-L (right).]
    {
        \includegraphics[width=0.24\textwidth]{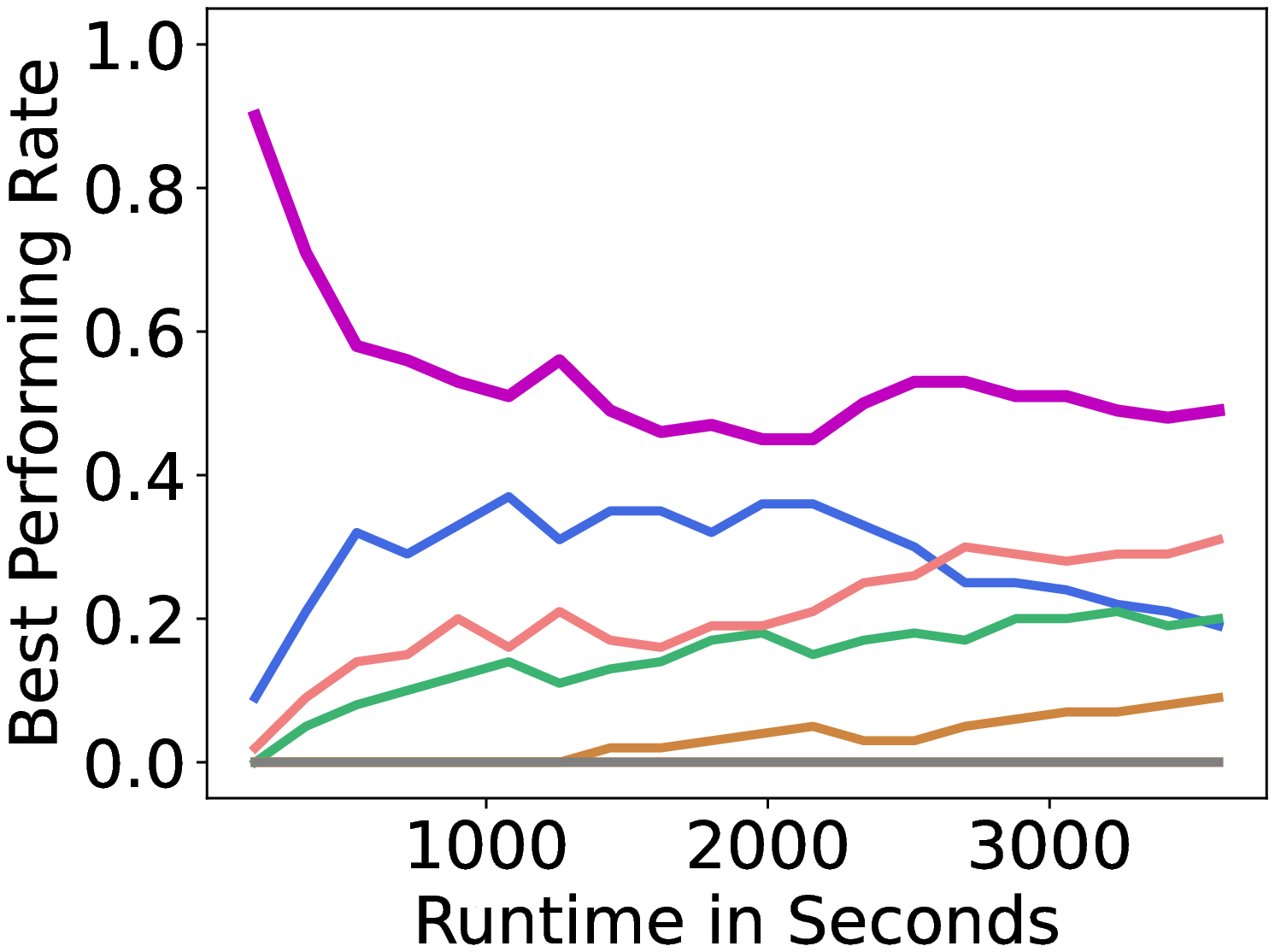}
        \includegraphics[width=0.24\textwidth]{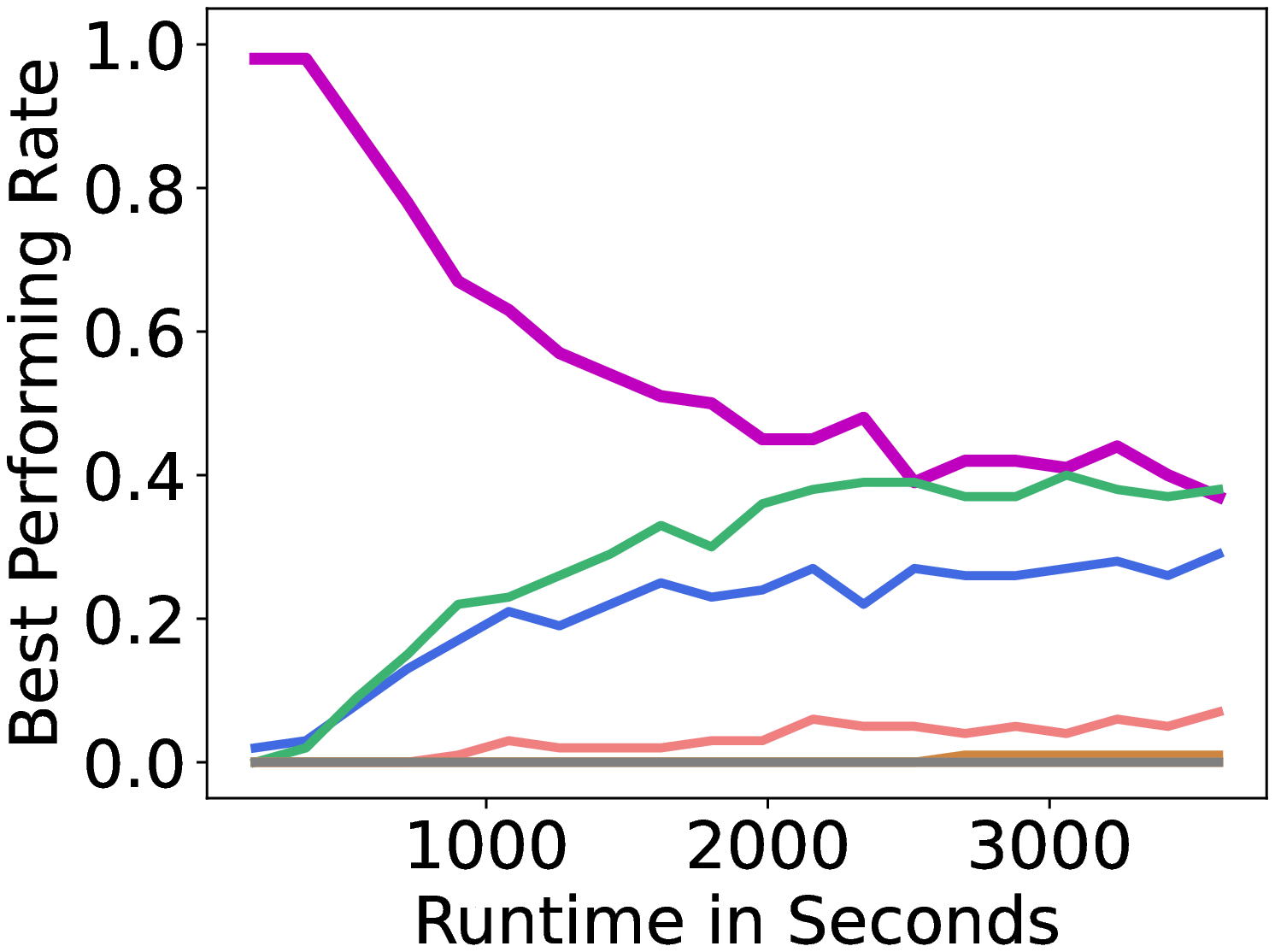}    
    }\\
    \subfloat[CA-S (left) and CA-L (right).]
    {
        \includegraphics[width=0.24\textwidth]{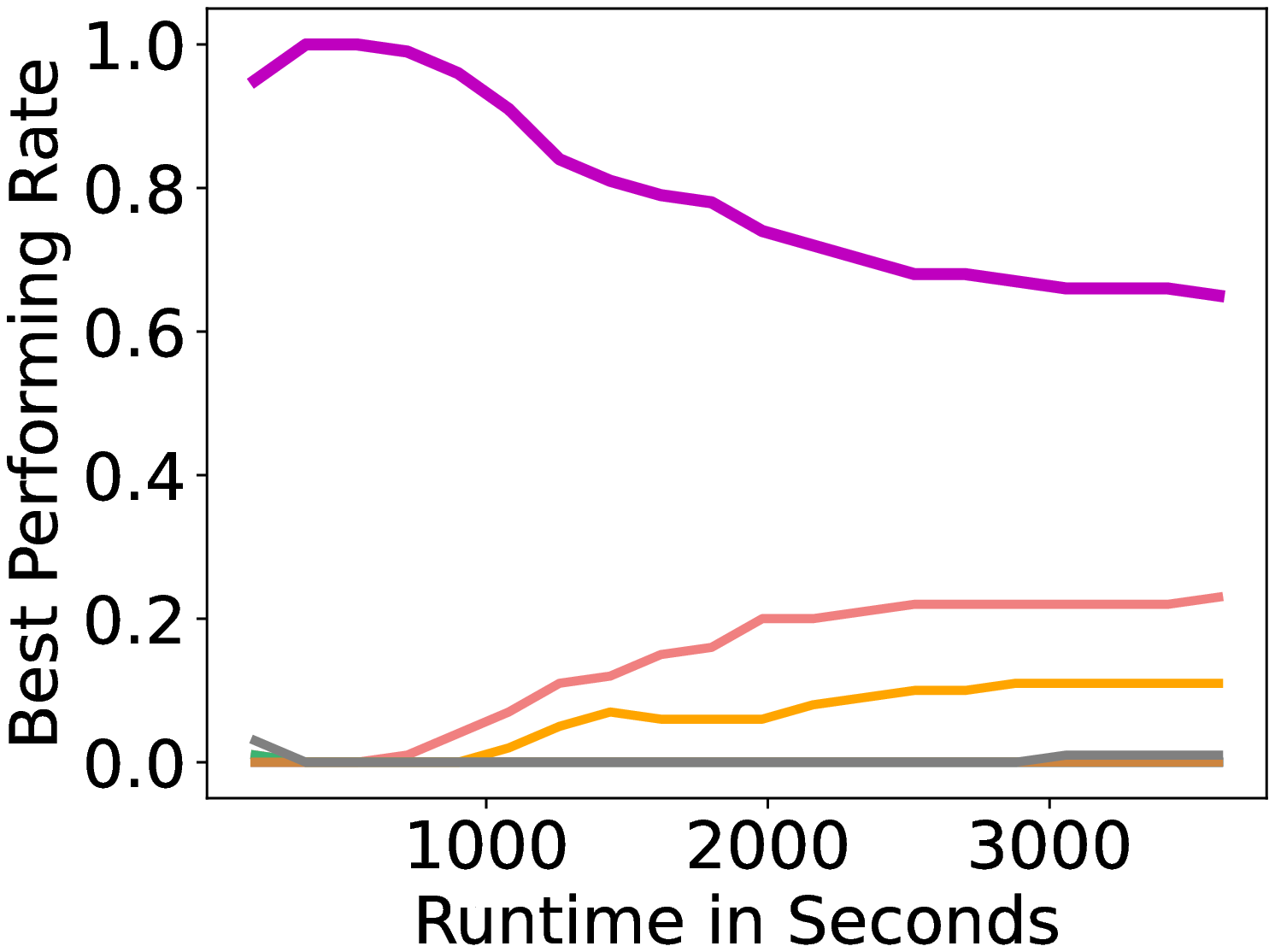}
        \includegraphics[width=0.24\textwidth]{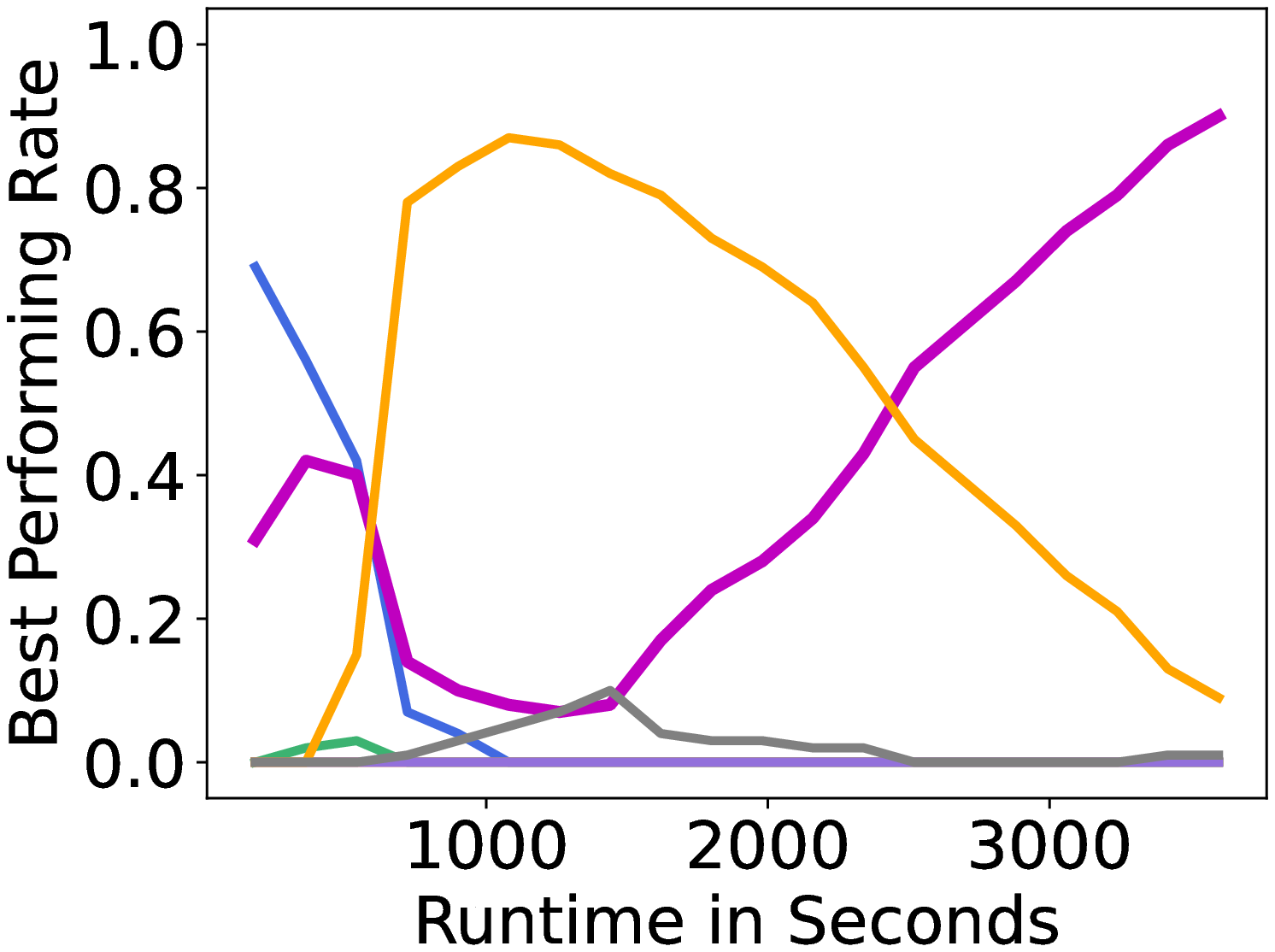}    
    }
    \subfloat[SC-S (left) and SC-L (right).]
    {
        \includegraphics[width=0.24\textwidth]{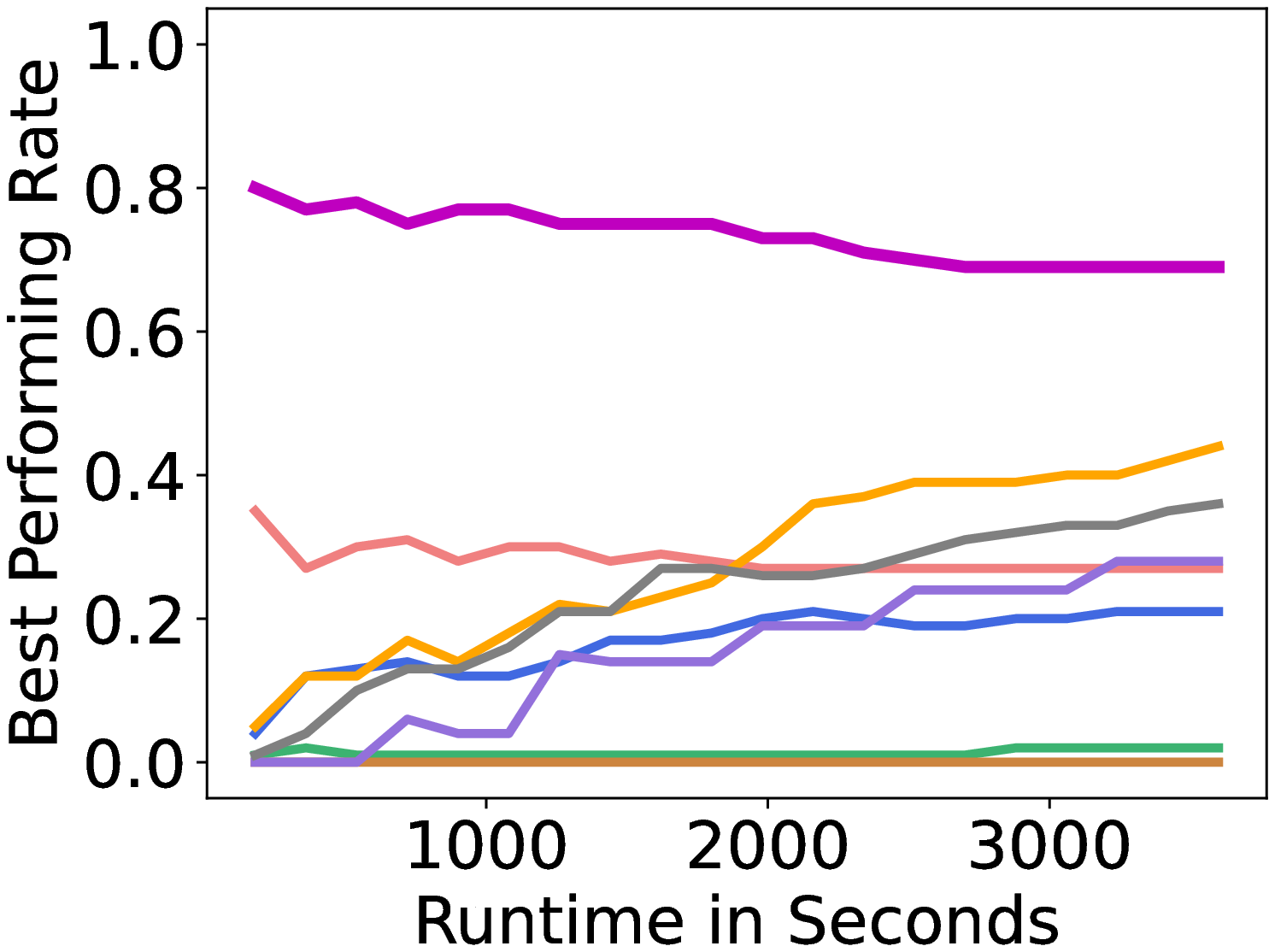}
        \includegraphics[width=0.24\textwidth]{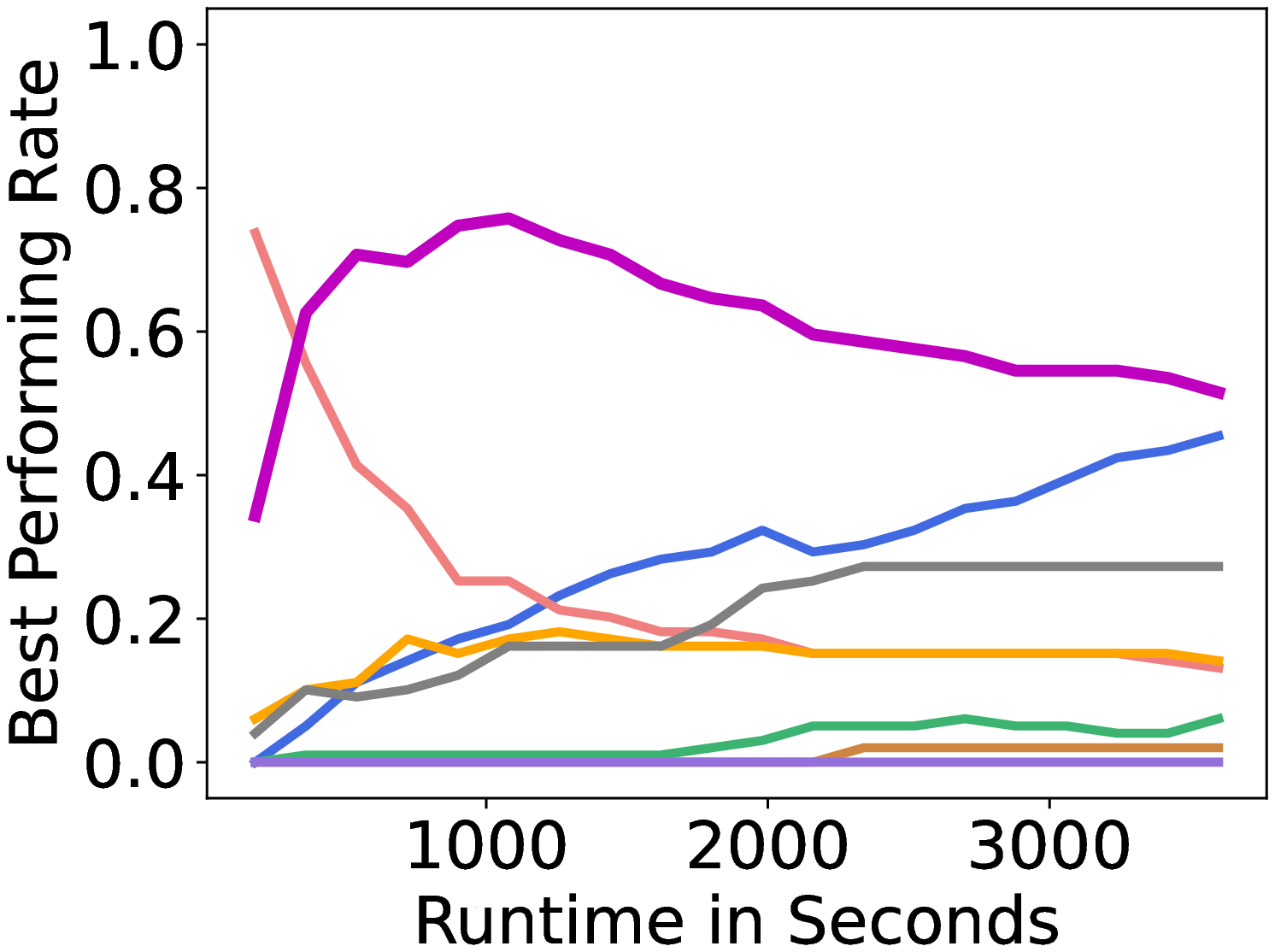}    
    }
    \caption{The best performing rate (the higher the better) as a function of runtime over 100 test instances. For ML approaches, the policies are trained on only small training instances but tested on both small and large test instances.\label{res_all::winRate}}
\end{figure*}

\begin{figure*}[btp]
    \centering
    \includegraphics[width=\textwidth]{figure_all/legend_ML_horizontal_timeVSobj_3600.eps}

    \subfloat[MVC-S (left) and MVC-L (right).]
    {
        \includegraphics[width=0.24\textwidth]{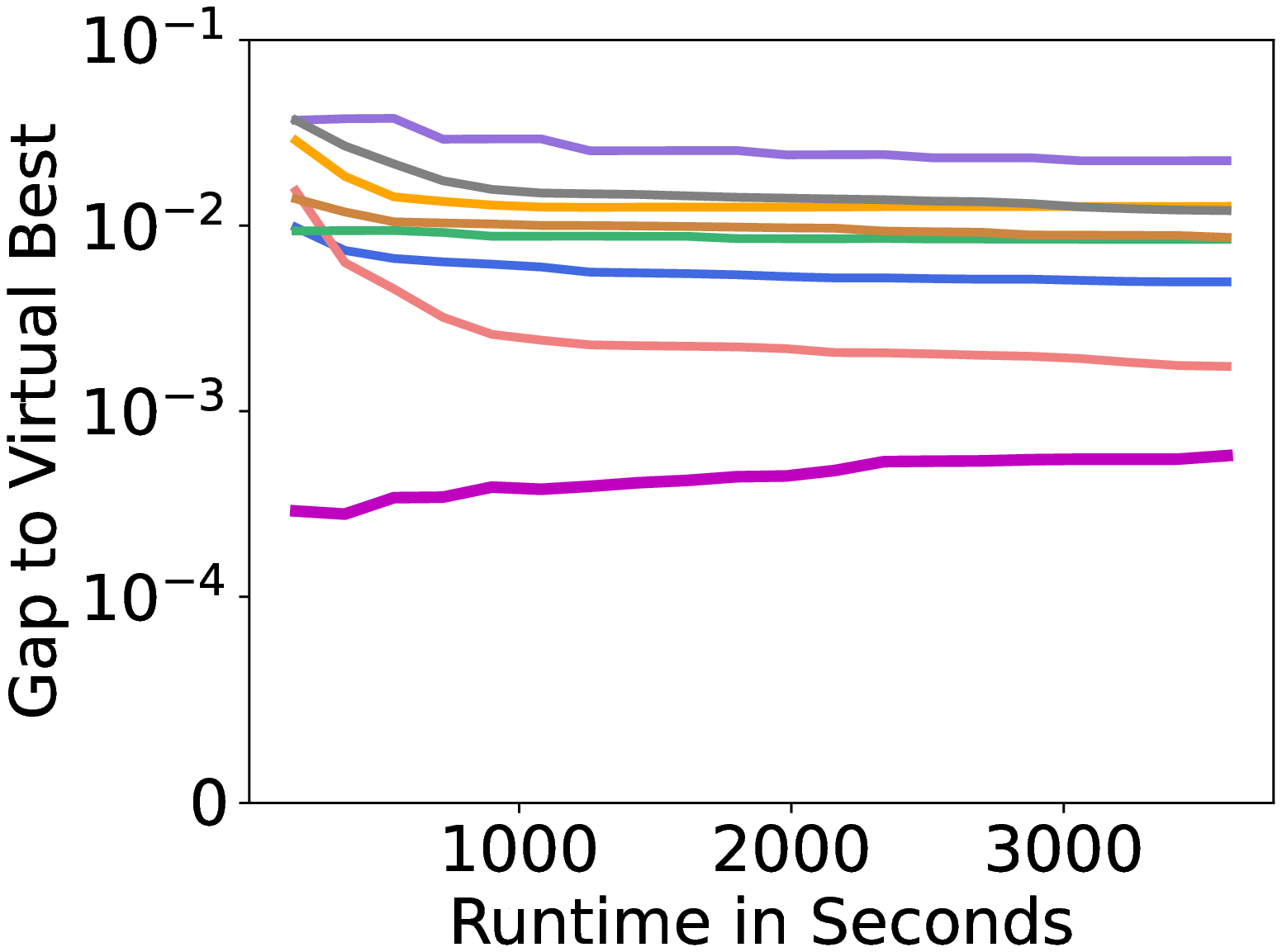}
        \includegraphics[width=0.24\textwidth]{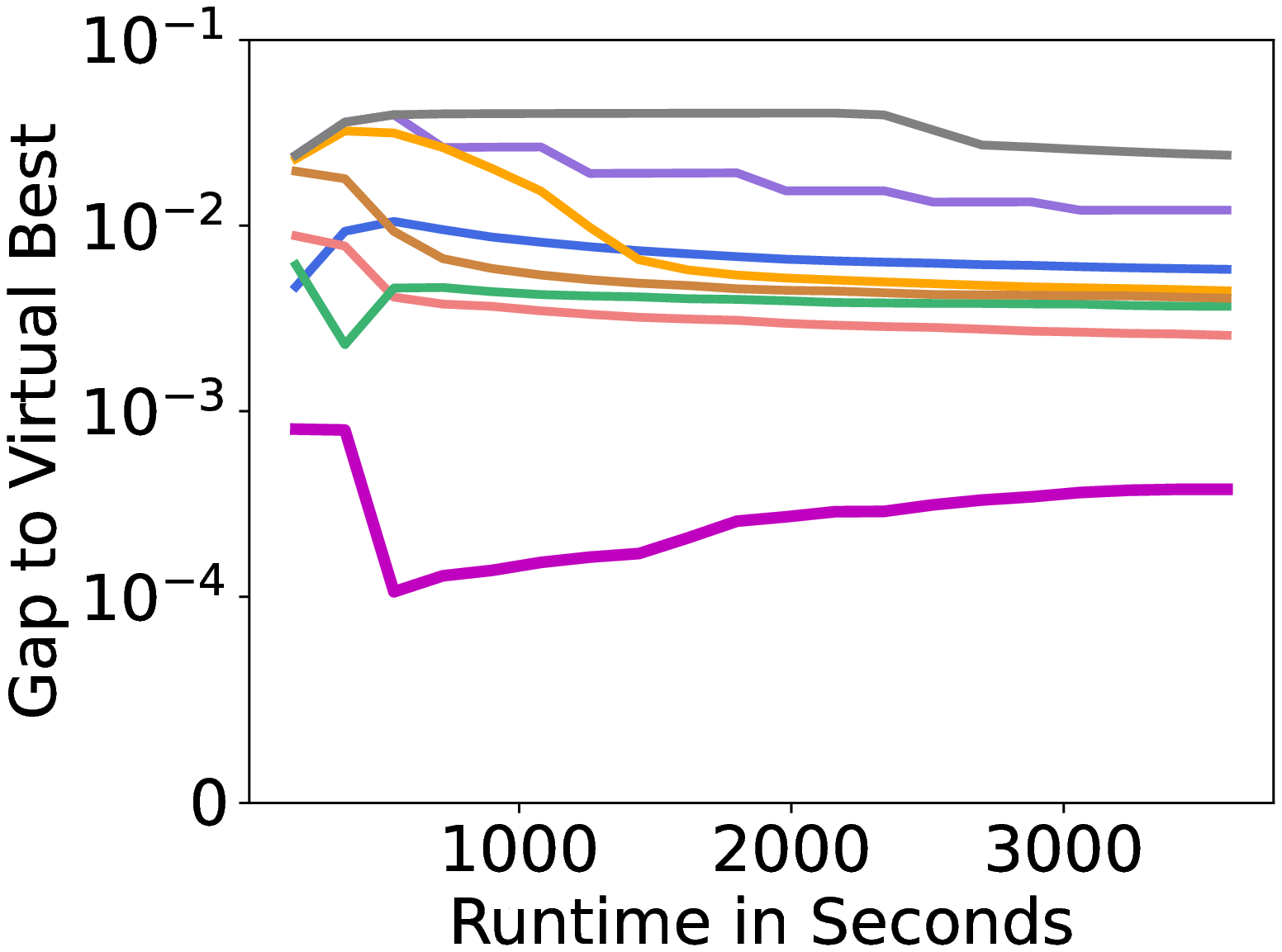}    
    }
    \subfloat[MIS-S (left) and MIS-L (right).]
    {
        \includegraphics[width=0.24\textwidth]{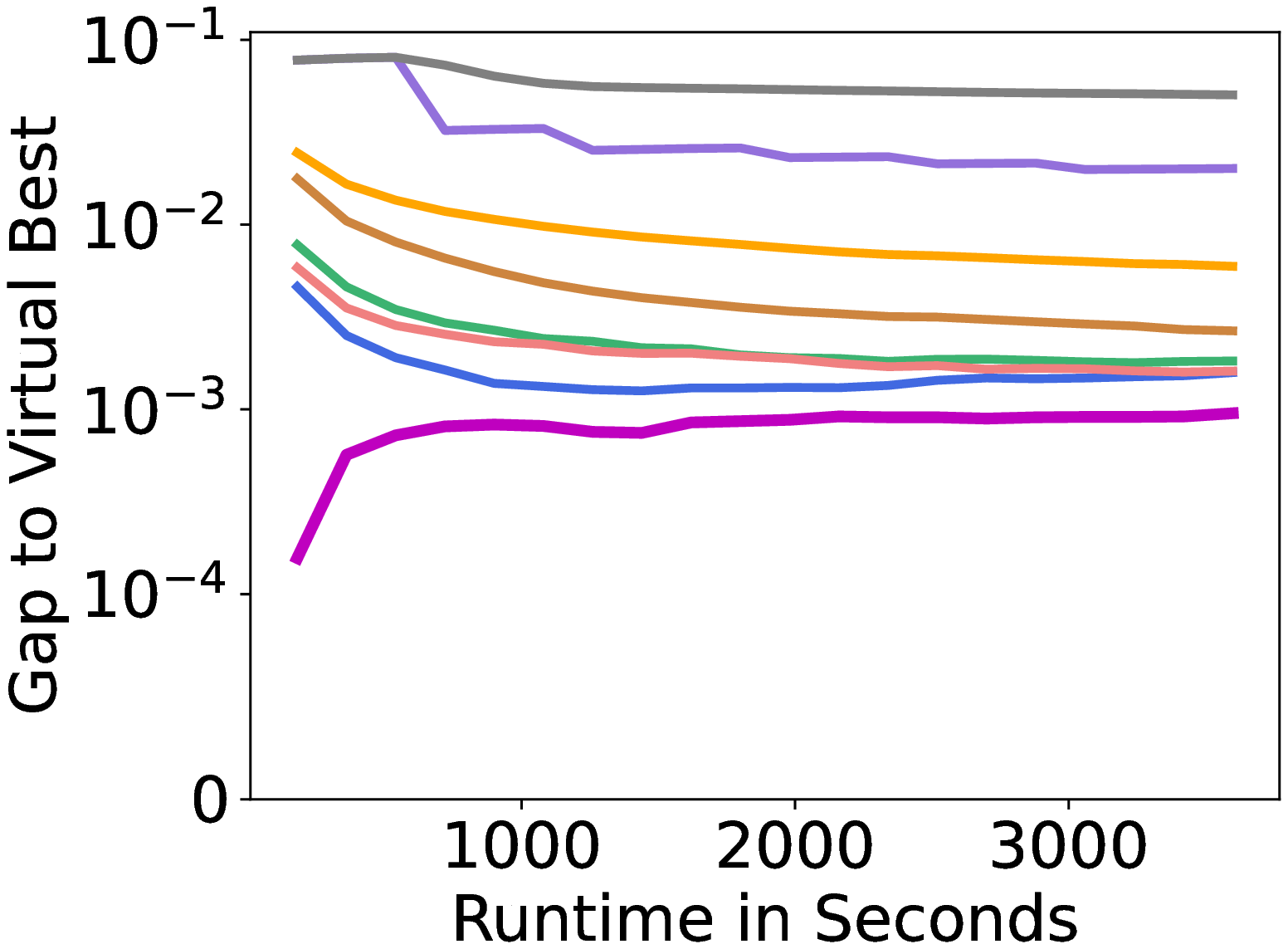}
        \includegraphics[width=0.24\textwidth]{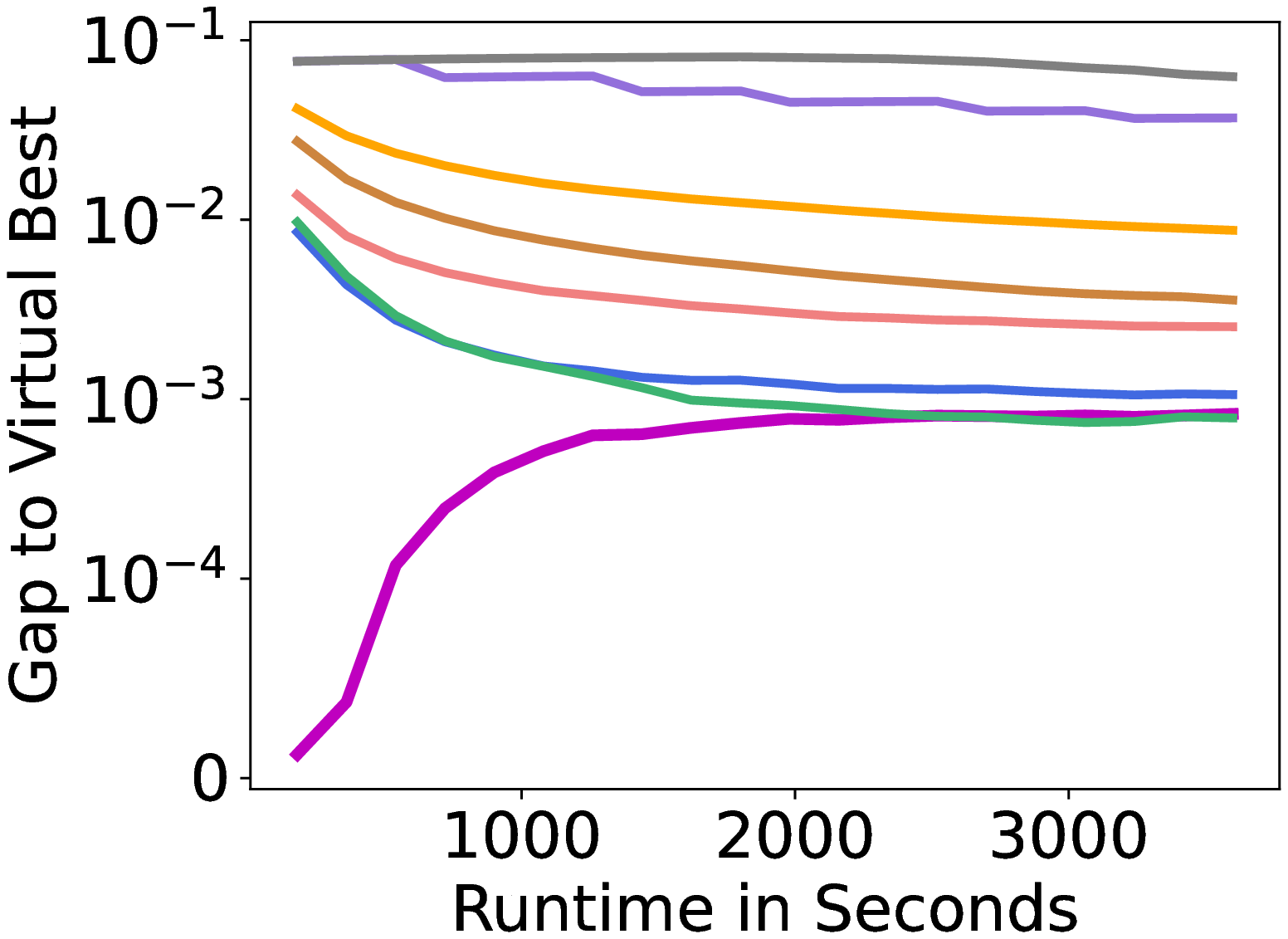}    
    }\\
    \subfloat[CA-S (left) and CA-L (right).]
    {
        \includegraphics[width=0.24\textwidth]{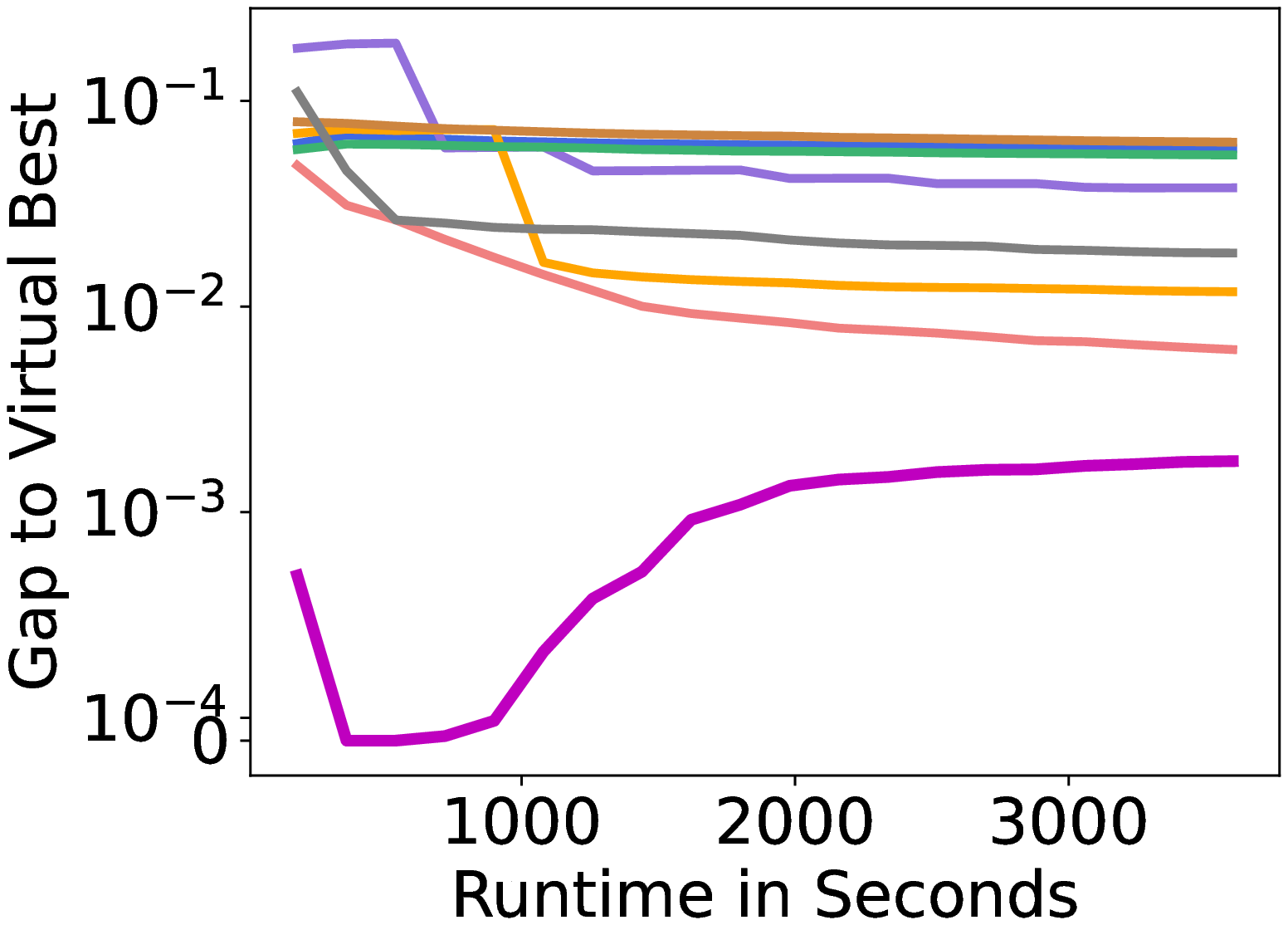}
        \includegraphics[width=0.24\textwidth]{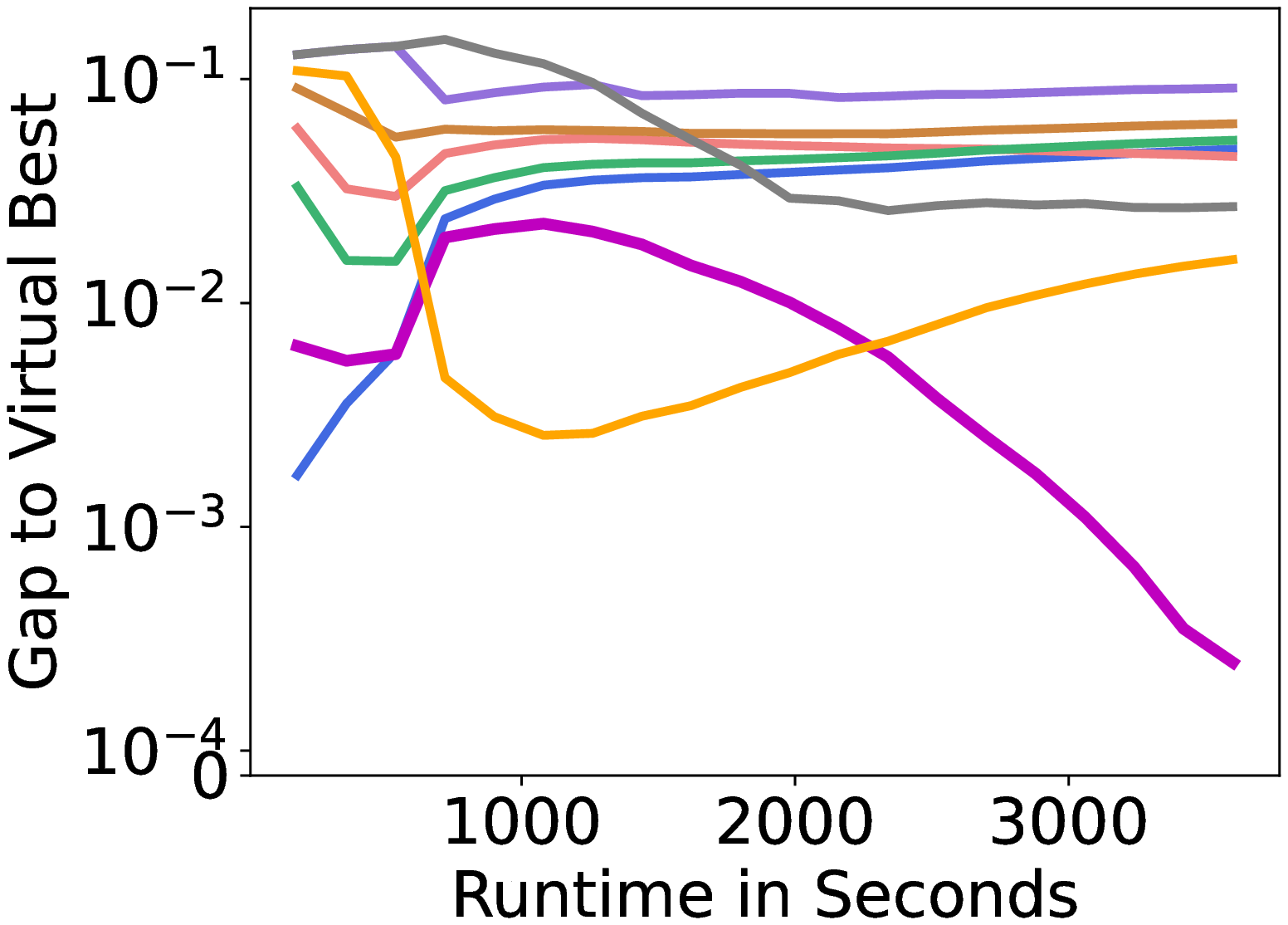}    
    }
    \subfloat[SC-S (left) and SC-L (right).]
    {
        \includegraphics[width=0.24\textwidth]{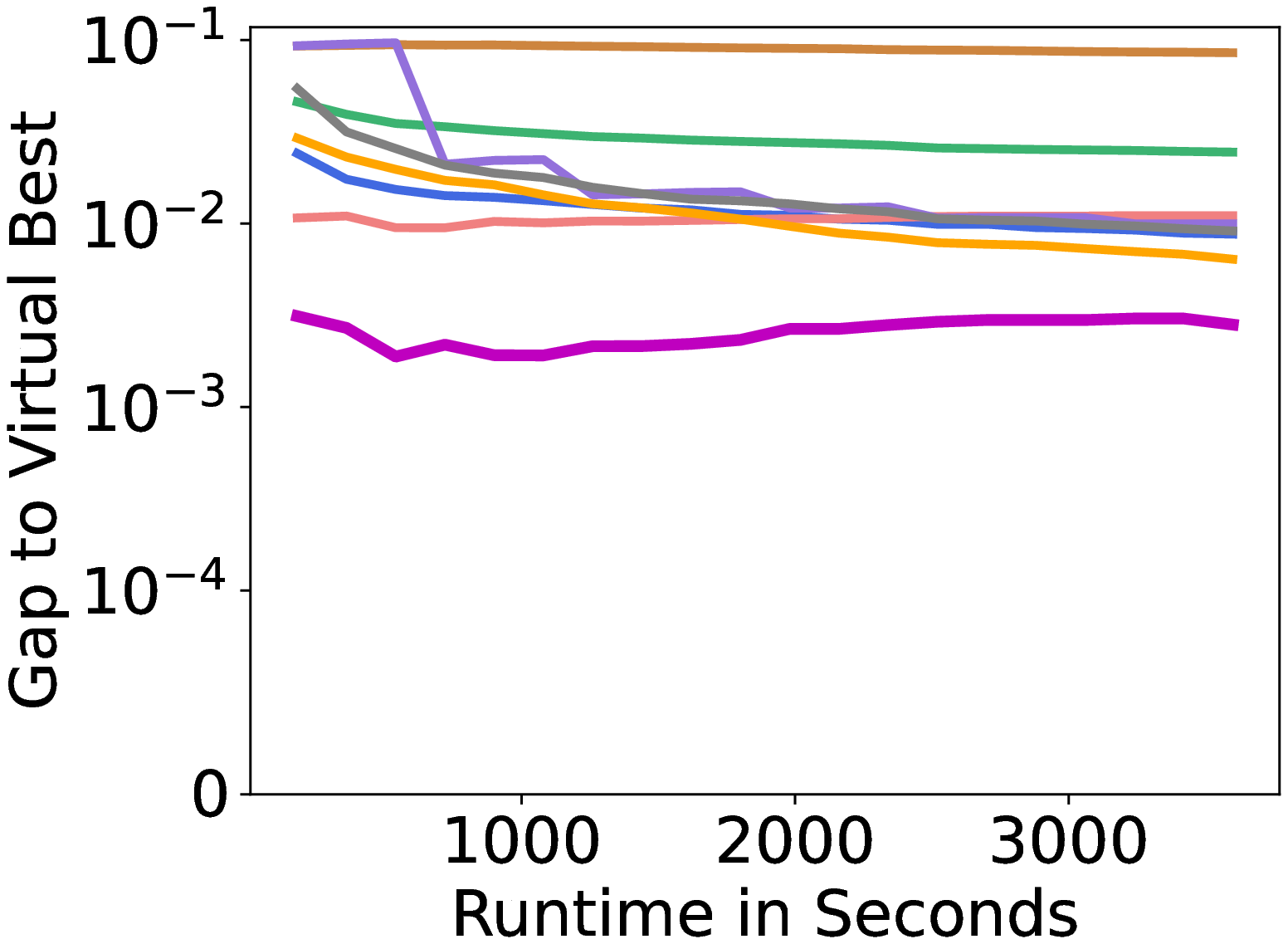}
        \includegraphics[width=0.24\textwidth]{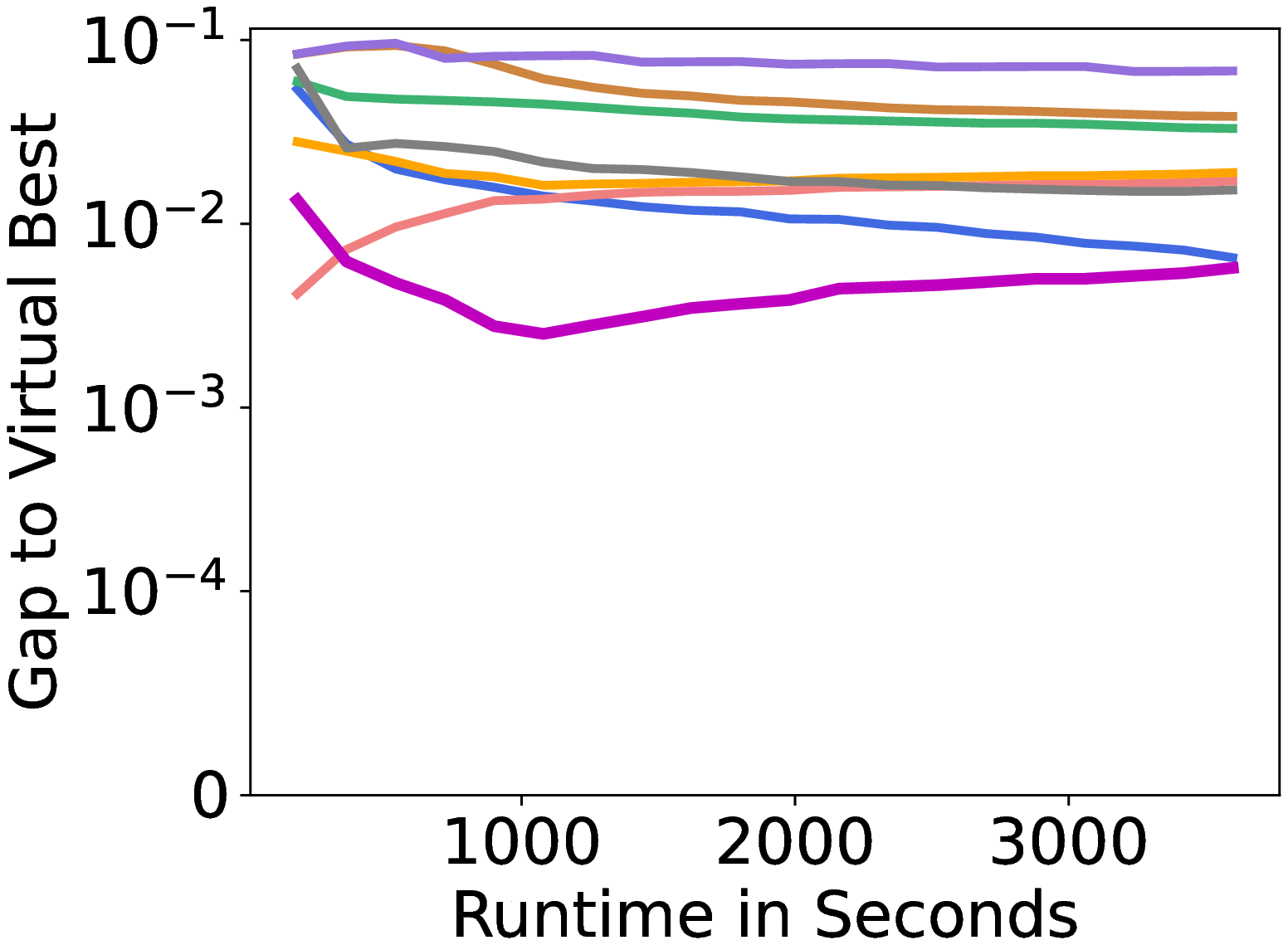}    
    }
    \caption{The gap to virtual best (the lower the better) as a function of runtime, averaged over 100 test instances. For ML approaches, the policies are trained on only small training instances but tested on both small and large test instances.\label{res_all::virtualBest}}
\end{figure*}

Figure \ref{res_all::winRate} shows the full results on the best performing rate as a function of runtime. 
Figure \ref{res_all::virtualBest} shows the full results on the gap to virtual best as a function of runtime.

\begin{table*}[htbp]
\scriptsize
\centering
\caption{Test results on small instances: Primal bound (PB), primal gap (PG) (in percent), primal integral (PI) at 15 minutes time cutoff, averaged over 100 instances and their standard deviations. \label{res::smalltable15}}

\begin{tabular}{c|rrr|rrr}
\hline
                       & \multicolumn{1}{c}{PB}             & \multicolumn{1}{c}{PG (\%)}         & \multicolumn{1}{c|}{PI} & \multicolumn{1}{c}{PB}              & \multicolumn{1}{c}{PG (\%)}         & \multicolumn{1}{c}{PI} \\ \hline

                       & \multicolumn{3}{c|}{MVC}                                                                        & \multicolumn{3}{c}{MIS}                                                                         \\ \hline
BnB                    & 450.41$\pm$9.85 & 1.71$\pm$0.48 & 25.7$\pm$3.3           & -1,981.72$\pm$23.49 & 6.66$\pm$0.89 & 74.2$\pm$4.4           \\ 
LB                     &  456.78$\pm$11.22 & 3.07$\pm$1.00 & 32.9$\pm$5.1            & -2,047.01$\pm$18.76 & 3.58$\pm$0.60 & 62.4$\pm$3.8              \\ 
\RANDOM                 & 447.33$\pm$11.33 & 1.02$\pm$1.28 & 11.5$\pm$11.3              & -2,110.73$\pm$11.86 & 0.58$\pm$0.19 & 12.8$\pm$1.6             \\ 
\GRAPH                  & 447.98$\pm$11.30 & 1.16$\pm$1.28 & 14.0$\pm$10.6             & -2,104.62$\pm$12.23 & 0.87$\pm$0.17 & 18.5$\pm$1.7             \\ 
\LBRELAX                & 449.23$\pm$11.49 & 1.43$\pm$1.51 & 19.6$\pm$10.9            & -2,093.80$\pm$12.07 & 1.38$\pm$0.23 & 22.9$\pm$2.1             \\ 
\DM       & 444.50$\pm$9.69 & 0.40$\pm$0.28 & 10.2$\pm$5.5              & -2,111.49$\pm$12.10 & 0.54$\pm$0.20 & 10.5$\pm$1.8            \\ 
\RL               & 446.12$\pm$10.10 & 0.76$\pm$0.36 & 11.9$\pm$2.9             & -2,113.48$\pm$11.72 & 0.45$\pm$0.17 & 9.5$\pm$1.7            \\ 
\CL               & {\bf443.51$\pm$9.58} & {\bf0.18$\pm$0.10} & {\bf4.0$\pm$2.1}             & {\bf-2,114.66$\pm$12.42} & {\bf0.39$\pm$0.19} & {\bf6.4$\pm$1.6}            \\ \hline
\multicolumn{1}{l|}{} & \multicolumn{3}{c|}{CA}                                                                         & \multicolumn{3}{c}{SC}                                                                          \\ \hline
BnB                 & -112,703$\pm$1,682 & 3.06$\pm$0.70 & 67.4$\pm$16.6      & 173.26$\pm$13.00 & 2.28$\pm$1.34 & 45.9$\pm$13.0                         \\ 
LB                  & -108,647$\pm$2,227 & 6.55$\pm$1.42 & 140.7$\pm$9.9    & 173.83$\pm$12.93 & 2.60$\pm$1.31 & 70.6$\pm$15.6      \\ 
\RANDOM            & -108,576$\pm$1,709 & 6.61$\pm$1.12 & 69.1$\pm$8.5     & 175.61$\pm$12.76 & 3.60$\pm$1.44 & 43.6$\pm$13.8                    \\ 
\GRAPH             & -107,189$\pm$1,977 & 7.81$\pm$1.15 & 84.7$\pm$9.8     & 187.69$\pm$14.24 & 9.77$\pm$2.17 & 89.9$\pm$19.9           \\ 
\LBRELAX        & -107,133$\pm$1,816 & 7.86$\pm$0.76 & 89.5$\pm$6.2      & 172.79$\pm$12.76 & 2.02$\pm$1.21 & 30.0$\pm$11.4    \\ 
\DM            & -113,501$\pm$1,611 & 2.38$\pm$0.66 & 52.4$\pm$10.9  & 171.72$\pm$12.42 & 1.43$\pm$1.00 & 26.9$\pm$9.2            \\ 
\RL            & -108,120$\pm$1,906 & 7.01$\pm$1.10 & 71.8$\pm$9.3  & 172.35$\pm$12.45 & 1.79$\pm$0.96 & 41.4$\pm$8.2            \\ 
\CL & {\bf-115,499$\pm$1,626} & {\bf0.66$\pm$0.33} & {\bf33.3$\pm$6.8}  & {\bf170.27$\pm$12.21} & {\bf0.59$\pm$0.67} & {\bf11.7$\pm$7.4}\\ \hline
\end{tabular}
\end{table*}

\begin{table*}[htbp]
\scriptsize
\centering
\caption{Test results on small instances: Primal bound (PB), primal gap (PG) (in percent), primal integral (PI) at 30 minutes time cutoff, averaged over 100 instances and their standard deviations. \label{res::smalltable30}}

\begin{tabular}{c|rrr|rrr}
\hline
                       & \multicolumn{1}{c}{PB}             & \multicolumn{1}{c}{PG (\%)}         & \multicolumn{1}{c|}{PI} & \multicolumn{1}{c}{PB}              & \multicolumn{1}{c}{PG (\%)}         & \multicolumn{1}{c}{PI} \\ \hline

                       & \multicolumn{3}{c|}{MVC}                                                                        & \multicolumn{3}{c}{MIS}                                                                         \\ \hline
BnB                    & 449.67$\pm$9.69 & 1.55$\pm$0.44 & 40.2$\pm$6.6           & -2,004.24$\pm$26.21 & 5.60$\pm$1.00 & 127.1$\pm$12.4\\ 
LB                    & 454.89$\pm$11.55 & 2.66$\pm$1.16 & 58.2$\pm$14.1              & -2,064.30$\pm$16.40 & 2.77$\pm$0.51 & 89.9$\pm$7.3             \\ 
\RANDOM                 & 447.16$\pm$11.22 & 0.98$\pm$1.26 & 20.6$\pm$22.5             & -2,115.23$\pm$11.82 & 0.37$\pm$0.16 & 16.9$\pm$2.7             \\ 
\GRAPH                  & 447.75$\pm$11.39 & 1.11$\pm$1.30 & 24.2$\pm$22.1             & -2,111.84$\pm$12.06 & 0.53$\pm$0.16 & 24.4$\pm$2.7            \\ 
\LBRELAX                & 449.02$\pm$11.53 & 1.38$\pm$1.51 & 32.1$\pm$24.2             & -2,102.85$\pm$11.97 & 0.95$\pm$0.19 & 33.0$\pm$3.6            \\ 
\DM       & 444.27$\pm$9.61 & 0.35$\pm$0.25 & 13.5$\pm$6.9             & -2,115.30$\pm$12.04 & 0.36$\pm$0.18 & 14.4$\pm$3.2             \\ 
\RL               & 445.71$\pm$9.98 & 0.67$\pm$0.35 & 18.2$\pm$5.7            & -2,116.64$\pm$11.53 & 0.30$\pm$0.15 & 12.7$\pm$2.9            \\ 
\CL              & {\bf443.48$\pm$9.56} & {\bf0.17$\pm$0.09} & {\bf5.5$\pm$3.6}             & {\bf-2,117.58$\pm$11.86} & {\bf0.26$\pm$0.17} & {\bf9.3$\pm$3.0}            \\ \hline
\multicolumn{1}{l|}{} & \multicolumn{3}{c|}{CA}                                                                         & \multicolumn{3}{c}{SC}                                                                          \\ \hline
BnB                 & -113,068$\pm$1,595 & 2.75$\pm$0.62 & 93.5$\pm$18.6     & 172.09$\pm$12.65 & 1.63$\pm$1.20 & 62.9$\pm$22.5                         \\ 
LB                  & -110,303$\pm$2,001 & 5.13$\pm$1.08 & 191.6$\pm$16.9     & 172.37$\pm$12.71 & 1.79$\pm$1.11 & 89.4$\pm$22.3      \\ 
\RANDOM            & -109,040$\pm$1,685 & 6.21$\pm$1.05 & 126.8$\pm$17.6     & 174.70$\pm$12.75 & 3.10$\pm$1.38 & 73.4$\pm$24.6                    \\ 
\GRAPH             & -107,802$\pm$1,892 & 7.28$\pm$1.07 & 152.2$\pm$18.9      & 186.79$\pm$14.13 & 9.33$\pm$2.28 & 175.7$\pm$38.8          \\ 
\LBRELAX        & -114,103$\pm$1,521 & 1.86$\pm$0.57 & 109.5$\pm$9.4     & 171.60$\pm$12.43 & 1.36$\pm$1.02 & 44.6$\pm$19.3    \\ 
\DM            & -114,621$\pm$1638 & 1.41$\pm$0.58 & 68.1$\pm$13.9   & 171.59$\pm$12.45 & 1.35$\pm$1.00 & 39.3$\pm$17.4             \\ 
\RL            & -108,562$\pm$1,854 & 6.63$\pm$1.05 & 132.9$\pm$18.2   & 171.70$\pm$12.30 & 1.42$\pm$0.88 & 55.7$\pm$15.6            \\ 
\CL & {\bf-115,513$\pm$1,621} & {\bf0.65$\pm$0.32} & {\bf39.1$\pm$11.6}  &  {\bf170.16$\pm$12.13} &{\bf0.53$\pm$0.63} & {\bf16.7$\pm$12.3}\\ \hline
\end{tabular}
\end{table*}

\begin{table*}[htbp]
\scriptsize
\centering
\caption{Test results on small instances: Primal bound (PB), primal gap (PG) (in percent), primal integral (PI) at 45 minutes time cutoff, averaged over 100 instances and their standard deviations. \label{res::smalltable45}}

\begin{tabular}{c|rrr|rrr}
\hline
                       & \multicolumn{1}{c}{PB}             & \multicolumn{1}{c}{PG (\%)}         & \multicolumn{1}{c|}{PI} & \multicolumn{1}{c}{PB}              & \multicolumn{1}{c}{PG (\%)}         & \multicolumn{1}{c}{PI} \\ \hline

                       & \multicolumn{3}{c|}{MVC}                                                                        & \multicolumn{3}{c}{MIS}                                                                         \\ \hline
BnB                    & 449.28$\pm$9.77 & 1.46$\pm$0.42 & 53.7$\pm$9.9           & -2,010.68$\pm$21.72 & 5.29$\pm$0.79 & 176.0$\pm$19.7            \\ 
LB                     & 453.84$\pm$11.65 & 2.44$\pm$1.26 & 80.7$\pm$24.6              & -2,075.43$\pm$14.84 & 2.24$\pm$0.46 & 111.6$\pm$10.5             \\ 
\RANDOM                 & 447.09$\pm$11.21 & 0.96$\pm$1.26 & 29.4$\pm$33.6             & -2,116.96$\pm$11.54 & 0.29$\pm$0.15 & 19.8$\pm$3.9             \\ 
\GRAPH                  & 447.42$\pm$11.19 & 1.04$\pm$1.27 & 33.9$\pm$33.4              & -2,114.42$\pm$11.74 & 0.41$\pm$0.16 & 28.6$\pm$3.8             \\ 
\LBRELAX               & 449.01$\pm$11.53 & 1.38$\pm$1.51 & 44.6$\pm$37.6             & -2,106.88$\pm$11.40 & 0.76$\pm$0.20 & 40.6$\pm$5.0             \\ 
\DM                     & 444.13$\pm$9.68 & 0.32$\pm$0.26 & 16.5$\pm$8.5              & -2,117.43$\pm$11.79 & 0.26$\pm$0.17 & 17.2$\pm$4.5             \\ 
\RL               & 445.54$\pm$9.98 & 0.63$\pm$0.34 & 24.0$\pm$8.6              & -2,117.79$\pm$11.34 & 0.25$\pm$0.14 & 15.2$\pm$4.1            \\ 
\CL               & {\bf443.48$\pm$9.56} & {\bf0.17$\pm$0.09} & {\bf7.1$\pm$5.1}             & {\bf-2,119.04$\pm$11.98} & {\bf0.19$\pm$0.16} & {\bf11.3$\pm$4.2}            \\ \hline
\multicolumn{1}{l|}{} & \multicolumn{3}{c|}{CA}                                                                         & \multicolumn{3}{c}{SC}                                                                          \\ \hline
BnB                & -113,421$\pm$1,599 & 2.45$\pm$0.62 & 116.3$\pm$22.0      & 171.47$\pm$12.67 & 1.27$\pm$1.01 & 75.9$\pm$30.6                         \\ 
LB                  & -111,113$\pm$1,835 & 4.43$\pm$0.81 & 233.3$\pm$22.3    & 171.54$\pm$12.85 & 1.30$\pm$0.98 & 102.4$\pm$28.5      \\ 
\RANDOM            & -109,253$\pm$1,697 & 6.03$\pm$1.02 & 181.9$\pm$26.2      & 174.15$\pm$12.94 & 2.78$\pm$1.30 & 99.8$\pm$35.3                     \\ 
\GRAPH             & -108,169$\pm$1,834 & 6.96$\pm$1.06 & 216.2$\pm$27.8      & 186.12$\pm$14.24 & 9.00$\pm$2.23 & 258.1$\pm$58.1           \\ 
\LBRELAX        & -114,268$\pm$1,512 & 1.72$\pm$0.57 & 125.3$\pm$13.6     & 170.98$\pm$12.38 & 1.00$\pm$0.88 & 54.8$\pm$25.6     \\ 
\DM            & -114,871$\pm$1,602 & 1.20$\pm$0.56 & 79.7$\pm$17.3  & 171.55$\pm$12.47 & 1.33$\pm$0.97 & 51.2$\pm$25.7            \\ 
\RL            & -108,776$\pm$1,813 & 6.44$\pm$1.04 & 191.7$\pm$27.0  & 171.35$\pm$12.29 & 1.22$\pm$0.85 & 67.5$\pm$22.6            \\ 
\CL & {\bf-115,513$\pm$1,621} & {\bf 0.65$\pm$0.32} & {\bf44.9$\pm$17.0} & {\bf170.15$\pm$12.12} & {\bf0.53$\pm$0.62} & {\bf21.5$\pm$17.5} \\ \hline
\end{tabular}
\end{table*}

\begin{table*}[htbp]
\scriptsize
\centering
\caption{Test results on small instances: Primal bound (PB), primal gap (PG) (in percent), primal integral (PI) at 60 minutes time cutoff, averaged over 100 instances and their standard deviations. \label{res::smalltable60_full}}

\begin{tabular}{c|rrr|rrr}
\hline
                       & \multicolumn{1}{c}{PB}             & \multicolumn{1}{c}{PG (\%)}         & \multicolumn{1}{c|}{PI} & \multicolumn{1}{c}{PB}              & \multicolumn{1}{c}{PG (\%)}         & \multicolumn{1}{c}{PI} \\ \hline

                       & \multicolumn{3}{c|}{MVC-S}                                                                        & \multicolumn{3}{c}{MIS-S}                                                                         \\ \hline
BnB                    & {448.63$\pm$9.58} & {1.32$\pm$0.43} & 66.1$\pm$13.1           & {-2,014.85$\pm$20.04} & {5.10$\pm$0.69} & 222.8$\pm$25.9           \\ 
LB                     & {453.45$\pm$11.81} & {2.35$\pm$1.30} & 102.2$\pm$35.9             & {-2,079.07$\pm$14.34} & {2.07$\pm$0.44} & 130.9$\pm$13.6             \\ 
\RANDOM                 & {447.06$\pm$11.21} & {0.96$\pm$1.26} & 38.0$\pm$44.8             & {-2,117.92$\pm$11.31} & {0.24$\pm$0.14} & 22.1$\pm$5.0             \\ 
\GRAPH                  & {447.14$\pm$10.83} & {0.98$\pm$1.20} & 42.9$\pm$44.0             & {-2,116.15$\pm$11.58} & {0.32$\pm$0.15} & 31.8$\pm$5.0             \\ 
\LBRELAX                & { 449.01$\pm$11.53} & {1.38$\pm$1.51} & 57.0$\pm$51.2             & {-2,109.17$\pm$11.17} & {0.65$\pm$0.20} & 46.9$\pm$6.5             \\ 
\DM       & {444.00$\pm$9.73} & {0.29$\pm$0.23} & {19.2$\pm$10.2}              & { -2,118.38$\pm$11.77} & {0.22$\pm$0.17} & { 19.4$\pm$5.8}             \\ 
\RL               & {445.45$\pm$9.99} & {0.61$\pm$0.34} & 29.6$\pm$11.5             & {-2,118.44$\pm$11.36} & {0.22$\pm$0.14} & 17.2$\pm$5.2            \\ 
{\bf\CL}               & {\bf 443.48$\pm$9.56} & {\bf 0.17$\pm$0.09} & {\bf 8.7$\pm$6.7}             & {\bf-2,119.78$\pm$12.14} & {\bf 0.15$\pm$0.15} & {\bf12.8$\pm$5.4}            \\ \hline
\multicolumn{1}{l|}{} & \multicolumn{3}{c|}{CA-S}                                                                         & \multicolumn{3}{c}{SC-S}                                                                          \\ \hline
BnB                 &    -113,608$\pm$1,611 & 2.28$\pm$0.59 & 137.4$\pm$25.9     & 171.22$\pm$12.50 & 1.13$\pm$0.95 & 86.7$\pm$37.9                         \\ 
LB                  & -111,342$\pm$1,732 & 4.23$\pm$0.75 & 272.1$\pm$26.9    & 171.39$\pm$12.81 & 1.22$\pm$0.97 & 113.7$\pm$35.2      \\ 
\RANDOM            & -109,397$\pm$1,684 & 5.90$\pm$1.02 & 235.6$\pm$34.9     & 173.95$\pm$12.98 & 2.67$\pm$1.29 & 124.3$\pm$45.4                    \\ 
\GRAPH             & -108,422$\pm$1,775 & 6.74$\pm$1.03 & 277.7$\pm$36.5     & 185.57$\pm$14.17 & 8.74$\pm$2.13 & 337.8$\pm$76.4          \\ 
\LBRELAX        & -114,348$\pm$1,516 & 1.65$\pm$0.57 & 140.5$\pm$18.3     & 170.74$\pm$12.35 & 0.86$\pm$0.83 & 63.2$\pm$31.6    \\ 
\DM            & -115,001$\pm$1,564 &1.09$\pm$0.51 & 90.0$\pm$20.8  & 171.55$\pm$12.47 & 1.33$\pm$0.97 & 63.2$\pm$34.3            \\ 
\RL            & -108,920$\pm$1,816 & 6.32$\pm$1.03 & 249.2$\pm$35.9  & 171.14$\pm$12.30 & 1.10$\pm$0.77 & 77.8$\pm$28.9            \\ 
{\bf\CL} & {\bf-115,513$\pm$1,621} & {\bf 0.65$\pm$0.32} & {\bf50.7$\pm$22.7} & {\bf170.11$\pm$12.10} &{\bf0.50$\pm$0.58} & {\bf26.2$\pm$12.8}\\ \hline
\end{tabular}
\end{table*}

\begin{table*}[htbp]
\scriptsize
\centering
\caption{Generalization results on large instances: Primal bound (PB), primal gap (PG) (in percent), primal integral (PI) at 15 minutes time cutoff, averaged over 100 instances and their standard deviations. \label{res::bigtable15}}

\begin{tabular}{c|rrr|rrr}
\hline
                       & \multicolumn{1}{c}{PB}             & \multicolumn{1}{c}{PG (\%)}         & \multicolumn{1}{c|}{PI} & \multicolumn{1}{c}{PB}              & \multicolumn{1}{c}{PG (\%)}         & \multicolumn{1}{c}{PI} \\ \hline

                       & \multicolumn{3}{c|}{MVC}                                                                        & \multicolumn{3}{c}{MIS}                                                                         \\ \hline
BnB            & 919.96$\pm$12.38 & 4.06$\pm$0.38 & 36.8$\pm$3.4                   & -3,888.39$\pm$20.62 & 8.24$\pm$0.31 & 76.3$\pm$2.8          \\ 
LB             & 907.06$\pm$12.46 & 2.69$\pm$0.36 & 32.7$\pm$3.2                   & -3,959.15$\pm$59.75 & 6.57$\pm$1.34 & 70.0$\pm$3.6              \\ 
\RANDOM        & 886.97$\pm$12.69 & 0.49$\pm$0.25 & 11.5$\pm$2.0                        & -4,215.32$\pm$15.86 & 0.52$\pm$0.12 & 12.4$\pm$1.0              \\ 
\GRAPH        & 888.28$\pm$12.61 & 0.64$\pm$0.26 & 18.0$\pm$2.3                       & -4,185.96$\pm$17.29 & 1.22$\pm$0.17 & 23.2$\pm$1.5             \\ 
\LBRELAX      & 901.37$\pm$12.66 & 2.08$\pm$0.30 & 30.1$\pm$2.8                      & -4,148.06$\pm$19.51 & 2.11$\pm$0.20 & 33.2$\pm$1.8             \\ 
\DM           & 886.32$\pm$12.63 & 0.42$\pm$0.26 & 12.6$\pm$1.8          & -4,203.74$\pm$16.80 & 0.80$\pm$0.17 & 14.8$\pm$1.7             \\ 
\RL           & 890.78$\pm$12.34 & 0.92$\pm$0.30 & 18.7$\pm$2.5                & -4,215.17$\pm$15.97 & 0.53$\pm$0.14 & 11.5$\pm$1.2            \\ 
\CL           & {\bf883.18$\pm$12.52} & {\bf0.06$\pm$0.05} & {\bf7.7$\pm$1.5}              & {\bf-4,220.96$\pm$15.68} & {\bf0.39$\pm$0.14} & {\bf6.8$\pm$1.5}             \\ \hline
\multicolumn{1}{l|}{} & \multicolumn{3}{c|}{CA}                                                                         & \multicolumn{3}{c}{SC}                                                                          \\ \hline
BnB                 & -194,128$\pm$14,403 & 15.43$\pm$6.20 & 164.4$\pm$11.8     & 110.42$\pm$7.44 & 2.92$\pm$1.49 & 63.3$\pm$12.2                         \\ 
LB                  & -203,872$\pm$4,522 & 11.18$\pm$1.72 & 149.9$\pm$8.6    & 117.36$\pm$8.84 & 8.58$\pm$2.85 & 89.3$\pm$19.3      \\ 
\RANDOM            & -215,183$\pm$2,670 & 6.26$\pm$0.74 & 75.8$\pm$6.0     & 112.91$\pm$7.72 & 5.04$\pm$2.03 & 59.9$\pm$16.8                    \\ 
\GRAPH             & -210,157$\pm$2,697 & 8.44$\pm$0.85 & 108.8$\pm$6.9     & 116.28$\pm$7.84 & 7.81$\pm$1.86 & 89.2$\pm$19.6           \\ 
\LBRELAX        & {\bf-222,638$\pm$4,846} & {\bf3.01$\pm$1.78} & 102.5$\pm$12.3      & 109.66$\pm$7.24 & 2.25$\pm$1.51 & 36.2$\pm$13.3     \\ 
\DM            & -211,938$\pm$3,323 & 7.67$\pm$1.22 & 89.9$\pm$8.9  & 109.12$\pm$6.97 & 1.79$\pm$1.26 & 32.4$\pm$10.7             \\ 
\RL            & -216,788$\pm$2,730 & 5.56$\pm$0.85 & {\bf58.1$\pm$6.9}   & 109.38$\pm$6.89 & 2.03$\pm$1.08 & 83.6$\pm$8.8            \\ 
\CL & -218,510$\pm$2,989 & {4.81$\pm$0.81} & 61.3$\pm$7.1   & {\bf107.95$\pm$6.78} & {\bf0.73$\pm$0.57} & {\bf23.1$\pm$8.6} \\ \hline
\end{tabular}
\end{table*}

\begin{table*}[htbp]
\scriptsize
\centering
\caption{Generalization results on large instances: Primal bound (PB), primal gap (PG) (in percent), primal integral (PI) at 30 minutes time cutoff, averaged over 100 instances and their standard deviations. \label{res::bigtable30}}

\begin{tabular}{c|rrr|rrr}
\hline
                       & \multicolumn{1}{c}{PB}             & \multicolumn{1}{c}{PG (\%)}         & \multicolumn{1}{c|}{PI} & \multicolumn{1}{c}{PB}              & \multicolumn{1}{c}{PG (\%)}         & \multicolumn{1}{c}{PI} \\ \hline

                       & \multicolumn{3}{c|}{MVC}                                                                        & \multicolumn{3}{c}{MIS}                                                                         \\ \hline
BnB             & 919.96$\pm$12.38 & 4.06$\pm$0.38 & 73.4$\pm$6.8                  & -3,888.39$\pm$20.62 & 8.24$\pm$0.31 & 150.5$\pm$5.6      \\ 
LB              & 900.15$\pm$12.32 & 1.95$\pm$0.35 & 52.6$\pm$6.0                    & -4,009.23$\pm$71.94 & 5.39$\pm$1.59 & 123.1$\pm$15.1             \\ 
\RANDOM         & 886.39$\pm$12.71 & 0.43$\pm$0.25 & 15.6$\pm$3.9                     & -4,225.74$\pm$15.63 & 0.28$\pm$0.10 & 15.8$\pm$1.8            \\ 
\GRAPH          & 886.89$\pm$12.79 & 0.48$\pm$0.23 & 22.9$\pm$3.9                     & -4,206.29$\pm$16.76 & 0.74$\pm$0.16 & 31.6$\pm$2.7            \\ 
\LBRELAX        & 887.64$\pm$12.21 & 0.57$\pm$0.23 & 39.4$\pm$4.4                      & -4,177.14$\pm$18.22 & 1.42$\pm$0.16 & 48.5$\pm$3.0            \\ 
\DM             & 885.58$\pm$12.65 & 0.33$\pm$0.26 & 15.9$\pm$4.0        & -4,216.32$\pm$17.30 & 0.50$\pm$0.17 & 20.4$\pm$3.0              \\ 
\RL             & 888.89$\pm$12.64 & 0.71$\pm$0.30 & 25.8$\pm$4.8              & -4,224.37$\pm$15.79 & 0.31$\pm$0.13 & 15.1$\pm$2.2           \\ 
\CL             & {\bf883.07$\pm$12.61} & {\bf0.05$\pm$0.04} & {\bf8.1$\pm$2.1}              & {\bf-4,226.65$\pm$15.56} & {\bf0.26$\pm$0.13} & {\bf9.7$\pm$2.6}            \\ \hline
\multicolumn{1}{l|}{} & \multicolumn{3}{c|}{CA}                                                                         & \multicolumn{3}{c}{SC}                                                                          \\ \hline
BnB                 & -216,772$\pm$13,060 & 5.58$\pm$5.42 & 257.1$\pm$56.4     & 109.39$\pm$7.26 & 2.02$\pm$1.36 & 84.4$\pm$22.2                         \\ 
LB                  & -206,526$\pm$3,750 & 10.03$\pm$1.39 & 245.1$\pm$19.2      & 116.43$\pm$8.97 & 7.84$\pm$2.88 & 162.6$\pm$39.2       \\ 
\RANDOM            & -216,326$\pm$2,603 & 5.76$\pm$0.74 & 129.4$\pm$12.1     & 111.71$\pm$7.65 & 4.02$\pm$1.86 & 100.6$\pm$32.0                    \\ 
\GRAPH             & -213,142$\pm$2,713 & 7.14$\pm$0.78 & 177.6$\pm$13.2     & 112.74$\pm$7.64 & 4.91$\pm$1.80 & 141.7$\pm$31.1           \\ 
\LBRELAX        & {\bf-225,154$\pm$4,366} & {\bf1.91$\pm$1.60} & 121.9$\pm$23.9     & 109.26$\pm$7.07 & 1.91$\pm$1.42 & 53.9$\pm$24.5    \\ 
\DM            & -214,495$\pm$3,148 & 6.56$\pm$1.01 & 154.0$\pm$17.9   & 109.04$\pm$6.94 & 1.72$\pm$1.19 & 48.1$\pm$21.3              \\ 
\RL            & -217,600$\pm$2,705 & 5.20$\pm$0.84 & 106.3$\pm$14.2    & 108.66$\pm$6.83 & 1.38$\pm$0.99 & 98.1$\pm$15.1           \\ 
\CL & {-223,257$\pm$2,667} & {2.74$\pm$0.71} & {\bf95.0$\pm$12.5}  &{\bf107.78$\pm$6.64} & {\bf0.58$\pm$0.45} & {\bf28.6$\pm$12.6} \\ \hline
\end{tabular}
\end{table*}

\begin{table*}[htbp]
\scriptsize
\centering
\caption{Generalization results on large instances: Primal bound (PB), primal gap (PG) (in percent), primal integral (PI) at 45 minutes time cutoff, averaged over 100 instances and their standard deviations. \label{res::bigtable45}}

\begin{tabular}{c|rrr|rrr}
\hline
                       & \multicolumn{1}{c}{PB}             & \multicolumn{1}{c}{PG (\%)}         & \multicolumn{1}{c|}{PI} & \multicolumn{1}{c}{PB}              & \multicolumn{1}{c}{PG (\%)}         & \multicolumn{1}{c}{PI} \\ \hline

                       & \multicolumn{3}{c|}{MVC}                                                                        & \multicolumn{3}{c}{MIS}                                                                         \\ \hline
BnB              & 907.44$\pm$12.77 & 2.73$\pm$0.43 & 107.2$\pm$9.4                  & -3,913.03$\pm$46.93 & 7.66$\pm$1.06 & 222.6$\pm$9.1            \\ 
LB               & 894.77$\pm$12.41 & 1.36$\pm$0.30 & 66.3$\pm$8.2                    & -4,063.18$\pm$54.80 & 4.11$\pm$1.18 & 165.2$\pm$25.7             \\ 
\RANDOM          & 886.15$\pm$12.71 & 0.40$\pm$0.24 & 19.2$\pm$5.9                    & -4,230.24$\pm$15.56 & 0.17$\pm$0.09 & 17.8$\pm$2.5              \\ 
\GRAPH           & 886.53$\pm$12.72 & 0.44$\pm$0.23 & 27.0$\pm$5.7                     & -4,215.85$\pm$16.16 & 0.51$\pm$0.16 & 37.1$\pm$3.9              \\ 
\LBRELAX         & 887.00$\pm$12.32 & 0.49$\pm$0.23 & 44.1$\pm$5.8                   & -4,191.17$\pm$17.76 & 1.09$\pm$0.16 & 59.7$\pm$4.2             \\ 
\DM              & 885.23$\pm$12.65 & 0.29$\pm$0.24 & 18.7$\pm$6.0                      & -4,222.04$\pm$16.64 & 0.36$\pm$0.16 & 24.2$\pm$4.3             \\ 
\RL              & 888.25$\pm$12.70 & 0.63$\pm$0.31 & 31.8$\pm$7.2                & -4,228.78$\pm$15.68 & 0.20$\pm$0.12 & 17.3$\pm$3.1            \\ 
\CL              & {\bf883.07$\pm$12.61} & {\bf0.05$\pm$0.04} & {\bf8.6$\pm$2.7}              & {\bf-4,230.20$\pm$15.19} & {\bf0.17$\pm$0.11} & {\bf11.6$\pm$3.6}             \\ \hline
\multicolumn{1}{l|}{} & \multicolumn{3}{c|}{CA}                                                                         & \multicolumn{3}{c}{SC}                                                                          \\ \hline
BnB                & -221,424$\pm$7,149 & 3.54$\pm$2.83 & 293.0$\pm$71.3      & 109.02$\pm$7.39 & 1.67$\pm$1.38 & 100.7$\pm$32.1                         \\ 
LB                  & -208,294$\pm$3,906 & 9.26$\pm$1.42 & 330.9$\pm$27.6    & 115.67$\pm$8.66 & 7.25$\pm$2.68 & 230.3$\pm$60.0      \\ 
\RANDOM            & -216,819$\pm$2,611 & 5.54$\pm$0.73 & 180.1$\pm$18.1       & 111.24$\pm$7.54 & 3.63$\pm$1.81 & 134.9$\pm$46.8                     \\ 
\GRAPH             & -214,331$\pm$2,641 & 6.63$\pm$0.83 & 239.2$\pm$19.7       & 111.96$\pm$7.60 & 4.25$\pm$1.78 & 182.5$\pm$43.6           \\ 
\LBRELAX        & -225,641$\pm$4,235 & 1.70$\pm$1.53 & 138.1$\pm$37.1      & 109.26$\pm$7.07 & 1.91$\pm$1.42 & 71.1$\pm$36.5      \\ 
\DM            & -216,705$\pm$3,062 & 5.59$\pm$0.97 & 208.7$\pm$25.7   & 109.04$\pm$6.94 & 1.72$\pm$1.19 & 63.6$\pm$31.8           \\ 
\RL            & -217,987$\pm$2,711 & 5.03$\pm$0.81 & 152.3$\pm$21.4   & 108.22$\pm$6.75 & 0.99$\pm$0.87 & 108.6$\pm$21.2            \\ 
\CL & {\bf-227,235$\pm$2,698} & {\bf1.01$\pm$0.54} & {\bf111.7$\pm$16.6}  & {\bf107.78$\pm$6.64} & {\bf0.58$\pm$0.45} & {\bf33.9$\pm$17.6} \\ \hline
\end{tabular}
\end{table*}

\begin{table*}[htbp]
\scriptsize
\centering
\caption{Generalization results on large instances: Primal bound (PB), primal gap (PG) (in percent) and primal integral (PI) at 60 minutes time cutoff, averaged over 100 instances and their standard deviations. \label{res::bigtable60_full}}

\begin{tabular}{c|rrr|rrr}
\hline
                       & \multicolumn{1}{c}{PB}             & \multicolumn{1}{c}{PG (\%)}         & \multicolumn{1}{c|}{PI} & \multicolumn{1}{c}{PB}              & \multicolumn{1}{c}{PG (\%)}         & \multicolumn{1}{c}{PI} \\ \hline

                       & \multicolumn{3}{c|}{MVC-L}                                                                        & \multicolumn{3}{c}{MIS-L}                                                                         \\ \hline
BnB                    & {904.41$\pm$12.95} & {2.41$\pm$0.40} & 130.2$\pm$11.1           & {-3,970.78$\pm$71.54} & {6.29$\pm$1.62} & 285.1$\pm$18.2           \\ 
LB                     & {893.56$\pm$12.62} & {1.22$\pm$0.30} & 77.8$\pm$10.1             & {-4,079.76$\pm$43.09} & {3.72$\pm$0.87} & 200.7$\pm$32.5             \\ 
\RANDOM                 & {886.00$\pm$12.74} & {0.38$\pm$0.24} & 22.7$\pm$8.0             & {\bf-4,232.68$\pm$15.42} & {\bf0.11$\pm$0.08} & 19.0$\pm$3.1             \\ 
\GRAPH                  & {886.34$\pm$12.67} & {0.42$\pm$0.23} & 30.9$\pm$7.6             & {-4,220.89$\pm$16.42} & {0.39$\pm$0.15} & 41.1$\pm$5.1             \\ 
\LBRELAX                & { 886.68$\pm$12.33} & {0.46$\pm$0.23} & 48.4$\pm$7.5              & {-4,199.04$\pm$17.54} & {0.91$\pm$0.16} & 68.6$\pm$5.5             \\ 
\DM              & {885.00$\pm$12.56} & {0.27$\pm$0.23} & {21.2$\pm$8.1}              & { -4,225.28$\pm$16.25} & {0.29$\pm$0.15} & { 27.1$\pm$5.5}             \\ 
\RL               & {887.90$\pm$12.67} & {0.59$\pm$0.30} & 37.3$\pm$9.6             & {-4,231.52$\pm$15.97} & {0.14$\pm$0.12} & 18.9$\pm$4.1            \\ 
{\bf\CL}               & {\bf 883.07$\pm$12.61} & {\bf 0.05$\pm$0.04} & {\bf 9.1$\pm$3.4}             & {-4,232.50$\pm$14.86} & { 0.12$\pm$0.11} & {\bf12.9$\pm$4.4}            \\ \hline
\multicolumn{1}{l|}{} & \multicolumn{3}{c|}{CA-L}                                                                         & \multicolumn{3}{c}{SC-L}                                                                          \\ \hline
BnB        &  -223,225$\pm$5,106 & 2.74$\pm$1.87 & 320.9$\pm$83.1          & 108.87$\pm$7.35 & 1.54$\pm$1.33 & 115.0$\pm$42.5            \\ 
LB         &   -208,500$\pm$3,976 & 9.17$\pm$1.43 & 414.0$\pm$36.9        & 115.12$\pm$8.77 & 6.80$\pm$2.73 & 293.5$\pm$79.7            \\ 
\RANDOM    &   -217,204$\pm$2,612 & 5.37$\pm$0.75 & 229.2$\pm$24.4          & 110.88$\pm$7.55 & 3.31$\pm$1.79 & 166.4$\pm$61.3         \\ 
\GRAPH     &   -214,926$\pm$2,649 & 6.37$\pm$0.86 & 297.5$\pm$26.9          & 111.49$\pm$7.51 & 3.85$\pm$1.74 & 218.9$\pm$56.7          \\ 
\LBRELAX   &   -225,848$\pm$4,201 & 1.61$\pm$1.50 & 153.0$\pm$50.3          & 109.26$\pm$7.07 & 1.91$\pm$1.42 & 88.3$\pm$48.9            \\ 
\DM        &  -219,074$\pm$3,278 & 4.56$\pm$0.98 & 254.2$\pm$33.4    & 109.04$\pm$6.94 & 1.72$\pm$1.19 & 79.1$\pm$42.4            \\ 
\RL        &  -218,273$\pm$2,725 & 4.91$\pm$0.81 & 197.0$\pm$28.5     & 107.87$\pm$6.74 & 0.66$\pm$0.72 & 116.2$\pm$27.1            \\ 
{\bf\CL} & {\bf-229,331$\pm$2,800} & {\bf0.09$\pm$0.10} & {\bf116.1$\pm$18.0} & {\bf107.78$\pm$6.64} & {\bf0.58$\pm$0.45} & {\bf39.2$\pm$23.2}\\ \hline
\end{tabular}
\end{table*}


\end{document}